\documentclass[10pt]{article}
\usepackage[utf8]{inputenc} 
\usepackage[T1]{fontenc}    
\usepackage{comment}      
\usepackage{url}            
\usepackage{nicefrac}       
\usepackage{microtype}     
\usepackage{fullpage}

\usepackage{xcolor}
\usepackage{framed}
\colorlet{shadecolor}{pink} 
\usepackage{authblk}
\usepackage{adjustbox}
\usepackage{comment}
\usepackage{multirow}
\usepackage{natbib}
\usepackage{tabularx}
\usepackage{color, colortbl}
\usepackage{graphicx}
\usepackage{soul}
\usepackage{subcaption}
\usepackage{booktabs}
\usepackage{tablefootnote}
\usepackage[most]{tcolorbox}
\tcbset{colback=white, colframe=black, boxrule=0.5pt, arc=2mm, left=1mm, right=1mm, top=1mm, bottom=1mm}
\usepackage{graphicx}%
\usepackage{multirow}%
\usepackage{amsmath,amssymb,amsfonts}%
\usepackage{amsthm}%
\usepackage{mathrsfs}%
\usepackage[title]{appendix}%
\usepackage{xcolor}%
\usepackage{textcomp}%
\usepackage{manyfoot}%
\usepackage{booktabs}%
\usepackage{algorithm}%
\usepackage{algpseudocode}
\usepackage{listings}%

\usepackage{wrapfig}
\usepackage{pifont}
\usepackage{subcaption}
\usepackage[table]{xcolor}
\definecolor{lightblue}{RGB}{235,245,255}
\usepackage[colorlinks,linkcolor=magenta,citecolor=blue, pagebackref=true,backref=true]{hyperref}

\usepackage{amsmath,amsthm,amssymb,amsfonts}
\usepackage{enumerate}
\usepackage{cleveref}
\usepackage{comment}
\usepackage{bm}
\usepackage{pifont}
\usepackage{adjustbox}

\newtheorem{theorem}{Theorem}
\newtheorem{corollary}{Corollary}
\usepackage{enumitem}
\usepackage{natbib}
\theoremstyle{plain} 
\newcommand{\vertiii}[1]{{\left\vert\kern-0.25ex\left\vert\kern-0.25ex\left\vert #1 
\right\vert\kern-0.25ex\right\vert\kern-0.25ex\right\vert}}

\definecolor{antiquewhite}{rgb}{0.98, 0.92, 0.84} 
\definecolor{blizzardblue}{rgb}{0.67, 0.9, 0.93}

\title{\bf Collaboration through Diversity and Specialization via Orthogonality for Improved LoRA-based MoE}

\author{Fuli Qiao \qquad Mehrdad Mahdavi \vspace*{.2em} \\ 
The Pennsylvania State University \vspace*{.2em} \\ 
\texttt{ \{fvq5015,mzm616\}@psu.edu}
}
\date{}

\begin{document}

\maketitle

\begin{abstract}
  Combining the Mixture-of-Experts (MoE) paradigm with Low-Rank Adaptation (LoRA) enables parameter-efficient adaptation of large language models (LLMs) to downstream tasks. However, existing LoRA-based MoE approaches often suffer from expert interference and weak specialization, where experts produce correlated outputs, leading to overlapping representations and reduced effective capacity. A key challenge is to balance collaboration and specialization: experts jointly capture task-specific knowledge while reducing redundancy and conflicts across experts. To address this challenge, we propose a principled framework that characterizes expert interactions in a data-dependent manner. We show that redundancy arises from correlated expert outputs and propose an explicit regularization objective that encourages orthogonality, promoting effective specialization and mitigating interference. Furthermore, built on an asymmetric LoRA framework with shared low-rank projections, we introduce a complementary regularization that encourages diversity in the shared representations. Our experiments across diverse benchmarks demonstrate that our method encourages representational specialization and improves task performance while maintaining the efficiency of parameter-efficient fine-tuning.
\end{abstract}

\section{Introduction}
Large Language Models (LLMs) have been powerful in multiple realistic scenarios~\citep{touvron2023llama,achiam2023gpt,grattafiori2024llama}, but high fine-tuning costs make adapting these models to resource-constrained devices for downstream tasks impractical~\citep{hu2022lora,zhao2024galore}. Low-Rank Adaptation (LoRA) has become a popular technique to solve the challenges when insufficient resources cannot support full fine-tuning (FFT)~\citep{hu2022lora}. However, the restricted update parameterization can limit adaptation, motivating efforts to improve its capacity and optimization~\citep{wang2025lorapro}. Combining LoRA with Mixture-of-Experts (MoE) framework offers a promising direction
, which uses multiple low-rank experts provide alternative updates and an input-dependent router determines how to combine them~\citep{zadouri2024pushing,liu2024adamole,tian2024hydralora,fan2025make}.

However, more experts do not necessarily give the router more meaningful choices. Different expert parameters can produce similar responses on the data, allowing different routing weights to yield nearly identical updates. Prior work has identified insufficient expert specialization and expert overlap as challenges in LoRA-based MoE~\citep{guo2025advancing, su2026variational}. Routing determines how expert outputs are combined, but does not directly ensure that those outputs are distinguishable~\citep{fan2025make}. Similarly, balanced expert utilization does not by itself establish that experts contribute different functions~\citep{guo2025advancing}. Understanding a LoRA mixture therefore requires examining the expert responses that the router actually combines.

Moreover, the LoRA factorization reveals a second source of redundancy. In a LoRA update $\Delta\boldsymbol{W}=\boldsymbol{B}\boldsymbol{A}$, $\boldsymbol{A}$ projects inputs into a low-dimensional space and $\boldsymbol{B}$ maps these features back to the output space.  Recent work~\citep{tian2024hydralora} leverages this structure to share one projection $\boldsymbol{A}$ matrix across multiple expert-specific reconstruction matrices $\boldsymbol{B}_i$ for each expert $i$, effectively reducing parameter redundancy. But even distinct reconstruction matrices can produce similar outputs when the shared features contain little variation along the directions in which those matrices differ. Preserving these directions is also insufficient to ensure distinguishable mixtures, since different combinations of distinct expert outputs can still produce the same residual. These perspectives shift the focus from expert multiplicity to the geometry of expert contributions and motivates our central question: \emph{How can we encourage distinguishable expert contributions while preserving informative shared features in a parameter-efficient LoRA mixture?}

We address this question through two theoretical results that establish when expert-output decorrelation yields distinguishable mixtures and quantitative routing-recovery bounds, and when a shared projection preserves differences between expert reconstruction matrices in their outputs. Our first result connects expert-output decorrelation to the distinguishability of routing weights. For fixed expert responses, we show that sufficiently decorrelated expert outputs with nonvanishing magnitudes make changes in routing weights observable in the combined residual. Specifically, we derive a lower bound on mixture-output separation in terms of routing-weight separation, which strengthens as correlations decrease while the output-norm bounds remain fixed. This bound establishes a unique oracle mixture for the quadratic residual objective and ensures that small excess residual loss implies small routing-weight error. This result motivates penalizing correlations among expert responses on activations. Our second result explains why output differentiation must be considered alongside the shared representation. We show that controlling the scale and determinant of the shared projection preserves reconstruction differences in mean-square output, provided the input distribution has sufficient variation in the retained directions. Together, these two results motivate reducing output redundancy while preserving informative shared directions. They provide conditional design principles rather than guarantees of training convergence or task-level specialization.

Building on these principles, we propose \textbf{OD-MoLE} (\textbf{O}rthogonality and \textbf{D}iversity for \textbf{M}ixture-\textbf{o}f-\textbf{L}oRA \textbf{E}xperts), a simple yet effective framework that promotes both specialization and collaboration. Built on the shared-projection architecture, OD-MoLE combines two complementary regularizers. An output-space orthogonality penalty reduces squared cosine similarities between expert responses on input activations. A shared-projection diversity penalty uses a sampled log determinant to encourage diversity within the shared low-rank representation. The first acts on the responses that the router combines, while the second acts on the features from which those responses are constructed. Our contribution is to connect these regularization targets to distinct properties of LoRA mixtures and evaluate their combined effect, rather than to introduce orthogonality or log-determinant regularization themselves. Experiments compare OD-MoLE with existing methods under both all-expert composition and selective routing, with ablations examining the contribution of each regularizer.

\paragraph{Contributions.}This paper makes the following main contributions:    
\begin{itemize}
\item We establish sufficient conditions under which expert-output decorrelation makes routing weights distinguishable and controls oracle routing-weight error through excess quadratic residual loss (Theorem~\ref{thm:routing-recovery}). We also show how scale-controlled shared-projection diversity preserves existing reconstruction differences in mean-square output (Theorem~\ref{thm:shared-diversity}).
\item Based on this analysis, we propose OD-MoLE, a novel asymmetric LoRA mixture that couples output orthogonality with projection diversity to address output redundancy and shared-feature collapse, respectively.
\item Extensive experiments demonstrate that OD-MoLE effectively achieves specialization and improves performance across diverse benchmarks while maintaining parameter efficiency.
\end{itemize}

\section{Related Work}\vspace{-0.25cm}
\paragraph{Parameter-Efficient Fine-tuning}Parameter-efficient fine-tuning (PEFT) has emerged to improve performance of LLMs under resource-constrained settings. LoRA~\citep{hu2022lora} has been a successful method and has various variants, which fundamentally introduce new low-rank trainable parameters within the existing model layer while keeping original parameters frozen. AdaLoRA~\citep{zhang2023adaptive} dynamically allocates rank across layers based on weight importance score, DoRA~\citep{liu2024dora} decomposes LoRA updates into magnitude and direction to improve stability during optimization, LoRA+~\citep{hayou2024lora} improves LoRA by setting different learning rates for LoRA adapter matrices, PiSSA~\citep{meng2024pissa} utilizes SVD to initialize LoRA weights with principal components of the pre-trained weight for preserving knowledge, and LoRA-Pro~\citep{wang2025lorapro} aligns LoRA's updates with full fine-tuning gradients. Despite these advances, a noticeable performance gap remains between LoRA-based PEFT and full fine-tuning.

\paragraph{Integration of MoE and LoRA}The integration of Mixture-of-Experts (MoE) and LoRA is a notable trend in recent advancements aimed at enhancing the performance of LLMs~\citep{wu2024mixture}. MoLoRA~\citep{wu2024mixture} proposes to integrate the MoE layer with LoRA modules, demonstrating that LoRA experts enhance specialization.  AdaMoLE~\citep{liu2024adamole} extends this idea by introducing learnable thresholds for dynamic expert selection. HydraLoRA~\citep{tian2024hydralora} adopts an asymmetric LoRA MoE framework where shared matrix $\boldsymbol{A}$ is combined with multiple expert-specific matrices $\boldsymbol{B}$, effectively reducing the number of expert parameters. GOAT~\citep{fan2025make} introduces an SVD-based MoE framework that adaptively integrates relevant priors and addresses weight misalignment and gradient dynamics through a theoretical scaling strategy. ~\citet{guo2025advancing} introduce complementary orthogonality and variance losses to encourage distinct expert token assignments and more discriminative routing decisions. FlyLoRA~\citep{zou2025flylora} is an implicit LoRA MoE that enables rank-wise experts in matrix $\boldsymbol{B}$ and a fixed sparse random projection for matrix $\boldsymbol{A}$ as an implicit router to reduce parameter interference.

\section{Preliminaries}
\label{sec:preliminaries}
In this section, we introduce the notations and briefly review the Mixture-of-Experts (MoE) formulation and its low-rank adaptation parameterization.

\paragraph{Problem Definition.}In standard MoE architecture, the output $f(\boldsymbol{x})\in\mathbb{R}^{d}$  for an input $\boldsymbol{x} \in \mathbb{R}^k$ is formulated as the weighted combination of $N$ specialized networks $\{f_i\}_{i=1}^{N}$: 
\begin{equation}
    f(\boldsymbol{x}) = \sum\nolimits_{i=1}^{N} g_{i}(\boldsymbol{x}) \, f_{i}(\boldsymbol{x}),
\end{equation}
where $f_{i}(\boldsymbol{x})$ is expert $i$ output, and $g_{i}(\boldsymbol{x})$ is a nonnegative gating coefficient for expert $i$. We assume that the router produces a probability vector $\boldsymbol{g}(\boldsymbol{x})\in\Delta_{N}=\{\boldsymbol{g}\in\mathbb{R}^{N}:g_i\geq 0, \boldsymbol{1}^{\top}\boldsymbol{g}=1\}$. A common parameterization uses learnable scoring functions $\{\phi_i\}_{i=1}^{N}$, often parameterized as:
\begin{align}
    g_{i}(\boldsymbol{x})=
  \text{exp}(\phi_i(\boldsymbol{x}))/\sum\nolimits_{j=1}^N\text{exp}(\phi_j(\boldsymbol{x}))
\end{align}
\paragraph{LoRA MoE.}In LoRA MoE framework, we leverage the efficiency of LoRA to parameterize individual experts within the mixture. Given a frozen pretrained model weight $\boldsymbol{W}_0\in \mathbb{R}^{d \times k}$, each expert $i$ is defined by a modular additive update. Decompose update $\Delta \boldsymbol{W}_i$ into two low-rank matrices $\boldsymbol{B}_i \in \mathbb{R}^{d \times r}, \boldsymbol{A}_i \in \mathbb{R}^{r \times k}$, where $r \ll \min (d, k)$. The effective weight for expert $i$ becomes 
\begin{align}
\boldsymbol{W}_i=\boldsymbol{W}_0+\Delta \boldsymbol{W}_i=\boldsymbol{W}_0+\frac{\gamma}{r} \boldsymbol{B}_i \boldsymbol{A}_i,
\end{align}
where scaling factor $\gamma$ scales update and we define $c=\gamma/r$ as the residual scaling factor. By applying this to input $\boldsymbol{x}$, the expert-specific transformation $f_i(\boldsymbol{x})$ becomes $ f_{i}(\boldsymbol{x})=\boldsymbol{W}_i \boldsymbol{x}=\boldsymbol{W}_0 \boldsymbol{x}+\gamma/r \boldsymbol{B}_i \boldsymbol{A}_i\boldsymbol{x}$. Here, $\boldsymbol{W}_0$ remains frozen during training, while only $\{\boldsymbol{A}_i, \boldsymbol{B}_i\}$ are updated.

Substituting this into the global MoE aggregation formula, and noting that the frozen term $\boldsymbol{W}_0\boldsymbol{x}$ is independent of expert index $i$, we can simplify by factoring out the shared linear transformation: 
\begin{align}
    f_{\mathrm{MoE}}(\boldsymbol{x}) = \boldsymbol{W}_0 \boldsymbol{x}+c\sum\nolimits_{i=1}^{N} g_{i}(\boldsymbol{x})\left(\boldsymbol{B}_i \boldsymbol{A}_i\boldsymbol{x}\right).
\end{align}
This parameterization effectively reduces per-expert training overhead to $\{\boldsymbol{A}_i, \boldsymbol{B}_i\}$, significantly enhancing parameter efficiency of MoE layer while maintaining the expressive capacity afforded by gating-dependent low-rank residuals. However, the integration of LoRA into mixture paradigms introduces distinct challenges. Specifically, we need to evaluate whether this architecture successfully partitions latent space among experts or if it inadvertently promotes redundant, correlated behaviors.
\section{Theoretical Analysis and Design Principles}
\label{sec:theory} 
We now analyze the shared-projection LoRA MoE through (1) when different routing weights produce distinguishable mixtures, and (2) when the shared projection preserves differences between experts. The insights from this analysis motivate the design of our proposed method in Section~\ref{sec:OD-MoLE}. 

For a model input $\boldsymbol{x}$, let $\boldsymbol{h}=\boldsymbol{h}_m(\boldsymbol{x})$ denote the input activation to module $m$. Suppressing the module index, we define the unscaled expert residual output of expert $i$ as $\boldsymbol{z}_i(\boldsymbol{h})=\boldsymbol{B}_i\boldsymbol{A}\boldsymbol{h}$, and collects all expert residuals as columns of $\boldsymbol{Z}(\boldsymbol{h})=[\boldsymbol{z}_1(\boldsymbol{h}),\dots,\boldsymbol{z}_N(\boldsymbol{h})]$. The adapted layer output is $\boldsymbol{W}_0\boldsymbol{h}+c\boldsymbol{Z}(\boldsymbol{h})\boldsymbol{g}(\boldsymbol{h})$. For fixed $\boldsymbol{h}$ and fixed expert parameters, consider
\begin{equation}
    Q_{\boldsymbol{h}}(\boldsymbol{g})=\frac{1}{2}\|c\boldsymbol{Z}(\boldsymbol{h})\boldsymbol{g}-\boldsymbol{t}_{\boldsymbol{h}}\|_2^2,\quad \boldsymbol{g}\in\Delta_{N}
    \label{eq:Q_h}
\end{equation}
where $\boldsymbol{t}_{\boldsymbol{h}}\in\mathbb{R}^d$ is a fixed target residual. For squared-error prediction with target $\boldsymbol{y}$, $\boldsymbol{t}_{\boldsymbol{h}}=\boldsymbol{y}-\boldsymbol{W}_0\boldsymbol{h}$ makes $Q_{\boldsymbol{h}}$ the prediction loss. For hidden-layer adaptation, it is an analytical surrogate.
\paragraph{Expert-output Decorrelation and Router Recovery.}We first examine whether different routing vectors produce distinguishable mixtures at a fixed input activation $\boldsymbol{h}$. The Hessian of $Q_{\boldsymbol{h}}$ (Eq.~\ref{eq:Q_h}) with respect to routing weights is $c^2\boldsymbol{Z}(\boldsymbol{h})^\top\boldsymbol{Z}(\boldsymbol{h})$, Thus, expert-output alignment determines the curvature of the routing objective. To quantify directional alignment, assume $N\geq2$ and $\boldsymbol{z}_i(\boldsymbol{h})\neq 0$ for every expert, and define average squared cosine similarity
\begin{equation}
    \omega(\boldsymbol{h})=\frac{1}{N(N-1)}\sum_{i\neq j}\left(\frac{\boldsymbol{z}_i(\boldsymbol{h})^\top\boldsymbol{z}_j({\boldsymbol{h}})}{\|\boldsymbol{z}_i(\boldsymbol{h})\|_2\|\boldsymbol{z}_j({\boldsymbol{h}})\|_2}\right)^2, \quad \delta(\boldsymbol{h})=\sqrt{N(N-1)\omega(\boldsymbol{h})}
\end{equation}
Here quantity $\omega(\boldsymbol{h})$ measures average squared similarity between expert residuals, treating positive and negative alignment symmetrically. The quantity $\delta(\boldsymbol{h})$ measures how far the normalized output Gram matrix is from the identity in Frobenius norm, giving the following sufficient condition with proof provided in Appendix~\ref{app:routing-recovery-proof}.
\begin{theorem}[Output decorrelation and routing recovery]
\label{thm:routing-recovery}
Fix input activation $\boldsymbol{h}$, the expert parameters, the target residual  $\boldsymbol{t}_{\boldsymbol{h}}$, and the scaling factor $c>0$. Suppose that $0<\underline{z}\leq\|\boldsymbol{z}_i(\boldsymbol{h})\|_2 \leq\overline{z}$ for every expert $i$ and that $\delta(\boldsymbol{h})<1$. Then, for any $\boldsymbol{g},\widehat{\boldsymbol{g}}\in\Delta_N$,
\begin{equation}
\|c\boldsymbol{Z}(\boldsymbol{h})(\boldsymbol{g}-\widehat{\boldsymbol{g}})\|_2^2 \geq c^2\underline{z}^2(1-\delta(\boldsymbol{h})) \|\boldsymbol{g}-\widehat{\boldsymbol{g}}\|_2^2.
\label{eq:output-separation}
\end{equation}
Moreover, $Q_{\boldsymbol{h}}$ has a unique minimizer $\boldsymbol{g}^{*}(\boldsymbol{h})$ on $\Delta_N$, and each $\boldsymbol{g}\in\Delta_{N}$ satisfies
\begin{equation}
\|\boldsymbol{g}-\boldsymbol{g}^{*}(\boldsymbol{h})\|_2^2\leq\frac{2[Q_{\boldsymbol{h}}(\boldsymbol{g})-Q_{\boldsymbol{h}}(\boldsymbol{g}^{*}(\boldsymbol{h}))]
}{c^2\underline{z}^2(1-\delta(\boldsymbol{h}))
}.
\label{eq:routing-recovery}
\end{equation}
\end{theorem}
Theorem~\ref{thm:routing-recovery} contains two statements of different scope. Eq.~\ref{eq:output-separation} bounds mixture-output separation at fixed input and expert outputs, independently of the downstream loss. Eq.~\ref{eq:routing-recovery} shows a recovery guarantee for the quadratic residual objective $Q_{\boldsymbol{h}}$ only. To extend the separation bound to selective routing, let $N\geq2$, $1\leq K\leq N$, and define
\begin{equation}
\Delta_N^K=\bigl\{\boldsymbol{g}\in\Delta_N:|\operatorname{supp}(\boldsymbol{g})|\leq K\bigr\}.
\label{eq:topk-simplex}
\end{equation}
The difference between two vectors in $\Delta_N^K$ is supported on at most $\kappa=\min\{2K,N\}$ experts. Let $\boldsymbol{R}_{\mathcal{T}}$ be the normalized Gram matrix of the outputs indexed by $\mathcal{T}$, with entries $R_{ij}=\boldsymbol{z}_i^\top\boldsymbol{z}_j/(\|\boldsymbol{z}_i\|_2\|\boldsymbol{z}_j\|_2)$. Define
\begin{equation}
\delta_{\mathcal{T}}(\boldsymbol{h})=\|\boldsymbol{R}_{\mathcal{T}}-\boldsymbol{I}\|_{\mathrm{F}},\qquad\delta_{2K}(\boldsymbol{h})=\max_{\substack{\mathcal{T}\subseteq[N]\\
1\leq|\mathcal{T}|\leq\kappa}}\delta_{\mathcal{T}}(\boldsymbol{h}),
\label{eq:delta2K}
\end{equation}
and
\begin{equation}
m_{\Delta}^{K}(\boldsymbol{h})=\inf_{\substack{\boldsymbol{g},\widehat{\boldsymbol{g}}\in\Delta_N^K\\
\boldsymbol{g}\neq\widehat{\boldsymbol{g}}}}\frac{\|\boldsymbol{Z}(\boldsymbol{h})(\boldsymbol{g}-\widehat{\boldsymbol{g}})\|_2^2}{\|\boldsymbol{g}-\widehat{\boldsymbol{g}}\|_2^2}.
\label{eq:mDeltaK}
\end{equation}
The following corollary accounts for sparsity through $\delta_{2K}(\boldsymbol{h})$ and relates separation across alternative supports to the all-pairs orthogonality penalty.
\begin{corollary}[Output decorrelation and top-$K$ routing separation]
\label{cor:topk}
Fix $\boldsymbol{h}$, the expert parameters, and the scaling factor $c>0$. Suppose that $0<\underline z\le\|\boldsymbol{z}_i(\boldsymbol{h})\|_2\le\overline z$ for every expert $i$.
\begin{enumerate}
\item[(i)] If $\delta_{2K}(\boldsymbol{h})<1$, then for any $\boldsymbol{g},\widehat{\boldsymbol{g}}\in\Delta_N^K$,
\begin{equation}
\|c\bm Z(\boldsymbol{h})(\boldsymbol{g}-\widehat{\boldsymbol{g}})\|_2^2\ge c^2\underline z^2\bigl(1-\delta_{2K}(\boldsymbol{h})\bigr)\|\boldsymbol{g}-\widehat{\boldsymbol{g}}\|_2^2,
\label{eq:topk-sep}
\end{equation}
thus $m_\Delta^K(\boldsymbol{h})\ge\underline z^2\bigl(1-\delta_{2K}(\boldsymbol{h})\bigr)$.
\item[(ii)] For every pair of experts $i\neq j$,
\begin{equation}
\begin{aligned}
m_\Delta^K(\boldsymbol{h})&\le\frac12\|\boldsymbol{z}_i(\boldsymbol{h})-\bm z_j(\boldsymbol{h})\|_2^2.
\end{aligned}
\label{eq:topk-upper}
\end{equation}
If $K<N$, the ratio in Eq.~\ref{eq:mDeltaK} takes this value for two routing vectors that each have exactly $K$ positive entries and neither of which selects $i$ and $j$ together.
\item[(iii)] The threshold quantity satisfies
\begin{equation}
2\max_{i\neq j}R_{ij}(\boldsymbol{h})^2\ \le\ \delta_{2K}(\boldsymbol{h})^2\ \le\ \min\Bigl\{\kappa(\kappa-1)\max_{i\neq j}R_{ij}(\boldsymbol{h})^2,\ N(N-1)\,\omega(\boldsymbol{h})\Bigr\}.
\label{eq:topk-pairs}
\end{equation}
In particular, $\delta_{2K}(\boldsymbol{h})\le\delta(\boldsymbol{h})$, with equality when $2K\ge N$.
\end{enumerate}
\end{corollary}
Since $\delta_{2K}(\boldsymbol{h})\leq\delta(\boldsymbol{h})$, the top-$K$ separation bound requires a weaker sufficient condition. Part~(ii) shows that separation across alternative supports still depends on every pair of expert outputs, while part~(iii) relates the sufficient condition to the all-pairs
orthogonality penalty. Appendix~\ref{sec:topk} provides the proof and extends the quadratic recovery guarantee to a fixed support and, for realizable targets, across supports.

The norm lower bound controls output scale, even orthogonal outputs provide weak separation if their magnitudes approach zero. This bound is an assumption that the training objective does not enforce, so both statements are conditional on it. Appendix~\ref{app:orth-analysis} verifies that near-zero expert outputs do not occur on the evaluated activations in Tab.~\ref{tab:min-norm}. The condition $\delta(\boldsymbol{h})<1$ requires $\omega(\boldsymbol{h})<1/[N(N-1)]$ and is sufficient rather than necessary for distinguishable mixtures, as opposite expert outputs can still uniquely determine routing weights (see \hyperref[app:example-opposite-outputs]{Appendix, Example~1}). Under the theorem's assumptions, small excess residual loss implies proximity to the best mixture weights for $Q_{\boldsymbol{h}}$, motivating penalizing expert-output alignment directly on input activations, since even orthogonal residuals can yield identical predictive distributions under different routing weights (\hyperref[app:example-task-predictions]{Appendix, Example~2}). The oracle need not represent a semantic task assignment, and the result does not guarantee that joint training achieves a small loss gap.
We therefore test routing distinguishability directly on models trained with cross-entropy in Tab.~\ref{tab:orth-mechanism}. 
\paragraph{Shared-projection diversity and expert distinguishability.}
We next examine whether differences between expert matrices remain visible after applying the shared projection. Even substantially different matrices $\boldsymbol{B}_i$ and $\boldsymbol{B}_j$ can produce identical outputs if their differences act only on latent directions that the input distribution does not activate. Assume $\mathbb{E}\|\boldsymbol{h}\|_{2}^{2}<\infty$, and define $\boldsymbol{\Sigma}_{\boldsymbol{h}}=\mathbb{E}\left[\boldsymbol{h}\boldsymbol{h}^\top\right]$ and
$\boldsymbol{K}_{\boldsymbol{A}}=\boldsymbol{A}\boldsymbol{\Sigma}_{\boldsymbol{h}}\boldsymbol{A}^\top$. The smallest eigenvalue of $\boldsymbol{K}_{\boldsymbol{A}}$ measures the least expected squared activation along a unit latent direction. In particular,
\begin{equation}
\mathbb{E}_h\|\boldsymbol{z}_i(\boldsymbol{h})-\boldsymbol{z}_j({\boldsymbol{h}})\|_2^2
\geq\lambda_{\min}(\boldsymbol{K}_{\boldsymbol{A}})
\|\boldsymbol{B}_i-\boldsymbol{B}_j\|_{\mathrm F}^2.
\label{eq:expert_differences}
\end{equation}
The following result bounds this eigenvalue using the shared projection and the input distribution.

\begin{theorem}[Shared diversity preserves expert differences]
\label{thm:shared-diversity}
Suppose $r>1$, $\text{tr}(\boldsymbol{A}\boldsymbol{A}^\top)=r$, and $\log\det(\boldsymbol{A}\boldsymbol{A}^\top)\geq-\eta$, where $\eta\geq0$. Assume
$\boldsymbol{u}^\top\boldsymbol{\Sigma}_{\boldsymbol{h}}\boldsymbol{u}\geq\nu\|\boldsymbol{u}\|_2^2$ for every $\boldsymbol{u}\in\text{row}(\boldsymbol{A})$, with $\nu>0$. Then
\begin{equation}
\lambda_{\min}(\boldsymbol{K}_{\boldsymbol{A}})\geq\nu e^{-\eta}\left(\frac{r-1}{r}\right)^{r-1}.
\label{eq:shared-diversity}
\end{equation}
\end{theorem}
The trace fixes the total squared singular-value scale, and the determinant prevents individual directions from collapsing at that scale. The input assumption ensures nonzero second-moment energy in the directions retained by $\boldsymbol{A}$. The trace condition is a normalization and can be removed, which makes the role of scale explicit. Corollary~\ref{cor:shared-diversity-scale} in Appendix~\ref{app:shared-diversity-proof} removes it and factors the bound into a scale term and a shape term that depends only on the normalized matrix. Our penalty is applied to the raw matrix as a practical surrogate, and Tab.~\ref{tab:div-mechanism} verifies that it improves scale-invariant statistics of the spectrum.

Together with Eq.~\ref{eq:expert_differences}, Theorem~\ref{thm:shared-diversity} shows that these controls preserve existing expert differences in mean square. But they do not create differences between identical experts or imply the pointwise routing condition. For example, one expert can equal the average of two others while all pairs retain positive mean-square output differences, allowing distinct routing weights to produce identical mixtures (\hyperref[app:example-ambiguous-mixture]{Appendix, Example~3}). Thus, the two results motivate complementary regularizers as output decorrelation discourages redundant mixtures and shared-projection diversity helps preserve the directions through which experts differ. Proofs are provided in Appendix~\ref{app:odmole-theory}, and the connection to the sampled log-determinant penalty is detailed in Appendix~\ref{app:sampled-determinants}.

\section{OD-MoLE: Orthogonality and Diversity for LoRA-based MoE}\label{sec:OD-MoLE}
Building on Section~\ref{sec:theory}, we introduce OD-MoLE, which combines two regularizers in a shared-projection LoRA mixture. Output-space orthogonality discourages redundant expert responses, while shared-projection diversity encourages the projection to retain distinct latent directions. These objectives are motivated by the routing-recovery and expert-difference preservation results, respectively. We first describe the architecture and then define the two regularizers.

\paragraph{Asymmetric LoRA MoE Formulation.}We adopt an asymmetric LoRA MoE architecture motivated by the distinct functional roles of LoRA components to navigate the trade-off between expert specialization and collaborative diversity. In LoRA, $\boldsymbol{A}$ projects inputs into a low-rank feature space and $\boldsymbol{B}$ reconstructs this representation into the output space. Prior work has shown their asymmetric roles that optimizing $\boldsymbol{B}$ is particularly important for capturing task-specific variations~\citep{zhu2024asymmetry}. This observation motivates an MoE with a shared $\boldsymbol{A}$ and multiple expert-specific $\boldsymbol{B}_i$ matrices. For an input activation $\boldsymbol{h}\in\mathbb{R}^{k}$, the effective weight update is
\begin{equation}
\Delta \boldsymbol{W}=c\sum_{i=1}^{N} g_i(\boldsymbol{h}) \boldsymbol{B}_i \boldsymbol{A},
\end{equation}
where $\boldsymbol{A} \in \mathbb{R}^{r \times k}$, $\boldsymbol{B}_i \in \mathbb{R}^{d \times r}$, and $c=\gamma/r$. Sharing $\boldsymbol{A}$ reduces parameter redundancy, while the expert-specific matrices $\boldsymbol{B}_i$ provide different mappings from shared features to output residuals. This structure also separates the two regularization targets: output orthogonality acts on $\boldsymbol{B}_i\boldsymbol{A}\boldsymbol{h}$, while projection diversity acts on $\boldsymbol{A}$. We next describe how these regularizers address output redundancy and shared-feature collapse.
\subsection{Specialization via Output Orthogonality}
\label{sub:oo}

Theorem~\ref{thm:routing-recovery} connects expert-output correlations to the distinguishability of routing weights. For fixed expert outputs, the Hessian of the quadratic residual objective is $c^2\boldsymbol{Z}(\boldsymbol{h})^\top\boldsymbol{Z}(\boldsymbol{h})$. Under the theorem's output-norm assumptions, sufficiently small correlations provide a positive curvature bound and control routing-weight error through excess residual loss. 

Corollary~\ref{cor:topk} extends the output-separation bound to selective routing. For a fixed input activation, two alternative top-$K$ routing assignments can select different experts, with at most $2K$ experts in the union of their supports. Their mixture-output difference therefore depends on correlations among experts in that union. The all-pairs squared-cosine penalty controls the corresponding correlation quantity through $\delta_{2K}(\boldsymbol{h})^2 \leq N(N-1)\omega(\boldsymbol{h})$. This motivates penalizing output alignment across all expert pairs, including pairs outside the current selection. Specifically, recalling  $\boldsymbol{z}_i(\boldsymbol{h})=\boldsymbol{B}_i\boldsymbol{A}\boldsymbol{h}$, we define the output-orthogonality loss as the average squared cosine similarity among expert responses:
\begin{equation}
\mathcal{L}_{\mathrm{orth}}
=
\mathbb{E}_{h}
\left[
\frac{1}{N(N-1)}
\sum_{i \neq j}
\left(
\frac{\boldsymbol{z}_i(\boldsymbol{h})^\top \boldsymbol{z}_j({\boldsymbol{h}})}
{\|\boldsymbol{z}_i(\boldsymbol{h})\|_2 \|\boldsymbol{z}_j({\boldsymbol{h}})\|_2}
\right)^2
\right].
\end{equation}
For nonzero expert outputs, normalization makes the penalty invariant to rescaling each output by a nonzero scalar, while squaring penalizes both positive and negative alignment. Evaluating the penalty on activation targets correlations between expert responses to observed inputs. It encourages output decorrelation, while the output-norm lower bound and the pointwise correlation conditions remain assumptions of Theorem~\ref{thm:routing-recovery}. The theorem motivates this regularization target, and we assess task-specific specialization empirically.
\subsection{Collaboration via Shared-Projection Diversity}
Output decorrelation addresses similarities between expert responses, but those responses also depend on the shared features $\boldsymbol{A}\boldsymbol{h}$. Differences between reconstruction matrices can remain unexpressed if the shared features contain little energy in the corresponding latent directions. Theorem~\ref{thm:shared-diversity} shows that controlling the projection's scale and determinant preserves these differences in mean-square output, provided the input distribution has sufficient variation in the retained directions. 

Motivated by this result, we encourage projection diversity through a sampled log-determinant penalty. In LoRA module $m$, for $\boldsymbol{A}\in\mathbb{R}^{r\times k}$ with $r\leq k$, sample a set $\mathcal{S}$ of $r$ columns uniformly. Let $\boldsymbol{A}_{\mathcal{S}}\in\mathbb{R}^{r\times r}$ denote the submatrix formed by these columns. We define
\begin{equation}
\mathcal{L}_{\mathrm{div}}
=
-
\log \det
\left(
\boldsymbol{A}_{\mathcal{S}}^\top
\boldsymbol{A}_{\mathcal{S}}
+
\epsilon \boldsymbol{I}
\right),\qquad \epsilon > 0
\label{eq:loss_div}
\end{equation}
where $\boldsymbol{I}$ is $r\times r$ identity matrix, serving as a stabilization term keeping the logarithm well-defined even when the sampled columns are linearly dependent. The determinant measures the squared volume spanned by the sampled columns before stabilization. At a controlled scale, larger volume discourages nearly dependent directions. The scale assumption is essential to this interpretation. Rescaling $\boldsymbol{A}\mapsto s\boldsymbol{A}$ and $\boldsymbol{B}_i\mapsto\boldsymbol{B}_i/s$  for any scalar $s>1$ preserves expert outputs but can decrease the diversity penalty. Consequently, the penalty alone does not enforce the assumptions of Theorem~\ref{thm:shared-diversity} or guarantee good conditioning. The theorem provides conditional motivation for preserving shared directions through which existing expert differences can affect the output.

We note that this objective is closely related to determinantal point processes (DPPs)~\citep{derezinski2021determinantal}, which favor diverse and non-redundant subsets by maximizing the determinant of a kernel matrix. In our setting, the determinant of $\boldsymbol{A}_{\mathcal{S}}^\top\boldsymbol{A}_{\mathcal{S}}$ equals the squared volume spanned by the sampled columns. We use this quantity as a regularization objective to encourage diversity in the shared projection. We also note that our approach is aligned with prior research~\citep{huang2025hira,albert2025randlora,shen2025kron,xu2026low}.

\subsection{Overall Training Objective}
We optimize OD-MoLE by augmenting the task loss with two complementary regularization terms that promote expert specialization and collaborative expressiveness. Let $\mathcal{D} = \{(\boldsymbol{x}, \boldsymbol{y})\}$ denote the training dataset. We use $m$ to index all LoRA MoE modules $\mathcal{M}$ across layers and submodules (e.g., query, key, value, and output projections), and let $\boldsymbol{h}_m(\boldsymbol{x})$ denote the input hidden representation to module $m$ induced by $\boldsymbol{x}$. We define the output-space orthogonality loss: 
\begin{equation}
\mathcal{L}_{\mathrm{orth}}(\boldsymbol{x})
=
\frac{1}{|\mathcal{M}|}\sum_{m\in\mathcal{M}}
\frac{1}{N(N-1)}
\sum_{i \neq j}
\left(
\frac{
\left\langle
\boldsymbol{B}_{m,i}\boldsymbol{A}_{m}\boldsymbol{h}_{m}(\boldsymbol{x}),
\boldsymbol{B}_{m,j}\boldsymbol{A}_{m}\boldsymbol{h}_{m}(\boldsymbol{x})
\right\rangle
}{
\left\|\boldsymbol{B}_{m,i}\boldsymbol{A}_{m}\boldsymbol{h}_{m}(\boldsymbol{x})\right\|_2
\left\|\boldsymbol{B}_{m,j}\boldsymbol{A}_{m}\boldsymbol{h}_{m}(\boldsymbol{x})\right\|_2
}
\right)^2,
\label{eq:loss_ortho}
\end{equation}
which penalizes squared cosine similarity between expert outputs. OD-MoLE training objective is:
\begin{equation}
\begin{aligned}
\mathcal{L}_{\mathrm{OD\text{-}MoLE}}&=\mathbb{E}_{(\boldsymbol{x},\boldsymbol{y})\sim\mathcal{D}}\Big[\ell\!\left(f_{\boldsymbol{W}_0 + \Delta \boldsymbol{W}_{\mathrm{MoE}}}(\boldsymbol{x}), \boldsymbol{y}\right)+\lambda_{\mathrm{orth}} \mathcal{L}_{\mathrm{orth}}(\boldsymbol{x})\Big]-\frac{\lambda_{\mathrm{div}}}{|\mathcal{M}|}\sum_{m\in\mathcal{M}}
\mathbb{E}_{\mathcal{S}_m}\left[\log\det\!\left(\boldsymbol{A}_{m,\mathcal{S}_m}^\top\boldsymbol{A}_{m,\mathcal{S}_m}+\epsilon\boldsymbol{I}\right)\right].
\label{eq:overall-objective}
\end{aligned}
\end{equation}
Here, $\ell(\cdot)$ denotes the task loss, while the two regularizers encourage less redundant expert outputs and more diverse shared projection directions. The coefficients $\lambda_{\mathrm{orth}}$ and $\lambda_{\mathrm{div}}$ control their contributions to training. Their effects on task performance are evaluated empirically in Tab.~\ref{tab:regularizer-ablation-selective}.

\paragraph{Efficient approximation.}Direct computation of the regularization terms over all tokens and modules can be expensive in Eq.~\ref{eq:overall-objective}. In practice, we adopt stochastic approximations: the orthogonality loss is estimated using a randomly sampled subset of token-level activations, while the diversity regularizer is computed on a randomly selected LoRA module $m^{*}$ and a subset $\mathcal{S}$ of its columns, then compute $\mathcal{L}_{\mathrm{div}}$ via Eq.~\ref{eq:loss_div} at each training step. These sampling procedures reduce regularization overhead. The completed algorithm is in Algorithm~\ref{alg:odmole}.

\begin{algorithm}[t]
\caption{OD-MoLE}
\label{alg:odmole}
\begin{algorithmic}[1]
\Require Dataset $\mathcal{D}$, frozen $\boldsymbol{W}_0$,
LoRA modules $\mathcal{M}$, $N$ experts, top-$K$ routing,
$\lambda_{\mathrm{orth}},\lambda_{\mathrm{div}}$
\State Initialize $\{\boldsymbol{A}_m,\boldsymbol{B}_{m,i}\}$, routers $\phi_m$ for each module $m\in\mathcal{M}$
\For{each minibatch $(\boldsymbol{x},\boldsymbol{y})\sim\mathcal{D}$}
    \State For each module $m$, $\boldsymbol{g}_m\gets\text{TopKSoftmax}(\phi_m(\boldsymbol{h}_m),K)$
    \State Compute $\boldsymbol{z}_{m,i}=
    \boldsymbol{B}_{m,i}\boldsymbol{A}_m \boldsymbol{h}_m$ for each expert $i$, and routed residual
        $c\sum_i g_{m,i}\boldsymbol{z}_{m,i}$
    \State Randomly select a module $m^{*}$ and columns $\mathcal{S}_{m^{*}}$ of $\boldsymbol{A}_{m^{*}}$ and compute $\mathcal{L}_{\mathrm{div}}$ via Eq.~\ref{eq:loss_div}
    \State Compute output-space orthogonality loss $\mathcal{L}_{\mathrm{orth}}$ via Eq.~\ref{eq:loss_ortho}
    \State Compute $\mathcal{L}=\mathcal{L}_{\mathrm{task}}+\lambda_{\mathrm{orth}}\mathcal{L}_{\mathrm{orth}}+\lambda_{\mathrm{div}}\mathcal{L}_{\mathrm{div}}$
    \State Update $\{\boldsymbol{A}_m,\boldsymbol{B}_{m,i},\phi_m\}$
    by backpropagation while keeping $\boldsymbol{W}_0$ frozen
\EndFor
\end{algorithmic}
\end{algorithm}

\begin{table}[t]
  \caption{Comparison of PEFT methods on commonsense reasoning tasks using Llama-2-7B. LoRA MoE methods are grouped into (i) all-expert composition and (ii) selective routing, including both adaptive routing (AdaMoLE) and top-$k$ selection. Task scores are mean accuracies (\%) across three seeds $\{0,1,2\}$. Avg.\ reports the mean $\pm$ sample standard deviation of the per-seed average accuracy across eight tasks. \textbf{Bold} and \underline{underlined} indicate the best and second-best results within each group.}
  \label{tab:main_la7B_commonsense_reason}
  \centering
  \footnotesize
  \begin{tabularx}{\linewidth}{l|*{8}{>{\centering\arraybackslash}X}c}
    \toprule
    Method & ComQA & CosQA & SIQA & PIQA & ScQA & BoolQ & \mbox{ARC-C} & OBQA & Avg \\
    \midrule
    \multicolumn{10}{l}{\textit{Single-LoRA baselines}} \\
    \midrule
    LoRA ($r$:$8$)  & 73.31 & 81.71 & 78.33 & 82.97 & 88.15 & 76.80 & 58.45 & 83.60 & $77.92\pm1.96$ \\
    LoRA ($r$:$16$) & 73.23 & 81.68 & 77.97 & 83.11 & 88.85 & 77.40 & 59.25 & 84.80 & $78.29\pm1.97$ \\
    LoRA ($r$:$32$) & 73.06 & 82.78 & 78.04 & 83.23 & 88.46 & 75.20 & 58.14 & 85.00 & $77.99\pm1.91$ \\
    PiSSA           & 72.60 & 81.15 & 77.45 & 81.85 & 87.18 & 74.80 & 57.97 & 81.40 & $76.80\pm1.32$ \\

    \midrule
    \multicolumn{10}{l}{\textit{Compose all experts}} \\
    \midrule
    HydraLoRA & \underline{75.76} & 82.78 & \underline{78.25} & 82.92 & 88.62 & 80.40 & 58.70 & 86.60 & $79.25\pm1.31$ \\
    MoLoRA    & 74.86 & \textbf{84.19} & \textbf{78.66} & \underline{83.35} & \textbf{89.32} & 78.60 & \underline{59.98} & \underline{87.20} & $\underline{79.52}\pm1.35$ \\
    GOAT      & 73.96 & 82.74 & 77.83 & \textbf{84.01} & 87.37 & \textbf{82.20} & 59.53 & 86.20 & $79.23\pm1.56$ \\
    OD-MoLE   & \textbf{76.41} & \underline{83.45} & 78.20 & 83.03 & \underline{89.18} & \underline{81.21} & \textbf{61.26} & \textbf{87.60} & $\textbf{80.04}\pm1.19$\\

    \midrule
    \multicolumn{10}{l}{\textit{Route to selective experts}} \\
    \midrule
    HydraLoRA & \underline{78.21} & \underline{84.05} & \textbf{78.86} & 82.56 & 90.11 & 82.20 & 61.69 & \textbf{89.40} & $\underline{80.89}\pm1.43$ \\
    MoLoRA    & 77.31 & 84.02 & 77.99 & 81.18 & \underline{90.16} & 81.60 & 59.64 & 87.20 & $79.89\pm1.38$ \\
    AdaMoLE   & 74.46 & 83.08 & 78.13 & \textbf{84.12} & 89.03 & 80.80 & 60.87 & 86.40 & $79.61\pm1.22$ \\
    GOAT      & 76.63 & 83.75 & 78.36 & 81.47 & \textbf{90.32} & \underline{82.80} & \underline{61.88} & 88.40 & $80.45\pm1.67$ \\
    OD-MoLE   & \textbf{78.79} & \textbf{84.25} & \underline{78.51} & \underline{83.03} & 89.74 & \textbf{83.60} & \textbf{62.37} & \underline{88.80} & $\textbf{81.14}\pm1.07$ \\
    \bottomrule
  \end{tabularx}
\end{table}
\section{Experiments}
We first describe the experimental setup, followed by the main results, then present ablation studies and additional analyses to provide deeper insights into our method. Additional experimental results are shown in Appendix~\ref{app:experiment-details}.

\paragraph{Baselines.}We evaluate OD-MoLE against both single-LoRA fine-tuning and existing LoRA MoE approaches. For single-LoRA baselines, we consider LoRA~\citep{hu2022lora} with ranks $r \in \{8,16,32\}$ and PiSSA~\citep{meng2024pissa}. For LoRA MoE methods, we compare with HydraLoRA~\citep{tian2024hydralora}, GOAT~\citep{fan2025make}, AdaMoLE~\citep{liu2024adamole}, and MoLoRA~\citep{zadouri2024pushing}.

\paragraph{Model and Datasets.}We conduct experiments on two pre-trained models, Llama-2-7B and Qwen2.5-7B~\citep{yang2024qwen2}. We consider a diverse set of commonsense reasoning benchmarks, including CommonsenseQA (ComQA)~\citep{talmor-etal-2019-commonsenseqa}, CosmosQA (CosQA)~\citep{huang-etal-2019-cosmos}, SocialIQA (SIQA)~\citep{sap-etal-2019-social}, PhysicalIQA (PIQA)~\citep{bisk2020piqa}, ScienceQA (ScQA)~\citep{lu2022learn}, BoolQ~\citep{clark-etal-2019-boolq}, ARC-Challenge (ARC-C)~\citep{clark2018think}, and OpenBookQA (OBQA)~\citep{mihaylov-etal-2018-suit}.

\paragraph{Experimental Setup.}For all LoRA MoE methods, we assign LoRA experts to four attention projection submodules ($W_q$, $W_k$, $W_v$, $W_o$) in every Transformer layer and fix the total effective expert rank at 32, excluding gating-related parameters. We consider two configurations: (i) \textit{expert composition}, with 2 experts per submodule, all active; and (ii) \textit{selective routing}, with 8 experts per submodule and top-$2$ gating. Single-LoRA baselines use ranks $r\in\{8,16,32\}$. Following~\citep{liu2024adamole}, we truncate input sequences to a maximum length of 256 tokens. We use a total batch size of 16 and a learning rate of $10^{-4}$, with $\lambda_{\mathrm{orth}}=\lambda_{\mathrm{div}}=10^{-3}$. The number of training epochs for each dataset is selected based on validation performance and is reported in Appendix Tab.~\ref{tab:training_epochs}. To promote balanced expert utilization, we include a load-balancing loss~\citep{fedus2022switch} with coefficient $10^{-3}$. All experiments use Hugging Face Transformers~\citep{wolf2019huggingface} and PEFT on a single NVIDIA A100 GPU.

\subsection{Overall Performance}

\begin{table}[t]
\caption{Comparison of OD-MoLE and HydraLoRA on commonsense reasoning tasks using Qwen2.5-7B in all-expert composition and selective top-$k$ routing. Task scores are mean accuracies (\%) across seeds $\{0,1,2\}$. Avg.\ reports mean $\pm$ sample standard deviation of per-seed average accuracy across eight tasks. \textbf{Bold} indicates the highest mean average within each routing group.}
\label{tab:main_Qwen7B_commonsense_reason}
\centering
\footnotesize
\setlength{\tabcolsep}{3pt}
\begin{tabularx}{\linewidth}{l|*{8}{>{\centering\arraybackslash}X}c}
\toprule
Method & ComQA & CosQA & SIQA & PIQA & ScQA & BoolQ
& \mbox{ARC-C} & OBQA & Avg. \\
\midrule
\multicolumn{10}{l}{\textit{Compose all experts}} \\
\midrule
HydraLoRA
& 85.42 & 84.35 & 80.14 & 87.65 & 92.49 & 83.20 & 83.36 & 92.80
& $86.18{\pm}1.25$ \\
OD-MoLE
& 85.50 & 84.85 & 79.63 & 87.65 & 92.58 & 83.80 & 83.96 & 94.80
& $\mathbf{86.60}{\pm}0.92$ \\
\midrule
\multicolumn{10}{l}{\textit{Route to selective experts}} \\
\midrule
HydraLoRA
& 85.01 & 84.57 & 81.24 & 87.25 & 93.32 & 84.10 & 84.73 & 93.20
& $86.68{\pm}1.37$ \\
OD-MoLE
& 85.09 & 84.93 & 81.35 & 87.84 & 93.11 & 84.50 & 84.73 & 94.80
& $\mathbf{87.04}{\pm}0.95$ \\
\bottomrule
\end{tabularx}
\end{table}

Tab.~\ref{tab:main_la7B_commonsense_reason} reports the performance of OD-MoLE and baseline methods on commonsense reasoning benchmarks using Llama-2-7B. We group LoRA MoE methods into (i) expert composition and (ii) selective routing to enable a systematic comparison under different routing strategies. Under all expert compositions, OD-MoLE outperforms baselines on average performance, attaining the best results on ComQA, ARC-C, and OBQA, and remains competitive on other tasks. Under selective routing, OD-MoLE further improves the average performance, outperforming HydraLoRA, GOAT, and AdaMoLE. It achieves the best results on ComQA, CosQA, BoolQ, and ARC-C, and ranks among top performers on most other tasks. Tab.~\ref{tab:main_Qwen7B_commonsense_reason} compares OD-MoLE with HydraLoRA on Qwen2.5-7B. OD-MoLE outperforms HydraLoRA under both expert composition and top-$k$ routing, achieving higher average accuracy in both settings. The improvements are consistent across most tasks, with notable gains on CosQA, BoolQ, and OBQA. These results demonstrate that enforcing output-level orthogonality and diversity leads to more effective and robust expert representations.

\subsection{Ablation Study}

\begin{figure}[t]
  \centering
  \footnotesize

  \begin{subfigure}{0.24\linewidth}
    \centering
    \includegraphics[width=\linewidth]{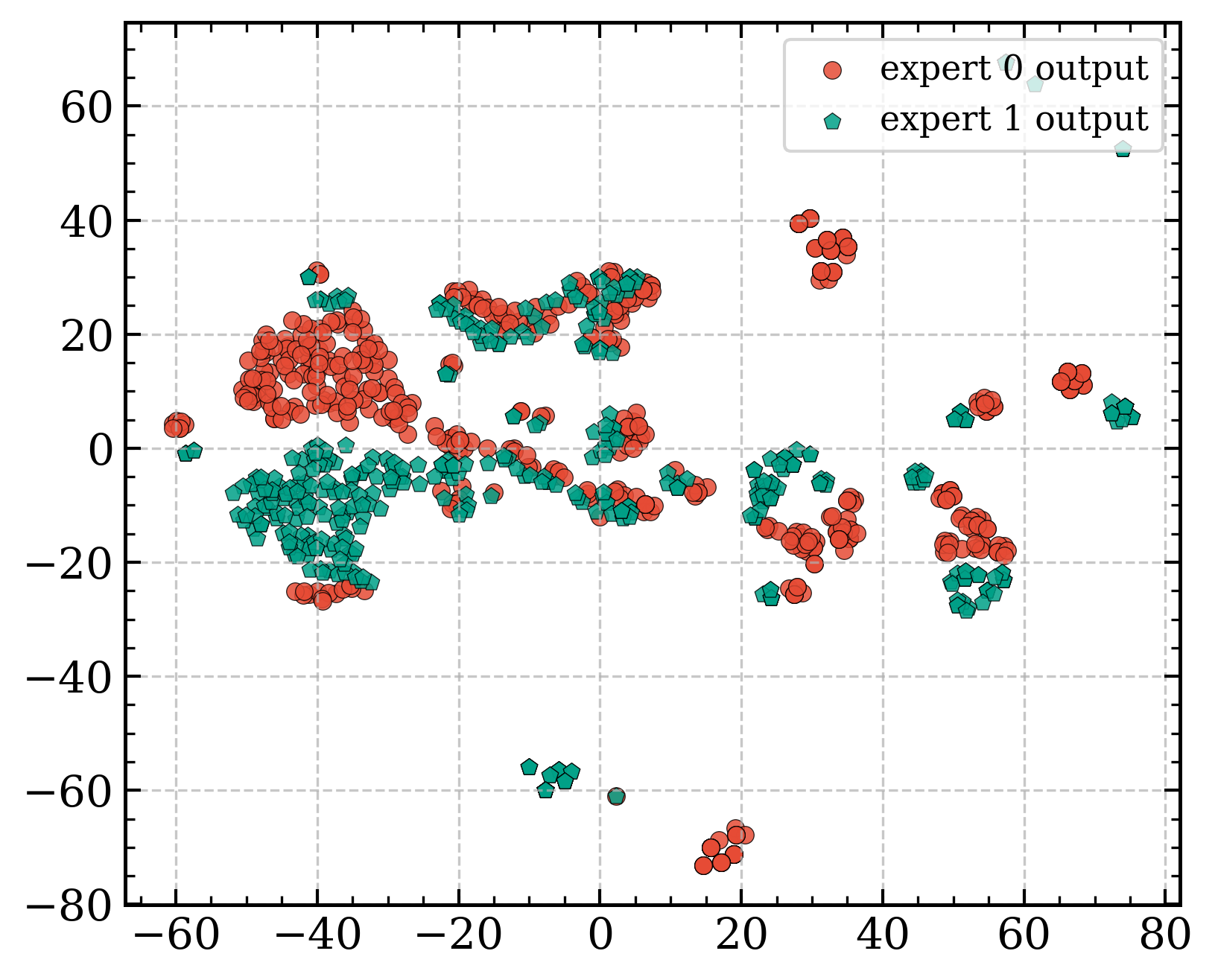}
    \caption*{HydraLoRA ($q$-proj)}
  \end{subfigure}
  \hfill
  \begin{subfigure}{0.24\linewidth}
    \centering
    \includegraphics[width=\linewidth]{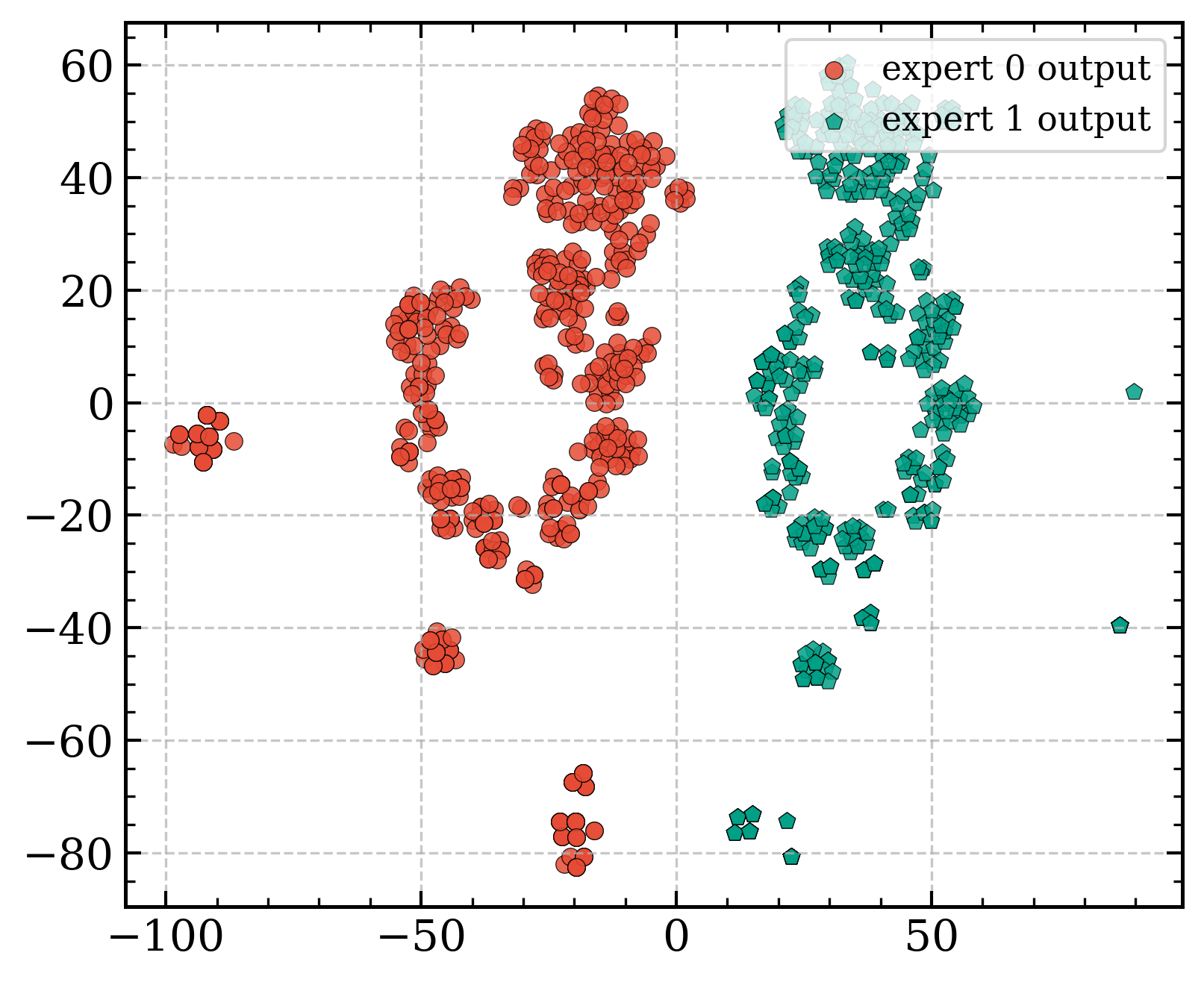}
    \caption*{OD-MoLE ($q$-proj)}
  \end{subfigure}
  \hfill
  \begin{subfigure}{0.24\linewidth}
    \centering
    \includegraphics[width=\linewidth]{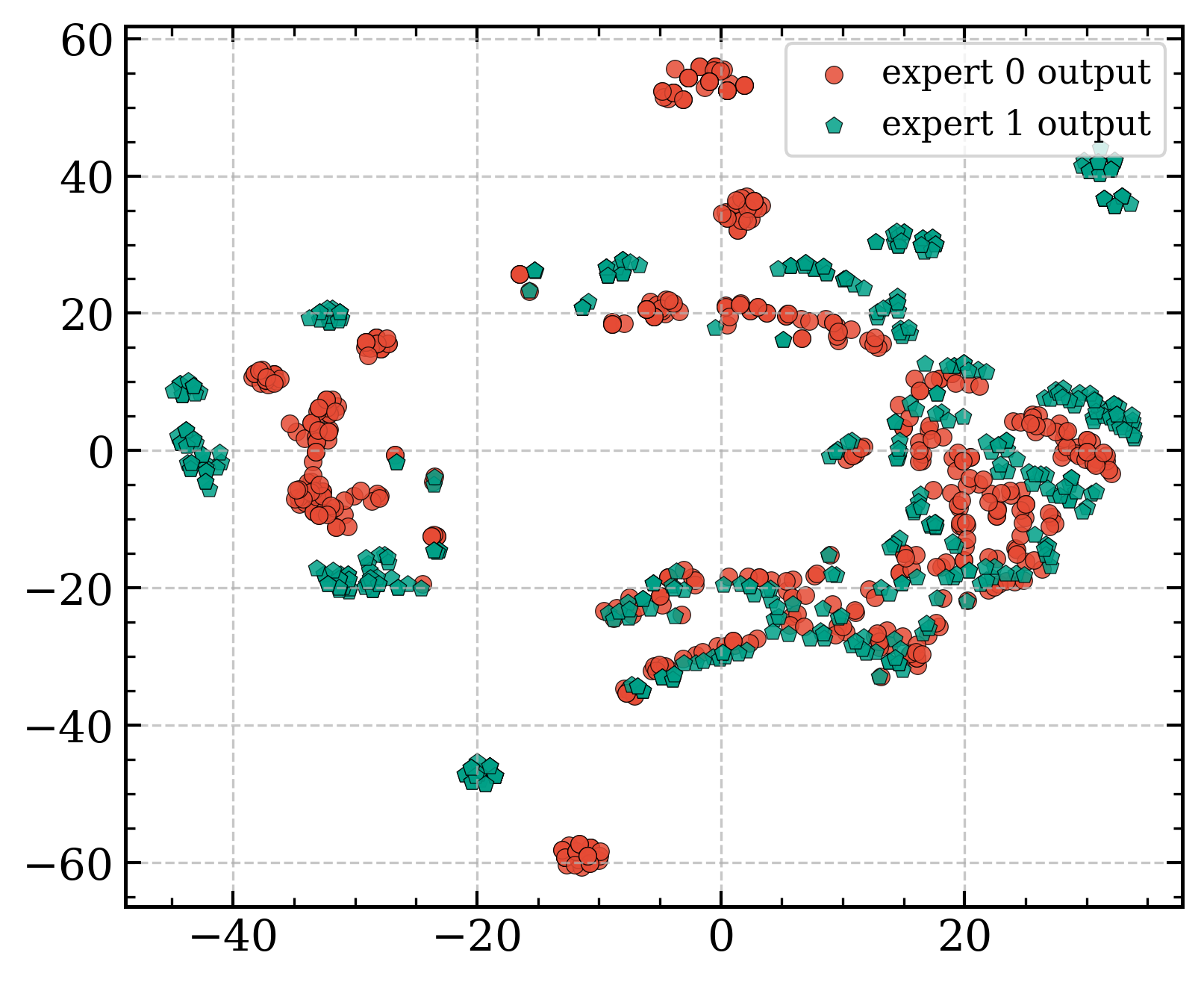}
    \caption*{HydraLoRA ($k$-proj)}
  \end{subfigure}
  \hfill
  \begin{subfigure}{0.24\linewidth}
    \centering
    \includegraphics[width=\linewidth]{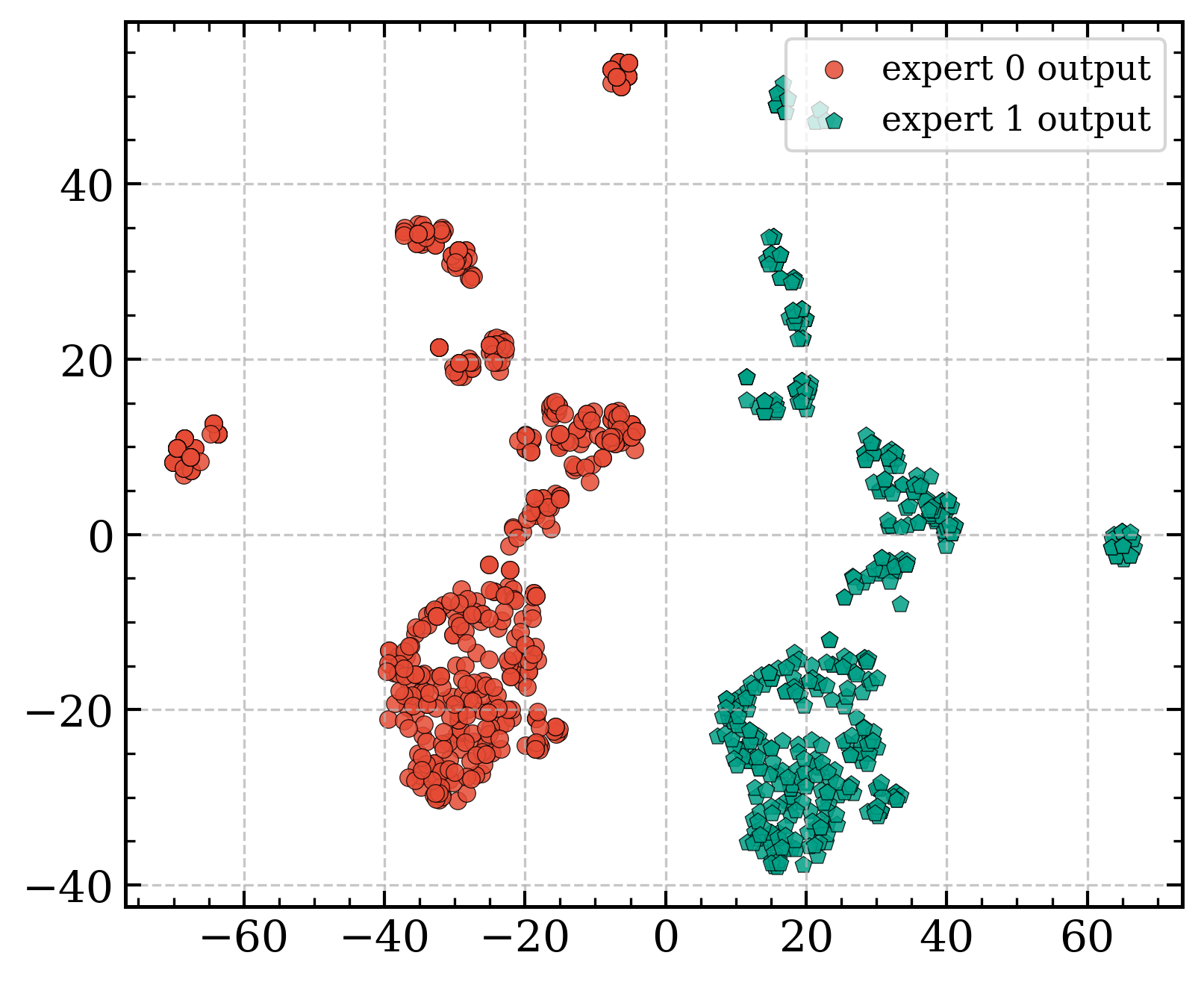}
    \caption*{OD-MoLE ($k$-proj)}
  \end{subfigure}

  \caption{
  t-SNE visualization of expert output representations $\boldsymbol{B}_i\boldsymbol{A}\boldsymbol{h}$ for $q$- and $k$-projection modules at layer 0. From left to right: HydraLoRA ($q$), OD-MoLE ($q$),  HydraLoRA ($k$), and OD-MoLE ($k$). OD-MoLE exhibits clearer separation between expert outputs, improving specialization.
  }
  \label{fig:tsne_hydra_odmole_layer0_qk}\vspace{-0.2cm}
\end{figure}

\paragraph{Expert Output Representation Analysis.}~To better understand how different methods affect expert specialization, we visualize expert output representations $\boldsymbol{B}_i\boldsymbol{A}\boldsymbol{h}$ using t-SNE~\citep{van2008visualizing}. Fig.~\ref{fig:tsne_hydra_odmole_layer0_qk} compares HydraLoRA and OD-MoLE on the $q$- and $k$-projection modules at layer 0 under expert composition, using Llama-2-7B on ARC-C. We observe that HydraLoRA produces heavily overlapping clusters between expert outputs, indicating that different experts tend to learn similar transformations and exhibit limited specialization. In contrast, OD-MoLE yields clearly separated clusters for different experts across both $q$- and $k$-projection modules. It suggests that OD-MoLE encourages experts to capture more diverse and distinct output representations.

\paragraph{Impact of Each Regularizer.}~
Tab.~\ref{tab:regularizer-ablation-selective} compares output orthogonality and shared-projection diversity under selective routing with top-2 routing over 8 experts. Each regularizer improves macro-average over HydraLoRA in all three seeds. The mean gains are $0.51$ percentage points (pp) for only orthogonality and $0.45$ pp for only diversity. Combining both regularizers achieves the highest macro-average, a gain of $0.63$ pp over HydraLoRA, and the highest mean accuracy on 5 of 8 tasks. We also test a setting with diversity disabled, where computing orthogonality over selected pairs or all pairs yields nearly identical macro-averages, and it shows that computing every pair of experts, including those experts that are not selected by the router, does not get in the way of what each expert is trying to learn, since we make routed output spaces orthogonal but do not destroy their learning space. Appendix~\ref{app:ablation-per-seed} (Tabs.~\ref{tab:orth-only-seeds}--\ref{tab:pairs-seeds}) reports the per-seed results and paired comparisons underlying Tab.~\ref{tab:regularizer-ablation-selective}.

\begin{table}[t]
\caption{Regularizer ablation on Qwen2.5-7B in selective routing with top-2 over 8 experts. Orth. and Div. denote output orthogonality and shared-projection diversity. $\checkmark^{\dagger}$ computes the orthogonality loss over the router-selected expert pair only. Task scores are mean accuracies (\%) with standard deviation as subscript across seeds $\{40,41,42\}$. Avg.\ is mean and standard deviation of per-seed average accuracy across eight tasks. Bold marks the highest mean in each column.}
\label{tab:regularizer-ablation-selective}
\centering
\small
\resizebox{\linewidth}{!}{%
\begin{tabular}{cc*{9}{c}}
\toprule
Orth. & Div. & ComQA & CosQA & SIQA & PIQA
& ScQA & BoolQ & ARC-C & OBQA & Avg. \\
\midrule
$\times$ & $\times$
& $84.58_{\pm0.21}$ & $85.72_{\pm1.70}$ & $79.38_{\pm0.14}$ & $89.28_{\pm0.76}$
& $91.75_{\pm0.52}$ & $87.35_{\pm0.26}$ & $88.25_{\pm0.42}$ & $95.20_{\pm0.20}$ & $87.69_{\pm0.28}$ \\

$\times$ & $\checkmark$
& $84.63_{\pm0.54}$ & $87.16_{\pm0.54}$ & $\textbf{79.87}_{\pm0.39}$ & $89.63_{\pm0.77}$
& $92.21_{\pm0.54}$ & $87.68_{\pm0.38}$ & $88.60_{\pm0.18}$ & $95.33_{\pm0.23}$ & $88.14_{\pm0.10}$ \\

$\checkmark^{\dagger}$ & $\times$
& $84.50_{\pm0.60}$ & $\textbf{87.53}_{\pm1.08}$ & $79.58_{\pm0.53}$ & $89.30_{\pm0.56}$
& $92.52_{\pm0.40}$ & $87.97_{\pm0.64}$ & $88.31_{\pm0.76}$ & $95.80_{\pm0.40}$ & $88.19_{\pm0.19}$ \\

$\checkmark$ & $\times$
& $84.77_{\pm0.64}$ & $86.65_{\pm0.40}$ & $79.75_{\pm0.30}$ & $89.59_{\pm0.27}$
& $92.37_{\pm0.03}$ & $88.12_{\pm0.10}$ & $88.05_{\pm0.23}$ & $\textbf{96.27}_{\pm0.46}$ & $88.20_{\pm0.10}$ \\

$\checkmark$ & $\checkmark$
& $\textbf{84.85}_{\pm0.29}$ & $86.43_{\pm1.20}$ & $79.75_{\pm0.58}$ & $\textbf{89.90}_{\pm0.52}$
& $\textbf{92.90}_{\pm0.68}$ & $\textbf{88.21}_{\pm0.29}$ & $\textbf{88.68}_{\pm0.27}$
& $95.87_{\pm0.50}$ & $\textbf{88.32}_{\pm0.26}$ \\
\bottomrule
\end{tabular}%
}
\end{table}

\paragraph{Mechanisms of the Orthogonality and Diversity Regularizers.}~To test local routing distinguishability, we hold the two selected experts fixed at each evaluated token and attention projection and transfer $0.05$ routing weight from the higher-weight expert to its partner. We measure the resulting change in the mixture output $\boldsymbol{r}$ using $S=\lVert\Delta \boldsymbol{r}\rVert_2/\lVert\Delta \boldsymbol{g}\rVert_2$, where $\Delta \boldsymbol{g}$ is the routing-weight perturbation. With diversity disabled, output orthogonality increases median $S$ by $0.5$--$10.4\%$ and decreases mean squared cosine similarity by $4.9$--$17.3\%$ across the eight tasks, shown in Tab.~\ref{tab:orth-mechanism}. The fifth percentile of $S$ increases on five tasks and decreases on three. The sensitivity results show that the same weight transfer typically produces a larger change in the mixture output. The cosine results show that the selected expert outputs are less aligned on average. Appendix~\ref{app:orth-analysis} provides additional analyses
of routing sensitivity, expert-output norms, routing distributions, and predictive contributions under expert interventions.

For shared-projection diversity, we compare only applying diversity loss with HydraLoRA on PIQA under all-expert composition in Tab.~\ref{tab:div-mechanism}. Effective rank and stable rank decline in both settings. Only applying diversity loss reduces the decline, retaining higher mean effective rank and stable rank at the end epoch. Both the median and p90 condition numbers are about $10\%$ lower with only applying diversity loss at the final epoch. These metrics are invariant to uniform rescaling, so the observed differences reflect changes in the relative singular values of $\boldsymbol{A}$. Appendix~\ref{app:div-analysis} reports additional spectral diagnostics and comparisons of raw and scale-normalized log-determinant regularization.

\begin{table}[t]
\centering
\caption{Routing distinguishability under selective routing with top-2 routing over eight experts using $\lambda_{\mathrm{div}}=0$. Orth. indicates whether the output orthogonality penalty is enabled. $S$ measures the response to a $0.05$ routing-weight transfer within the selected pair. The cosine rows report 100 times the mean squared cosine similarity between the selected expert outputs.}
\label{tab:orth-mechanism}
\footnotesize
\begin{tabular}{lc*{8}{c}}
\toprule
Metric & Orth. & ComQA & CosQA & SIQA & PIQA
& ScQA & BoolQ & ARC-C & OBQA \\
\midrule
\multirow{2}{*}{Median $S\uparrow$}
& $\times$ & 0.610 & 3.444 & 1.758 & 2.578 & 3.230 & 1.393 & 1.396 & 1.786 \\
& $\checkmark$ & \textbf{0.673} & \textbf{3.560} & \textbf{1.802} & \textbf{2.705}
& \textbf{3.246} & \textbf{1.463} & \textbf{1.412} & \textbf{1.805} \\
\midrule
\multirow{2}{*}{$100\,\overline{\cos^2}\downarrow$}
& $\times$ & 4.57 & 1.95 & 3.74 & 2.92 & 2.08 & 2.45 & 4.75 & 3.61 \\
& $\checkmark$ & \textbf{3.78} & \textbf{1.64} & \textbf{3.21} & \textbf{2.42}
& \textbf{1.93} & \textbf{2.33} & \textbf{4.30} & \textbf{3.17} \\
\bottomrule
\end{tabular}
\end{table}

\begin{table}[t]
\centering
\caption{Shared-projection spectrum and conditioning on PIQA under all-expert composition. Orth. and Div. denote output orthogonality and shared-projection diversity ($\lambda_{\mathrm{div}}=0.001$). Init.\ is the shared initialization of both. The other rows are measured at the end of training. Effective and stable ranks are means over 112 shared projection matrices. Median and p90 summarize the per-matrix spectral condition number $\kappa_2(\boldsymbol{A})$.}
\label{tab:div-mechanism}
\footnotesize
\begin{tabular}{cccccc}
\toprule
Orth. & Div. & Eff. rank & Stable rank & Median $\kappa_2$ & p90 $\kappa_2$ \\
\midrule
\multicolumn{2}{c}{Init.} & 15.96 & 14.34 & 1.12 & 1.13 \\
\midrule
$\times$ & $\times$ & 12.19 & 4.21 & 2.83 & 3.19 \\
$\times$ & $\checkmark$
& \textbf{12.92} & \textbf{4.63} & \textbf{2.54} & \textbf{2.87} \\
\bottomrule
\end{tabular}
\end{table}

\paragraph{Effect of the Number of Experts.}We vary $N\in\{8,16,32\}$ while fixing the total expert rank $Nr_i=32$ and routing ratio $K/N=1/4$, keeping the active expert rank at $Kr_i=8$, where $r_i$ is rank of each expert $i$. Using Qwen2.5-7B, OD-MoLE achieves average accuracies of 87.04\%, 86.23\%, and 86.53\%, exceeding HydraLoRA by 0.36, 0.18, and 0.54 percentage points, respectively. Increasing expert count while reducing per-expert rank therefore does not consistently improve performance. Full results are reported in
Appendix Tab.~\ref{tab:expert-count}.

\begin{table}[h!]
\centering
\footnotesize
\caption{Comparison of OD-MoLE and HydraLoRA in peak memory usage, training time per epoch, and performance on CommonsenseQA using Llama-2-7B and Qwen2.5-7B. Both methods are evaluated under identical settings on a single NVIDIA A100 GPU.}
\label{tab:computation_cost}

\begin{tabularx}{\linewidth}{l*{6}{>{\centering\arraybackslash}X}}
\toprule
 & \multicolumn{3}{c}{Llama-2-7B} & \multicolumn{3}{c}{Qwen2.5-7B} \\
\cmidrule(lr){2-4} \cmidrule(lr){5-7}
Method & Peak Mem & Time/Epoch & Accuracy(\%) 
       & Peak Mem & Time/Epoch & Accuracy(\%)\\
\midrule

\multicolumn{7}{l}{\textit{Compose all experts}} \\
HydraLoRA &37.30GB & 34.0min& 75.76 & 33.07GB & 37.0min & 85.42 \\
OD-MoLE   &37.18GB & 40.0min & \textbf{76.41} &  34.77GB&42.5min & \textbf{85.50} \\

\midrule
\multicolumn{7}{l}{\textit{Route to selective experts}} \\
HydraLoRA & 39.10GB & 48.5min & 78.21 & 35.35GB & 52.5min & 85.01 \\
OD-MoLE   & 39.33GB & 56.5min & \textbf{78.79} & 36.85GB & 59.0min & \textbf{85.09} \\
\bottomrule
\end{tabularx}\vspace{-0.1cm}
\end{table}
\paragraph{Computation Analysis.}Tab.~\ref{tab:computation_cost} compares the computational cost and performance of different LoRA MoE methods under identical experimental settings. We evaluate both peak memory usage and per-epoch training time on a single NVIDIA A100 GPU. OD-MoLE achieves consistently better performance than HydraLoRA across both routing strategies, with only a modest increase in computational cost. In \textit{compose-all-experts} setting, OD-MoLE improves accuracy while incurring a negligible increase in memory and a small increase in training time. In \textit{selective routing} setting, OD-MoLE again outperforms HydraLoRA, but with a more noticeable increase in training time. We attribute this overhead to the additional computations required for enforcing output diversity and orthogonality among experts. Despite this increase, the memory usage remains comparable, indicating that our method does not introduce significant additional parameters. These results highlight a favorable trade-off: OD-MoLE consistently improves model performance while maintaining similar memory consumption and only moderately increasing training time.

\section{Conclusion}
In this work, we have presented OD-MoLE, which combines expert-output orthogonality and shared-projection diversity in an asymmetric LoRA mixture. Our analysis establishes sufficient conditions under which output decorrelation makes routing weights distinguishable and shared-projection diversity preserves existing expert differences in mean-square output. These results motivate the two regularization targets while making their assumptions explicit. Experiments on Llama-2-7B and Qwen2.5-7B show OD-MoLE performs well under both all-expert composition and selective routing. Complementarily, the results support the usefulness of the combined regularization, establishing task-specific expert functions requires further evaluation beyond output separation.

\section*{Acknowledgments}
This material is based upon work supported by the National Science Foundation (NSF) under Grant Number 2318101. Any opinions, findings, conclusions, or recommendations expressed in this material are those of the author(s) and do not necessarily reflect the views of the National Science Foundation.

\bibliographystyle{unsrtnat}
\bibliography{references}

@article{shen2025kron,
  title={Kron-LoRA: Hybrid Kronecker-LoRA Adapters for Scalable, Sustainable Fine-tuning},
  author={Shen, Yixin},
  journal={arXiv preprint arXiv:2508.01961},
  year={2025}
}

@article{albert2025randlora,
  title={Randlora: Full-rank parameter-efficient fine-tuning of large models},
  author={Albert, Paul and Zhang, Frederic Z and Saratchandran, Hemanth and Rodriguez-Opazo, Cristian and Hengel, Anton van den and Abbasnejad, Ehsan},
  journal={arXiv preprint arXiv:2502.00987},
  year={2025}
}

@inproceedings{huang2025hira,
  title={HiRA: Parameter-efficient hadamard high-rank adaptation for large language models},
  author={Huang, Qiushi and Ko, Tom and Zhuang, Zhan and Tang, Lilian and Zhang, Yu},
  booktitle={The Thirteenth International Conference on Learning Representations},
  year={2025}
}

@article{derezinski2021determinantal,
  title={Determinantal point processes in randomized numerical linear algebra},
  author={Derezinski, Micha{\l} and Mahoney, Michael W},
  journal={Notices of the American Mathematical Society},
  volume={68},
  number={1},
  pages={34--45},
  year={2021}
}

@inproceedings{
tian2024hydralora,
title={HydraLo{RA}: An Asymmetric Lo{RA} Architecture for Efficient Fine-Tuning},
author={Chunlin Tian and Zhan Shi and Zhijiang Guo and Li Li and Cheng-zhong Xu},
booktitle={The Thirty-eighth Annual Conference on Neural Information Processing Systems},
year={2024},
url={https://openreview.net/forum?id=qEpi8uWX3N}
}

@inproceedings{
fan2025make,
title={Make Lo{RA} Great Again: Boosting Lo{RA} with Adaptive Singular Values and Mixture-of-Experts Optimization Alignment},
author={Chenghao Fan and Zhenyi Lu and Sichen Liu and Chengfeng Gu and Xiaoye Qu and Wei Wei and Yu Cheng},
booktitle={Forty-second International Conference on Machine Learning},
year={2025},
url={https://openreview.net/forum?id=SUxq4HeIAd}
}

@inproceedings{
liu2024adamole,
title={AdaMo{LE}: Fine-Tuning Large Language Models with Adaptive Mixture of Low-Rank Adaptation Experts},
author={Zefang Liu and Jiahua Luo},
booktitle={First Conference on Language Modeling},
year={2024},
url={https://openreview.net/forum?id=ndY9qFf9Sa}
}

@article{fedus2022switch,
  title={Switch transformers: Scaling to trillion parameter models with simple and efficient sparsity},
  author={Fedus, William and Zoph, Barret and Shazeer, Noam},
  journal={Journal of Machine Learning Research},
  volume={23},
  number={120},
  pages={1--39},
  year={2022}
}

@article{wolf2019huggingface,
  title={Huggingface's transformers: State-of-the-art natural language processing},
  author={Wolf, Thomas and Debut, Lysandre and Sanh, Victor and Chaumond, Julien and Delangue, Clement and Moi, Anthony and Cistac, Pierric and Rault, Tim and Louf, R{\'e}mi and Funtowicz, Morgan and others},
  journal={arXiv preprint arXiv:1910.03771},
  year={2019}
}

@inproceedings{
meng2024pissa,
title={Pi{SSA}: Principal Singular Values and Singular Vectors Adaptation of Large Language Models},
author={Fanxu Meng and Zhaohui Wang and Muhan Zhang},
booktitle={The Thirty-eighth Annual Conference on Neural Information Processing Systems},
year={2024},
url={https://openreview.net/forum?id=6ZBHIEtdP4}
}

@inproceedings{liu2024dora,
  title={Dora: Weight-decomposed low-rank adaptation},
  author={Liu, Shih-Yang and Wang, Chien-Yi and Yin, Hongxu and Molchanov, Pavlo and Wang, Yu-Chiang Frank and Cheng, Kwang-Ting and Chen, Min-Hung},
  booktitle={Forty-first International Conference on Machine Learning},
  year={2024}
}

@inproceedings{
zhu2024asymmetry,
title={Asymmetry in Low-Rank Adapters of Foundation Models},
author={Jiacheng Zhu and Kristjan Greenewald and Kimia Nadjahi and Haitz S{\'a}ez de Oc{\'a}riz Borde and Rickard Br{\"u}el Gabrielsson and Leshem Choshen and Marzyeh Ghassemi and Mikhail Yurochkin and Justin Solomon},
booktitle={Forty-first International Conference on Machine Learning},
year={2024},
url={https://openreview.net/forum?id=txRZBD8tBV}
}

@article{achiam2023gpt,
  title={Gpt-4 technical report},
  author={Achiam, Josh and Adler, Steven and Agarwal, Sandhini and Ahmad, Lama and Akkaya, Ilge and Aleman, Florencia Leoni and Almeida, Diogo and Altenschmidt, Janko and Altman, Sam and Anadkat, Shyamal and others},
  journal={arXiv preprint arXiv:2303.08774},
  year={2023}
}

@article{touvron2023llama,
  title={Llama 2: Open foundation and fine-tuned chat models},
  author={Touvron, Hugo and Martin, Louis and Stone, Kevin and Albert, Peter and Almahairi, Amjad and Babaei, Yasmine and Bashlykov, Nikolay and Batra, Soumya and Bhargava, Prajjwal and Bhosale, Shruti and others},
  journal={arXiv preprint arXiv:2307.09288},
  year={2023}
}

@article{grattafiori2024llama,
  title={The llama 3 herd of models},
  author={Grattafiori, Aaron and Dubey, Abhimanyu and Jauhri, Abhinav and Pandey, Abhinav and Kadian, Abhishek and Al-Dahle, Ahmad and Letman, Aiesha and Mathur, Akhil and Schelten, Alan and Vaughan, Alex and others},
  journal={arXiv preprint arXiv:2407.21783},
  year={2024}
}

@inproceedings{
zhao2024galore,
title={GaLore: Memory-Efficient {LLM} Training by Gradient Low-Rank Projection},
author={Jiawei Zhao and Zhenyu Zhang and Beidi Chen and Zhangyang Wang and Anima Anandkumar and Yuandong Tian},
booktitle={Forty-first International Conference on Machine Learning},
year={2024},
url={https://openreview.net/forum?id=hYHsrKDiX7}
}

@inproceedings{
hu2022lora,
title={Lo{RA}: Low-Rank Adaptation of Large Language Models},
author={Edward J Hu and yelong shen and Phillip Wallis and Zeyuan Allen-Zhu and Yuanzhi Li and Shean Wang and Lu Wang and Weizhu Chen},
booktitle={International Conference on Learning Representations},
year={2022},
url={https://openreview.net/forum?id=nZeVKeeFYf9}
}

@inproceedings{
wu2024mixture,
title={Mixture of Lo{RA} Experts},
author={Xun Wu and Shaohan Huang and Furu Wei},
booktitle={The Twelfth International Conference on Learning Representations},
year={2024},
url={https://openreview.net/forum?id=uWvKBCYh4S}
}

@inproceedings{
zadouri2024pushing,
title={Pushing Mixture of Experts to the Limit: Extremely Parameter Efficient MoE for Instruction Tuning},
author={Ted Zadouri and Ahmet {\"U}st{\"u}n and Arash Ahmadian and Beyza Ermis and Acyr Locatelli and Sara Hooker},
booktitle={The Twelfth International Conference on Learning Representations},
year={2024},
url={https://openreview.net/forum?id=EvDeiLv7qc}
}

@inproceedings{
wang2025lorapro,
title={Lo{RA}-Pro: Are Low-Rank Adapters Properly Optimized?},
author={Zhengbo Wang and Jian Liang and Ran He and Zilei Wang and Tieniu Tan},
booktitle={The Thirteenth International Conference on Learning Representations},
year={2025},
url={https://openreview.net/forum?id=gTwRMU3lJ5}
}

@inproceedings{
hayou2024lora,
title={Lo{RA}+: Efficient Low Rank Adaptation of Large Models},
author={Soufiane Hayou and Nikhil Ghosh and Bin Yu},
booktitle={Forty-first International Conference on Machine Learning},
year={2024},
url={https://openreview.net/forum?id=NEv8YqBROO}
}

@inproceedings{
zhang2023adaptive,
title={Adaptive Budget Allocation for Parameter-Efficient Fine-Tuning },
author={Qingru Zhang and Minshuo Chen and Alexander Bukharin and Pengcheng He and Yu Cheng and Weizhu Chen and Tuo Zhao},
booktitle={The Eleventh International Conference on Learning Representations },
year={2023},
url={https://openreview.net/forum?id=lq62uWRJjiY}
}

@inproceedings{
guo2025advancing,
title={Advancing Expert Specialization for Better MoE},
author={Hongcan Guo and Haolang Lu and Guoshun Nan and Bolun Chu and Jialin Zhuang and Yuan Yang and Wenhao Che and Xinye Cao and Sicong Leng and Qimei Cui and Xudong Jiang},
booktitle={The Thirty-ninth Annual Conference on Neural Information Processing Systems},
year={2025},
url={https://openreview.net/forum?id=iydmH9boLb}
}

@article{su2026variational,
  title={Variational Inference, Entropy, and Orthogonality: A Unified Theory of Mixture-of-Experts},
  author={Su, Ye and Liu, Yong},
  journal={arXiv preprint arXiv:2601.03577},
  year={2026}
}

@inproceedings{roemmele2011choice,
  title={Choice of Plausible Alternatives: An Evaluation of Commonsense Causal Reasoning.},
  author={Roemmele, Melissa and Bejan, Cosmin Adrian and Gordon, Andrew S},
  booktitle={AAAI spring symposium: logical formalizations of commonsense reasoning},
  pages={90--95},
  year={2011}
}

@inproceedings{clark-etal-2019-boolq,
    title = "{B}ool{Q}: Exploring the Surprising Difficulty of Natural Yes/No Questions",
    author = "Clark, Christopher  and
      Lee, Kenton  and
      Chang, Ming-Wei  and
      Kwiatkowski, Tom  and
      Collins, Michael  and
      Toutanova, Kristina",
    editor = "Burstein, Jill  and
      Doran, Christy  and
      Solorio, Thamar",
    booktitle = "Proceedings of the 2019 Conference of the North {A}merican Chapter of the Association for Computational Linguistics: Human Language Technologies, Volume 1 (Long and Short Papers)",
    month = jun,
    year = "2019",
    address = "Minneapolis, Minnesota",
    publisher = "Association for Computational Linguistics",
    url = "https://aclanthology.org/N19-1300/",
    doi = "10.18653/v1/N19-1300",
    pages = "2924--2936"
}

@inproceedings{talmor-etal-2019-commonsenseqa,
    title = "{C}ommonsense{QA}: A Question Answering Challenge Targeting Commonsense Knowledge",
    author = "Talmor, Alon  and
      Herzig, Jonathan  and
      Lourie, Nicholas  and
      Berant, Jonathan",
    editor = "Burstein, Jill  and
      Doran, Christy  and
      Solorio, Thamar",
    booktitle = "Proceedings of the 2019 Conference of the North {A}merican Chapter of the Association for Computational Linguistics: Human Language Technologies, Volume 1 (Long and Short Papers)",
    month = jun,
    year = "2019",
    address = "Minneapolis, Minnesota",
    publisher = "Association for Computational Linguistics",
    url = "https://aclanthology.org/N19-1421/",
    doi = "10.18653/v1/N19-1421",
    pages = "4149--4158"
}

@inproceedings{huang-etal-2019-cosmos,
    title = "Cosmos {QA}: Machine Reading Comprehension with Contextual Commonsense Reasoning",
    author = "Huang, Lifu  and
      Le Bras, Ronan  and
      Bhagavatula, Chandra  and
      Choi, Yejin",
    editor = "Inui, Kentaro  and
      Jiang, Jing  and
      Ng, Vincent  and
      Wan, Xiaojun",
    booktitle = "Proceedings of the 2019 Conference on Empirical Methods in Natural Language Processing and the 9th International Joint Conference on Natural Language Processing (EMNLP-IJCNLP)",
    month = nov,
    year = "2019",
    address = "Hong Kong, China",
    publisher = "Association for Computational Linguistics",
    url = "https://aclanthology.org/D19-1243/",
    doi = "10.18653/v1/D19-1243",
    pages = "2391--2401"
}

@inproceedings{sap-etal-2019-social,
    title = "Social {IQ}a: Commonsense Reasoning about Social Interactions",
    author = "Sap, Maarten  and
      Rashkin, Hannah  and
      Chen, Derek  and
      Le Bras, Ronan  and
      Choi, Yejin",
    editor = "Inui, Kentaro  and
      Jiang, Jing  and
      Ng, Vincent  and
      Wan, Xiaojun",
    booktitle = "Proceedings of the 2019 Conference on Empirical Methods in Natural Language Processing and the 9th International Joint Conference on Natural Language Processing (EMNLP-IJCNLP)",
    month = nov,
    year = "2019",
    address = "Hong Kong, China",
    publisher = "Association for Computational Linguistics",
    url = "https://aclanthology.org/D19-1454/",
    doi = "10.18653/v1/D19-1454",
    pages = "4463--4473"
}

@inproceedings{bisk2020piqa,
  title={Piqa: Reasoning about physical commonsense in natural language},
  author={Bisk, Yonatan and Zellers, Rowan and Gao, Jianfeng and Choi, Yejin and others},
  booktitle={Proceedings of the AAAI conference on artificial intelligence},
  volume={34},
  number={05},
  pages={7432--7439},
  year={2020}
}

@inproceedings{
lu2022learn,
title={Learn to Explain: Multimodal Reasoning via Thought Chains for Science Question Answering},
author={Pan Lu and Swaroop Mishra and Tony Xia and Liang Qiu and Kai-Wei Chang and Song-Chun Zhu and Oyvind Tafjord and Peter Clark and Ashwin Kalyan},
booktitle={Advances in Neural Information Processing Systems},
editor={Alice H. Oh and Alekh Agarwal and Danielle Belgrave and Kyunghyun Cho},
year={2022},
url={https://openreview.net/forum?id=HjwK-Tc_Bc}
}

@article{clark2018think,
  title={Think you have solved question answering? try arc, the ai2 reasoning challenge},
  author={Clark, Peter and Cowhey, Isaac and Etzioni, Oren and Khot, Tushar and Sabharwal, Ashish and Schoenick, Carissa and Tafjord, Oyvind},
  journal={arXiv preprint arXiv:1803.05457},
  year={2018}
}

@inproceedings{mihaylov-etal-2018-suit,
    title = "Can a Suit of Armor Conduct Electricity? A New Dataset for Open Book Question Answering",
    author = "Mihaylov, Todor  and
      Clark, Peter  and
      Khot, Tushar  and
      Sabharwal, Ashish",
    editor = "Riloff, Ellen  and
      Chiang, David  and
      Hockenmaier, Julia  and
      Tsujii, Jun{'}ichi",
    booktitle = "Proceedings of the 2018 Conference on Empirical Methods in Natural Language Processing",
    month = oct # "-" # nov,
    year = "2018",
    address = "Brussels, Belgium",
    publisher = "Association for Computational Linguistics",
    url = "https://aclanthology.org/D18-1260/",
    doi = "10.18653/v1/D18-1260",
    pages = "2381--2391"
}

@inproceedings{
zou2025flylora,
title={FlyLo{RA}: Boosting Task Decoupling and Parameter Efficiency via Implicit Rank-Wise Mixture-of-Experts},
author={Heming Zou and Yunliang Zang and Wutong Xu and Yao Zhu and Xiangyang Ji},
booktitle={The Thirty-ninth Annual Conference on Neural Information Processing Systems},
year={2025},
url={https://openreview.net/forum?id=nGQLYn13Xf}
}

@article{yang2024qwen2,
  title={Qwen2. 5 Technical Report},
  author={Yang, An and Yang, Baosong and Zhang, Beichen and Hui, Binyuan and Zheng, Bo and Yu, Bowen and Li, Chengyuan and Liu, Dayiheng and Huang, Fei and Wei, Haoran and others},
  journal={arXiv preprint arXiv:2412.15115},
  year={2024}
}

@article{van2008visualizing,
  title={Visualizing data using t-SNE.},
  author={Van der Maaten, Laurens and Hinton, Geoffrey},
  journal={Journal of machine learning research},
  volume={9},
  number={11},
  year={2008}
}

@inproceedings{
xu2026low,
title={Low Kruskal-Rank Adaptation},
author={Yixing Xu and Guanchen Li and Chao Li and Xuanwu Yin and Dong Li and Spandan Tiwari and Ashish Sirasao and Emad Barsoum},
booktitle={Forty-third International Conference on Machine Learning},
year={2026},
url={https://openreview.net/forum?id=UcnpfSsXtu}
}

\clearpage
\appendix
\section*{Appendix}

The appendix is organized as follows:
\begin{itemize}
    \item Appendix~\ref{app:odmole-theory} provides proofs and supporting theoretical analysis.
    \begin{itemize}
        \item Proof of the routing-recovery theorem (Section~\ref{app:routing-recovery-proof}).
        \item Output decorrelation and routing recovery in top-$K$ routing (Section~\ref{sec:topk}).
        \item Task-weighted extension and gradient outer products (Section~\ref{app:task-weighted-extension}).
        \item Population guarantees for output decorrelation (Section~\ref{app:population-decorrelation}).
        \item Proof of the expert-output difference bound (Section~\ref{app:expert-differences-proof}).
        \item Proof of the shared-diversity theorem (Section~\ref{app:shared-diversity-proof}).
        \item Sampled determinants and the implemented regularizer (Section~\ref{app:sampled-determinants}).
    \end{itemize}

    \item Appendix~\ref{app:experiment-details} provides
    experimental details and additional analyses.
    \begin{itemize}
        \item Datasets (Section~\ref{app:datasets}).
        \item Baselines (Section~\ref{app:baselines}).
        \item Dataset-specific training schedules (Section~\ref{app:training-schedules}).
        \item Sensitivity to regularization coefficients (Section~\ref{app:regularization-sensitivity}).
        \item Expert-output similarity analysis (Section~\ref{app:output-similarity}).
        \item Effect of the number of experts (Section~\ref{app:effect_number_of_experts}).
        \item Multi-task learning with task-assigned experts (Section~\ref{app:task-assigned-routing}).
        \item Comparison with alternative specialization objectives (Section~\ref{app:guo-comparison}).
         \item Statistical significance of the comparison with HydraLoRA (Section~\ref{app:five-seed}).
        \item Per-seed results of the regularizer ablation (Section~\ref{app:ablation-per-seed}).
        \item Analysis of output orthogonality (Section~\ref{app:orth-analysis}).
        \item Analysis of shared-projection diversity (Section~\ref{app:div-analysis}).
        \item Mathematical reasoning and code generation (Section~\ref{app:more-tasks}).
        \item Inference latency (Section~\ref{app:latency}).
    \end{itemize}

    \item Appendix~\ref{subsec:param_num} compares the parameter counts of OD-MoLE and the baselines.
\end{itemize}

\section{Proofs and Details of the Theoretical Analysis}
\label{app:odmole-theory}
In this section, we follow the notation of Sections~\ref{sec:preliminaries} and~\ref{sec:theory}. Recall, the unscaled expert residual is $\boldsymbol{z}_i(\boldsymbol{h})=\boldsymbol{B}_i\boldsymbol{A}\boldsymbol{h}$, and the full output is $\boldsymbol{W}_{0}\boldsymbol{h}+c\sum_{i=1}^{N}g_i(\boldsymbol{h})\boldsymbol{z}_i(\boldsymbol{h})$, where $c=\gamma/r>0$. The routing analysis fixes $\boldsymbol{h}$ and expert parameters, whereas the diversity analysis takes expectations over a fixed activation distribution with finite second moments. We first prove routing recovery and clarify its scope, then establish the shared-projection bound and its connection to the regularizer.

\subsection{Proof of output decorrelation and routing recovery (Theorem~\ref{thm:routing-recovery})}
\label{app:routing-recovery-proof}
\begin{proof}
Fix an input activation $\boldsymbol{h}$, the expert parameters, and the target $\boldsymbol{t}_{\boldsymbol{h}}$. Assume $c>0$, $0<\underline{z}\leq\|\boldsymbol{z}_i(\boldsymbol{h})\|_2\leq\overline{z}<\infty$ for every expert $i$, and $\delta(\boldsymbol{h})<1$.

Define
\begin{equation}\label{eq:app-DVR}
\boldsymbol{D}=\text{diag}
(\|\boldsymbol{z}_1\|_2,\ldots,\|\boldsymbol{z}_N\|_2),
\qquad \boldsymbol{V}=\boldsymbol{Z}\boldsymbol{D}^{-1},
\qquad \boldsymbol{R}=\boldsymbol{V}^\top\boldsymbol{V}.
\end{equation}
The positive norm lower bound makes $\boldsymbol{D}$ invertible. The columns of $\boldsymbol{V}$ are normalized expert outputs, so $\boldsymbol{R}$ has unit diagonal and pairwise cosine similarities off the diagonal. Hence,
\begin{equation}
\|\boldsymbol{R}-\boldsymbol{I}\|_{\mathrm F}^2
=N(N-1)\omega(\boldsymbol{h})=\delta(\boldsymbol{h})^2.
\label{eq:app-RI}
\end{equation}
Since the operator norm is bounded by the Frobenius norm, we have $\|\boldsymbol{R}-\boldsymbol{I}\|_{\mathrm{op}}\leq\delta(\boldsymbol{h})$. Every eigenvalue of $\boldsymbol{R}$ therefore lies between $1-\delta(\boldsymbol{h})$ and $1+\delta(\boldsymbol{h})$, giving
\begin{equation}
(1-\delta(\boldsymbol{h}))\boldsymbol{I}\preceq\boldsymbol{R}
\preceq(1+\delta(\boldsymbol{h}))\boldsymbol{I}.
\end{equation}
Define the routing Gram matrix $\boldsymbol{G}=c^2\boldsymbol{Z}^{\top}\boldsymbol{Z}$. The $\boldsymbol{Z}^\top\boldsymbol{Z}=\boldsymbol{D}\boldsymbol{R}\boldsymbol{D}$ and the output-norm bounds imply
\begin{equation}
c^2\underline z^2(1-\delta(\boldsymbol{h}))\boldsymbol{I}\leq\boldsymbol{G}\leq c^2\overline z^2(1+\delta(\boldsymbol{h}))\boldsymbol{I},
\qquad \boldsymbol{G}=c^2\boldsymbol{Z}^\top\boldsymbol{Z}.
\label{eq:app-gram-spectrum}
\end{equation}
For any $\boldsymbol{g},\widehat{\boldsymbol{g}}\in\Delta_N$, we have
\begin{equation}
    \|c\boldsymbol{Z}(\boldsymbol{g}-\widehat{\boldsymbol{g}})\|_2^2
=(\boldsymbol{g}-\widehat{\boldsymbol{g}})^\top
\boldsymbol{G}(\boldsymbol{g}-\widehat{\boldsymbol{g}})\geq c^2\underline{z}^2(1-\delta(\boldsymbol{h}))
\|\boldsymbol{g}-\widehat{\boldsymbol{g}}\|_2^2.
\end{equation}
This proves Eq.~\ref{eq:output-separation}.

Since $\Delta_N$ is compact and $Q_{\boldsymbol{h}}$ is continuous, a minimizer $\boldsymbol{g}^{*}$ exists. Its Hessian with respect to routing weights is $\boldsymbol{G}\succ0$ by $\delta(\boldsymbol{h})<1$ and $c>0$, so this minimizer $\boldsymbol{g}^{*}$ is unique. At this constrained minimizer,
\begin{equation}
\nabla Q_{\boldsymbol{h}}(\boldsymbol{g}^{{*}})^{\top}
(\boldsymbol{g}-\boldsymbol{g}^{{*}})\geq0,
\qquad \boldsymbol{g}\in\Delta_N.
\label{eq:app-routing-optimality}
\end{equation}
Since the simplex is convex, moving from $\boldsymbol{g}^{{*}}$ toward any feasible $\boldsymbol{g}$ remains feasible. Therefore, the optimality of $\boldsymbol{g}^{{*}}$ requires the directional derivative along $\boldsymbol{g}-\boldsymbol{g}^{{*}}$ to be  nonnegative.

For $\boldsymbol{v}=\boldsymbol{g}-\boldsymbol{g}^{*}$, the exact quadratic expansion gives
\begin{equation}
Q_{\boldsymbol{h}}(\boldsymbol{g})-Q_{\boldsymbol{h}}(\boldsymbol{g}^{*})
=\nabla Q_{\boldsymbol{h}}(\boldsymbol{g}^{*})^\top\boldsymbol{v}
+\frac{1}{2}\boldsymbol{v}^\top\boldsymbol{G}\boldsymbol{v}\geq\frac{c^2\underline z^2(1-\delta(\boldsymbol{h}))}{2}
\|\boldsymbol{g}-\boldsymbol{g}^{*}\|_2^2.
\label{app:exact_qua}
\end{equation}
Rearranging the above proves Eq.~\ref{eq:routing-recovery}. Eq.~\ref{app:exact_qua} also proves uniqueness, since the distance between two minimizers must be zero. The proof also applies when some optimal routing weights are zero.
\end{proof}
The output orthogonality regularizer of OD-MoLE targets $\omega(\boldsymbol{h})$ on input activations. The theorem explains when the norm assumptions hold and $\delta(\boldsymbol{h})<1$, differences in routing weights cannot diminish in the mixture output. Consequently, a small excess residual loss implies proximity to the oracle mixture weights for the fixed target $\boldsymbol{t}_{\boldsymbol{h}}$.
\paragraph{Role and scope of the assumptions.}
The positive norm lower bound matters because directional separation alone does not control the magnitude of output changes. For example, outputs of the form $\boldsymbol{z}_i=\epsilon\boldsymbol{e}_i$ are mutually orthogonal, but satisfy
\begin{equation}
\|c\boldsymbol{Z}
(\boldsymbol{g}-\widehat{\boldsymbol{g}})\|_2
=c\varepsilon
\|\boldsymbol{g}-\widehat{\boldsymbol{g}}\|_2.
\end{equation}
As $\epsilon$ approaches zero, distinct routing vectors become difficult to distinguish from their outputs even though their normalized directions remain perfectly separated. 

The condition $\delta(\boldsymbol{h})<1$, equivalently $\omega(\boldsymbol{h})<1/[N(N-1)]$, ensures positive definiteness of the normalized Gram matrix. It is a sufficient rather than necessary condition for routing identifiability. In particular, it implies that $\boldsymbol{Z}$ has full column rank and hence $N\leq d$.

Theorem~\ref{thm:routing-recovery} imposes these conditions at the activation being analyzed. If $\boldsymbol{A}\boldsymbol{h}=0$, every expert residual vanishes. More generally, for any scalar $s>0$, because $\boldsymbol{z}_i(s\boldsymbol{h})=s\boldsymbol{z}_i(\boldsymbol{h})$, a uniform positive lower bound cannot hold on an input set containing arbitrary small scalings of a nonzero activation. 

\paragraph{Example 1 (routing recovery without decorrelation).}\label{app:example-opposite-outputs}
The following example shows that routing weights can be recovered uniquely and stably even when expert outputs point in opposite directions, so the correlation condition is sufficient but not necessary for recovery. Fix a nonzero vector $\boldsymbol{v}\in\mathbb{R}^d$ and let $\boldsymbol{z}_1=\boldsymbol{v}$ and $\boldsymbol{z}_2=-\boldsymbol{v}$. Their squared cosine similarity is one, so $\omega(\boldsymbol{h})=1$ and $\delta(\boldsymbol{h})=\sqrt{2}>1$. Nevertheless, for $\boldsymbol{g}=(a,1-a)^\top$ with $a\in[0,1]$,
\begin{equation}
c\boldsymbol{Z}\boldsymbol{g}=c(2a-1)\boldsymbol{v}.
\end{equation}
This output uniquely determines $a$. More precisely, for $\widehat{\boldsymbol{g}}=(b,1-b)^\top$,
\begin{equation}
\|c\boldsymbol{Z}(\boldsymbol{g}-\widehat{\boldsymbol{g}})\|_2^2
=2c^2\|\boldsymbol{v}\|_2^2
\|\boldsymbol{g}-\widehat{\boldsymbol{g}}\|_2^2.
\end{equation}
Thus, squared-cosine regularization can penalize configurations that already distinguish routing weights. The theorem assumes nonzero expert outputs so that cosine similarities are well-defined, but this assumption is not necessary for routing weights to be uniquely recoverable.

\paragraph{The relevant routing directions.}
Example 1 above is possible because routing differences satisfy $\boldsymbol{1}^\top(\boldsymbol{g}-\widehat{\boldsymbol{g}})=0$. For $N\geq2$, define
\begin{equation}
m_{\Delta}(\boldsymbol{h})=
\min_{\substack{\boldsymbol{1}^\top\boldsymbol{v}=0\\
\|\boldsymbol{v}\|_2=1}}
\|\boldsymbol{Z}(\boldsymbol{h})\boldsymbol{v}\|_2^2.
\label{eq:app-simplex-curvature}
\end{equation}
Different routing vectors in $\Delta_N$ produce different mixture outputs if and only if $m_{\Delta}(\boldsymbol{h})>0$. Equivalently, the expert outputs are affinely independent, so each mixture has a unique representation by routing weights. If the minimum is zero, there exists a nonzero vector $\boldsymbol{v}$ satisfying $\boldsymbol{1}^\top\boldsymbol{v}=0$ and $\boldsymbol{Z}(\boldsymbol{h})\boldsymbol{v}=\boldsymbol{0}$. For sufficiently small $s>0$, both routing vectors $\boldsymbol{g}_0=\frac{1}{N}\boldsymbol{1}$ and $\boldsymbol{g}_0+s\boldsymbol{v}$ belong to $\Delta_N$ and produce the same mixture output. Conversely, a positive minimum gives
\begin{equation}
    \|\boldsymbol{Z}(\boldsymbol{h})
(\boldsymbol{g}-\widehat{\boldsymbol{g}})\|_2^2
\geq m_{\Delta}(\boldsymbol{h})
\|\boldsymbol{g}-\widehat{\boldsymbol{g}}\|_2^2,
\end{equation}
so distinct routing vectors produce distinct outputs. We use the same reasoning as in the previous proof, where the routing weights must remain in the simplex. Then we have
\begin{equation}
\|\boldsymbol{g}-\boldsymbol{g}^{*}\|_2^2
\leq\frac{2[Q_{\boldsymbol{h}}(\boldsymbol{g})-Q_{\boldsymbol{h}}(\boldsymbol{g}^{*})]}
{c^2m_{\Delta}(\boldsymbol{h})}.
\end{equation}
Theorem~\ref{thm:routing-recovery} provides a convenient lower bound $m_{\Delta}(\boldsymbol{h})\geq\underline{z}^2(1-\delta(\boldsymbol{h}))$. Unique routing recovery can hold with more experts than the theorem’s sufficient condition permits. There are two different requirements. First, affine independence only requires that different routing weights summing to one give different outputs. In $\mathbb{R}^{d}$, up to $d+1$ expert outputs can satisfy this. Second, full column rank requires all $N$ output vectors to be linearly independent, a stronger condition that permits at most $d$ vectors in $\mathbb{R}^{d}$. Thus, $\delta(\boldsymbol{h})<1$ is sufficient for routing recovery but excludes some configurations where recovery is still possible.

\subsection{Output decorrelation and routing recovery in top-$K$ routing}
\label{sec:topk}
We prove Corollary~\ref{cor:topk} using the definitions in Eqs.~\ref{eq:topk-simplex}--\ref{eq:mDeltaK}, then establish recovery guarantees for the quadratic objective $Q_{\boldsymbol{h}}$.

Fix $\boldsymbol{h}$. Let $\boldsymbol{D}$ and $\boldsymbol{R}$ be the diagonal matrix of output norms and the normalized Gram matrix from Eq.~\ref{eq:app-DVR}. For a nonempty $\mathcal{T}\subseteq[N]$, let $\boldsymbol{D}_{\mathcal{T}}$ and $\boldsymbol{R}_{\mathcal{T}}$ denote the corresponding principal submatrices, and let $\boldsymbol{Z}_{\mathcal{T}}$ collect the corresponding expert outputs. Every routing difference has zero sum, so $m_{\Delta}^{K}(\boldsymbol{h})\geq m_{\Delta}(\boldsymbol{h})$ by Eq.~\ref{eq:app-simplex-curvature}.

\begin{proof}
Suppress $\boldsymbol{h}$. For any nonempty $\mathcal{T}$ with $\delta_{\mathcal{T}}<1$, the operator norm bound $\|\boldsymbol{R}_{\mathcal{T}}-\boldsymbol{I}\|_{\mathrm{op}}\leq\delta_{\mathcal{T}}$ gives$\boldsymbol{R}_{\mathcal{T}}\succeq(1-\delta_{\mathcal{T}})\boldsymbol{I}$. As in the proof of Eq.~\ref{eq:app-gram-spectrum}, this yields
\begin{equation}
\boldsymbol{g}_{\mathcal{T}}=c^2\bm D_{\mathcal{T}}\boldsymbol{R}_{\mathcal{T}}\bm D_{\mathcal{T}}\succeq c^2(1-\delta_{\mathcal{T}})\bm D_{\mathcal{T}}^2\succeq c^2\underline z^2(1-\delta_{\mathcal{T}})\boldsymbol{I}.
\label{eq:topk-Gbound}
\end{equation}
{\sffamily{Part (i).}} Let $\boldsymbol{g},\widehat{\boldsymbol{g}}\in\Delta_N^K$, $\boldsymbol{v}=\boldsymbol{g}-\widehat{\boldsymbol{g}}$, and $\mathcal{T}=\operatorname{supp}(\boldsymbol{g})\cup\operatorname{supp}(\widehat{\boldsymbol{g}})$. Since $\boldsymbol{1}^\top\boldsymbol{g}=1$, $\mathcal{T}$ is nonempty. Hence $1\le|\mathcal{T}|\le\kappa$ and $\delta_{\mathcal{T}}\le\delta_{2K}<1$. Since $v_i=0$ for $i\notin\mathcal{T}$, we have $\boldsymbol{v}^\top\boldsymbol{g}\boldsymbol{v}=\boldsymbol{v}_{\mathcal{T}}^\top\boldsymbol{g}_{\mathcal{T}}\boldsymbol{v}_{\mathcal{T}}$ and $\|\boldsymbol{v}\|_2=\|\boldsymbol{v}_{\mathcal{T}}\|_2$. By Eq.~\ref{eq:topk-Gbound}, we have
\begin{equation}
\begin{aligned}
\|c\bm Z(\boldsymbol{g}-\widehat{\boldsymbol{g}})\|_2^2&=\boldsymbol{v}_{\mathcal{T}}^\top\boldsymbol{g}_{\mathcal{T}}\boldsymbol{v}_{\mathcal{T}} \ge c^2\underline z^2(1-\delta_{\mathcal{T}})\|\boldsymbol{v}\|_2^2 \ge c^2\underline z^2(1-\delta_{2K})\|\boldsymbol{g}-\widehat{\boldsymbol{g}}\|_2^2.
\end{aligned}
\end{equation}
leading the the bound in Eq.~\ref{eq:topk-sep}. Dividing by $c^2\|\boldsymbol{g}-\widehat{\boldsymbol{g}}\|_2^2$ and taking the infimum gives the bound on $m_\Delta^K$ as desired.

{\sffamily{Part (ii).}} Fix $i\neq j$. Since $N-2\ge K-1$, choose $\mathcal{W}\subseteq[N]\setminus\{i,j\}$ with $|\mathcal{W}|=K-1$. If $K=1$, set $\boldsymbol{g}=\bm e_i$ and $\widehat{\boldsymbol{g}}=\bm e_j$. If $K\ge 2$, fix $a\in(0,1)$ and set
\begin{equation}
\boldsymbol{g}=a\,\bm e_i+\frac{1-a}{K-1}\sum_{k\in\mathcal{W}}\bm e_k,\qquad
\widehat{\boldsymbol{g}}=a\,\bm e_j+\frac{1-a}{K-1}\sum_{k\in\mathcal{W}}\bm e_k.
\end{equation}
Both vectors are nonnegative, satisfy $\bm 1^\top\boldsymbol{g}=\bm 1^\top\widehat{\boldsymbol{g}}=1$, and have exactly $K$ positive entries, so $\boldsymbol{g},\widehat{\boldsymbol{g}}\in\Delta_N^K$. Moreover, $j\notin\operatorname{supp}(\boldsymbol{g})$ and $i\notin\operatorname{supp}(\widehat{\boldsymbol{g}})$. In both cases $\boldsymbol{g}-\widehat{\boldsymbol{g}}=a(\bm e_i-\bm e_j)$ with $a\in(0,1]$, so
\begin{equation}
\frac{\|\bm Z(\boldsymbol{g}-\widehat{\boldsymbol{g}})\|_2^2}{\|\boldsymbol{g}-\widehat{\boldsymbol{g}}\|_2^2}=\frac{a^2\|\boldsymbol{z}_i-\bm z_j\|_2^2}{2a^2}=\frac12\|\boldsymbol{z}_i-\bm z_j\|_2^2,
\end{equation}
and $m_\Delta^K$ is at most this value. The second expression in Eq.~\ref{eq:topk-upper} follows by expanding $\|\boldsymbol{z}_i-\boldsymbol{z}_j\|_2^2$ and using $\boldsymbol{z}_i^\top\boldsymbol{z}_j
=\|\boldsymbol{z}_i\|_2\|\boldsymbol{z}_j\|_2 R_{ij}$, where $R_{ij}$ is the cosine similarity between the two outputs.

{\sffamily{Part (iii).}} Since $K\ge 1$ and $N\ge 2$, every pair $\mathcal{T}=\{i,j\}$ satisfies $|\mathcal{T}|=2\le\kappa$, and $\delta_{\{i,j\}}^2=2R_{ij}^2$,  maximizing over pairs gives the lower bound. For any admissible $\mathcal{T}$, $\delta_{\mathcal{T}}^2=\sum_{i\neq j\in\mathcal{T}}R_{ij}^2\le|\mathcal{T}|(|\mathcal{T}|-1)\max_{i\neq j}R_{ij}^2\le\kappa(\kappa-1)\max_{i\neq j}R_{ij}^2$. Moreover, $\boldsymbol{R}_{\mathcal{T}}-\boldsymbol{I}$ is a principal submatrix of $\boldsymbol{R}-\boldsymbol{I}$, so by Eq.~\ref{eq:app-RI}, $\delta_{\mathcal{T}}^2\le\|\boldsymbol{R}-\boldsymbol{I}\|_{\mathrm F}^2=N(N-1)\omega(\boldsymbol{h})=\delta(\boldsymbol{h})^2$. Maximizing over admissible $\mathcal{T}$ gives the upper bound and $\delta_{2K}\le\delta(\boldsymbol{h})$. When $2K\ge N$, $\mathcal{T}=[N]$ is admissible, so equality holds.
\end{proof}

\begin{corollary}[Routing recovery under top-$K$ routing]
\label{cor:topk-rec}
Under the assumptions of Corollary~\ref{cor:topk}, fix a target residual $\boldsymbol{t}_{\boldsymbol{h}}$ and let $Q_{\boldsymbol{h}}$ be the objective of Eq.~\ref{eq:Q_h}.
\begin{enumerate}
\item[(i)] If $\delta_{2K}(\boldsymbol{h})<1$ and $\boldsymbol{t}_{\boldsymbol{h}}=c\bm Z(\boldsymbol{h})\boldsymbol{g}^\circ$ for some $\boldsymbol{g}^\circ\in\Delta_N^K$, then $\boldsymbol{g}^\circ$ is the unique minimizer of $Q_{\boldsymbol{h}}$ on $\Delta_N^K$, and each $\boldsymbol{g}\in\Delta_N^K$ satisfies
\begin{equation}
\|\boldsymbol{g}-\boldsymbol{g}^\circ\|_2^2\le\frac{2Q_{\boldsymbol{h}}(\boldsymbol{g})}{c^2\underline z^2\bigl(1-\delta_{2K}(\boldsymbol{h})\bigr)}.
\label{eq:topk-real}
\end{equation}
\item[(ii)] Let $\mathcal{T}\subseteq[N]$ be nonempty with $\delta_{\mathcal{T}}(\boldsymbol{h})<1$, and let $\Delta_{\mathcal{T}}=\{\boldsymbol{g}\in\Delta_N:\operatorname{supp}(\boldsymbol{g})\subseteq\mathcal{T}\}$. Then $Q_{\boldsymbol{h}}$ has a unique minimizer $\boldsymbol{g}^*_{\mathcal{T}}(\boldsymbol{h})$ on $\Delta_{\mathcal{T}}$, and each $\boldsymbol{g}\in\Delta_{\mathcal{T}}$ satisfies
\begin{equation}
\|\boldsymbol{g}-\boldsymbol{g}^*_{\mathcal{T}}(\boldsymbol{h})\|_2^2\le\frac{2[Q_{\boldsymbol{h}}(\boldsymbol{g})-Q_{\boldsymbol{h}}(\boldsymbol{g}^*_{\mathcal{T}}(\boldsymbol{h}))]}{c^2\underline z^2\bigl(1-\delta_{\mathcal{T}}(\boldsymbol{h})\bigr)}.
\label{eq:topk-face}
\end{equation}
\end{enumerate}
\end{corollary}

\begin{proof}
{\sffamily{Part (i).}} For $\boldsymbol{g}\in\Delta_N^K$, Eq.~\ref{eq:topk-sep} gives $Q_{\boldsymbol{h}}(\boldsymbol{g})=\frac12\|c\bm Z(\boldsymbol{g}-\boldsymbol{g}^\circ)\|_2^2\ge\frac12c^2\underline z^2(1-\delta_{2K})\|\boldsymbol{g}-\boldsymbol{g}^\circ\|_2^2$. Since $Q_{\boldsymbol{h}}(\boldsymbol{g}^\circ)=0\le Q_{\boldsymbol{h}}(\boldsymbol{g})$, $\boldsymbol{g}^\circ$ is a minimizer. The inequality forces any other minimizer to equal $\boldsymbol{g}^\circ$, and rearranging gives Eq.~\ref{eq:topk-real}.

{\sffamily{Part (ii).}} $\Delta_{\mathcal{T}}$ is a nonempty, compact, convex face of $\Delta_N$ and $Q_{\boldsymbol{h}}$ is continuous, so a minimizer $\boldsymbol{g}^*_{\mathcal{T}}$ exists. For $\boldsymbol{g}\in\Delta_{\mathcal{T}}$, let $\boldsymbol{v}=\boldsymbol{g}-\boldsymbol{g}^*_{\mathcal{T}}$. Then $\operatorname{supp}(\boldsymbol{v})\subseteq\mathcal{T}$, so Eq.~\ref{eq:topk-Gbound} gives $\boldsymbol{v}^\top\boldsymbol{g}\boldsymbol{v}=\boldsymbol{v}_{\mathcal{T}}^\top\boldsymbol{g}_{\mathcal{T}}\boldsymbol{v}_{\mathcal{T}}\ge c^2\underline z^2(1-\delta_{\mathcal{T}})\|\boldsymbol{v}\|_2^2$. Since $\Delta_{\mathcal{T}}$ is convex, moving from $\boldsymbol{g}^*_{\mathcal{T}}$ toward $\boldsymbol{g}$ remains feasible, so, as in Eq.~\ref{eq:app-routing-optimality}, $\nabla Q_{\boldsymbol{h}}(\boldsymbol{g}^*_{\mathcal{T}})^\top\boldsymbol{v}\ge 0$. The exact quadratic expansion of Eq.~\ref{app:exact_qua} then gives
\begin{equation}
Q_{\boldsymbol{h}}(\boldsymbol{g})-Q_{\boldsymbol{h}}(\boldsymbol{g}^*_{\mathcal{T}})=\nabla Q_{\boldsymbol{h}}(\boldsymbol{g}^*_{\mathcal{T}})^\top\boldsymbol{v}+\frac12\boldsymbol{v}^\top\boldsymbol{g}\boldsymbol{v}\ge\frac{c^2\underline z^2(1-\delta_{\mathcal{T}})}{2}\|\boldsymbol{g}-\boldsymbol{g}^*_{\mathcal{T}}\|_2^2.
\end{equation}
Rearranging proves Eq.~\ref{eq:topk-face}, and applying it to a second minimizer proves uniqueness.
\end{proof}

\paragraph{Non-uniqueness across supports.}
Theorem~\ref{thm:routing-recovery} and Corollary~\ref{cor:topk-rec}(ii) use convexity of the feasible set. For $K<N$, $\Delta_N^K$ is a union of faces of $\Delta_N$ and is not convex, so Eq.~\ref{eq:routing-recovery} does not extend across supports without further assumptions. Uniqueness can fail even for orthogonal expert outputs: take $N=2$, $K=1$, $c=1$, $\boldsymbol{z}_1=\boldsymbol{e}_1$, $\bm z_2=\boldsymbol{e}_2$, and $\boldsymbol{t}_{\boldsymbol{h}}=\frac12(\boldsymbol{e}_1+\boldsymbol{e}_2)$. Then $\delta_{2K}(\boldsymbol{h})=\delta(\boldsymbol{h})=0$, yet $\Delta_2^1=\{\boldsymbol{e}_1,\boldsymbol{e}_2\}$ and $Q_{\boldsymbol{h}}(\boldsymbol{e}_1)=Q_{\boldsymbol{h}}(\boldsymbol{e}_2)=\frac14$, so $Q_{\boldsymbol{h}}$ has two minimizers on $\Delta_2^1$ at distance $\sqrt2$. Across supports, the statements that transfer are separation in Eq.~\ref{eq:topk-sep} and recovery of realizable targets in Corollary~\ref{cor:topk-rec}(i).

\paragraph{Relation to Theorem~\ref{thm:routing-recovery}.}
Since $m_{\Delta}^{K}(\boldsymbol{h})\geq m_{\Delta}(\boldsymbol{h})\geq\underline{z}^2(1-\delta(\boldsymbol{h}))$ when $\delta(\boldsymbol{h})<1$, the bound of Theorem~\ref{thm:routing-recovery} remains valid under top-$K$ routing, and Corollary~\ref{cor:topk}(i) improves it to $\underline{z}^2(1-\delta_{2K}(\boldsymbol{h}))$ under a weaker condition. By Eq.~\ref{eq:topk-pairs}, $\delta(\boldsymbol{h})<1$ is guaranteed when $\max_{i\neq j}R_{ij}(\boldsymbol{h})^2<1/[N(N-1)]$, whereas $\delta_{2K}(\boldsymbol{h})<1$ is guaranteed when $\max_{i\neq j}R_{ij}(\boldsymbol{h})^2<1/[\kappa(\kappa-1)]$, a threshold that depends on $K$ and not on $N$. The top-$K$ condition also relaxes the dimension requirement: $\delta(\boldsymbol{h})<1$ forces $\boldsymbol{Z}(\boldsymbol{h})$ to have full column rank and hence $N\leq d$, whereas $\delta_{2K}(\boldsymbol{h})<1$ requires only that every $\kappa$ expert outputs are linearly independent, which is compatible with $N>d$.

\paragraph{Connection to the regularizer.}
The all-pairs orthogonality loss controls the cross-support separation bound of Corollary~\ref{cor:topk}(i) through $\delta_{2K}(\boldsymbol{h})^2\leq N(N-1)\,\omega(\boldsymbol{h})$. The selected-support loss penalizes the correlations that enter the fixed-support recovery bound in Eq.~\ref{eq:topk-face}. Eq.~\ref{eq:topk-upper} explains why cross-support comparisons also depend on pairs that are never selected together. In particular, a zero mean squared output difference for any pair implies $m_\Delta^K(\boldsymbol{h})=0$ almost surely. This connects top-$K$ separation to the pairwise output differences in Eq.~\ref{eq:expert_differences}.

\paragraph{Connection to the experiments.}
The routing-sensitivity experiment in Tab.~\ref{tab:orth-mechanism} holds the selected pair fixed and measures the response to a weight transfer within that pair, so it probes the within-support quantity of Corollary~\ref{cor:topk-rec}(ii). The ablation in Tab.~\ref{tab:pairs-seeds} compares the two training losses under top-$2$ routing with diversity disabled. The selected-pair and all-pairs losses reach eight-task average accuracies of $88.19\%$ and $88.20\%$ over three seeds. Both losses control the within-support correlation, and the mean paired difference (selected-pair minus all-pairs) across three seeds is $-0.007$ pp, with a $95\%$ confidence interval of $[-0.477,\,0.463]$.

\subsection{Extension to task-weighted routing objectives}
\label{app:task-weighted-extension}
Theorem~\ref{thm:routing-recovery} measures residual differences in Euclidean distance. We now consider a weighted quadratic objective to account for different sensitivities along output directions. Fix the input $\boldsymbol{h}$, expert parameters, target $\boldsymbol{t}_{\boldsymbol{h}}$, and a symmetric matrix $\boldsymbol{C}_h\succ0$. Define
\begin{equation}
Q_{\boldsymbol{h}}^C(\boldsymbol{g})
=\frac{1}{2}
(c\boldsymbol{Z}\boldsymbol{g}-\boldsymbol{t}_{\boldsymbol{h}})^\top
\boldsymbol{C}_h
(c\boldsymbol{Z}\boldsymbol{g}-\boldsymbol{t}_{\boldsymbol{h}}),
\qquad \boldsymbol{g}\in\Delta_N,
\end{equation}
where $c>0$ and $\boldsymbol{Z}=\boldsymbol{Z}(\boldsymbol{h})$. The Euclidean objective corresponds to $\boldsymbol{C}_h=\boldsymbol{I}$.

Assume the hypotheses of Theorem~\ref{thm:routing-recovery}. Suppose further that
\begin{equation}
c_-\|\boldsymbol{u}\|_2^2
\leq\boldsymbol{u}^\top\boldsymbol{C}_h\boldsymbol{u}
\leq c_+\|\boldsymbol{u}\|_2^2,
\qquad 0<c_-\leq c_+<\infty,
\end{equation}
for every $\boldsymbol{u}$ in the span of the expert outputs. Writing $\boldsymbol{G}=c^2\boldsymbol{Z}^\top\boldsymbol{Z}$, the Hessian of the weighted objective satisfies
\begin{equation}
c_-\boldsymbol{G}
\preceq
\boldsymbol{G}_C
:=c^2\boldsymbol{Z}^\top\boldsymbol{C}_h\boldsymbol{Z}
\preceq c_+\boldsymbol{G}.
\end{equation}

Thus, with $\mu_h=c^2\underline{z}^2(1-\delta(\boldsymbol{h}))>0$, the weighted objective has curvature at least $c_-\mu_h$. Applying the same constrained-optimality argument gives a unique minimizer $\boldsymbol{g}_C^{*}$ on $\Delta_N$ and
\begin{equation}
\|\boldsymbol{g}-\boldsymbol{g}_C^{*}\|_2^2
\leq
\frac{2[Q_{\boldsymbol{h}}^C(\boldsymbol{g})-Q_{\boldsymbol{h}}^C(\boldsymbol{g}_C^{*})]}
{c_-\mu_h}.
\end{equation}

The lower comparison bound ensures that the weighted objective remains sensitive to changes in the expert mixture. The upper bound controls smoothness but is not needed for recovery. If $\boldsymbol{C}_h$ represents a local approximation to task curvature, this result applies to the quadratic surrogate. Extending it to the task loss itself requires additional control of curvature and approximation error.

\paragraph{Example 2 (separated residuals with identical task predictions).}\label{app:example-task-predictions}
The following example highlights a key limitation of residual-space separation showing that even orthogonal expert residuals can produce identical predictive distributions, leaving the routing weights unidentifiable from the task loss. Consider two-class logit residuals with $c=1$, zero base logits, and
\begin{equation}
\boldsymbol{z}_1=\begin{pmatrix}1\\0\end{pmatrix},
\qquad
\boldsymbol{z}_2=\begin{pmatrix}0\\-1\end{pmatrix}.
\end{equation}
Both have unit norm and are orthogonal, so $\omega(\boldsymbol{h})=\delta(\boldsymbol{h})=0$. For $\boldsymbol{g}=(a,1-a)^\top$, however,
\begin{equation}
\boldsymbol{Z}\boldsymbol{g}
=\begin{pmatrix}a\\a-1\end{pmatrix},
\qquad
\operatorname{softmax}(\boldsymbol{Z}\boldsymbol{g})
=\begin{pmatrix}\frac{e}{e+1}\\[2pt]\frac{1}{e+1}\end{pmatrix}.
\end{equation}
Different values of $a$ produce different logits, but their probabilities remain unchanged because softmax is invariant to adding the same constant to every logit. And cross-entropy loss for any fixed label is constant in $a$ and cannot distinguish the routing weights. There is no contradiction with Theorem~\ref{thm:routing-recovery}, which concerns the Euclidean residual objective. The cross-entropy Hessian has zero curvature along this routing-induced direction and therefore fails the positive-curvature assumption needed to transfer the recovery argument to the task loss.

\subsection{Population guarantees for output decorrelation}
\label{app:population-decorrelation}
The routing theorem applies at a fixed input, whereas OD-MoLE averages its output penalty over sampled activations. To connect these settings, fix expert parameters and consider an activation distribution under which all expert outputs are nonzero with probability one, so the cosine similarities are well-defined. Then $\omega(\boldsymbol{h})$ is defined, and Markov's inequality gives, for $0<\zeta<1$,
\begin{equation}
\Pr_{\boldsymbol{h}}[\delta(\boldsymbol{h})>\zeta]=\Pr_{\boldsymbol{h}}[\omega(\boldsymbol{h})>\zeta^2/(N(N-1))]\leq\frac{N(N-1)\mathbb{E}_{\boldsymbol{h}}\left[\omega(\boldsymbol{h})\right]}{\zeta^2}.
\end{equation}
Therefore, a small expected penalty limits the probability of violating the correlation threshold. On inputs satisfying both $\delta(\boldsymbol{h})\leq\zeta$ and $\min_i\|\boldsymbol{z}_i(\boldsymbol{h})\|_2\geq\underline{z}>0$, the recovery constant is at least $c^2\underline{z}^2(1-\zeta)$. The expected penalty alone does not enforce the output-norm condition.

\subsection{Proof of Eq.~\ref{eq:expert_differences}}
\label{app:expert-differences-proof}

\begin{proof}
We next examine how the shared representation affects differences between expert outputs. Let $\boldsymbol{D}_{ij}=\boldsymbol{B}_i-\boldsymbol{B}_j$. By linearity of expectation and the trace identity, we have
\begin{equation}
\begin{aligned}
\mathbb{E}_{\boldsymbol{h}}\|\boldsymbol{z}_i(\boldsymbol{h})-\boldsymbol{z}_j({\boldsymbol{h}})\|_2^2=\mathbb{E}_{\boldsymbol{h}}\|\boldsymbol{D}_{ij}\boldsymbol{A}\boldsymbol{h}\|_2^2=\operatorname{tr}(\boldsymbol{D}_{ij}\boldsymbol{K}_{\boldsymbol{A}}
\boldsymbol{D}_{ij}^\top)\\
\geq\lambda_{\min}(\boldsymbol{K}_{\boldsymbol{A}})
\operatorname{tr}(\boldsymbol{D}_{ij}\boldsymbol{D}_{ij}^\top)=\lambda_{\min}(\boldsymbol{K}_{\boldsymbol{A}})
\|\boldsymbol{B}_i-\boldsymbol{B}_j\|_{\mathrm F}^2.
\end{aligned}
\end{equation}
The inequality follows from $\boldsymbol{K}_{\boldsymbol{A}}\succeq\lambda_{\min}(\boldsymbol{K}_{\boldsymbol{A}})
\boldsymbol{I}$.
\end{proof}
Thus, a positive lower bound on $\lambda_{\min}(\boldsymbol{K}_{\boldsymbol{A}})$ ensures that every nonzero reconstruction difference remains observable in mean-square output. The next theorem provides such a bound. For scaled expert residuals $c\boldsymbol{z}_i(\boldsymbol{h})$, the corresponding lower bound includes an additional factor $c^2$.

\subsection{Proof of Theorem~\ref{thm:shared-diversity}}
\label{app:shared-diversity-proof}

\begin{proof}
Let $\boldsymbol{P}=\boldsymbol{A}\boldsymbol{A}^\top$ with eigenvalues $0<\lambda_1\leq\cdots\leq\lambda_r$. The trace and log-determinant assumptions imply $\sum_j\lambda_j=r$ and $\prod_j\lambda_j\geq e^{-\eta}$. By the arithmetic--geometric mean inequality,
\begin{equation}
e^{-\eta}
\leq\lambda_1\prod_{j=2}^r\lambda_j\leq\lambda_1\left(\frac{r-\lambda_1}{r-1}\right)^{r-1}
\leq\lambda_1\left(\frac{r}{r-1}\right)^{r-1}.
\end{equation}
Therefore,
\begin{equation}
\lambda_{\min}(\boldsymbol{P})
\geq e^{-\eta}\left(\frac{r-1}{r}\right)^{r-1}.
\end{equation}
For any $\boldsymbol{w}\in\mathbb R^r$, $\boldsymbol{A}^\top\boldsymbol{w}$ belongs to the row space of $\boldsymbol{A}$. Applying the activation assumption gives
\begin{equation}
\boldsymbol{w}^\top\boldsymbol{K}_{\boldsymbol{A}}\boldsymbol{w}=(\boldsymbol{A}^\top\boldsymbol{w})^\top
\boldsymbol{\Sigma}_{\boldsymbol{h}}(\boldsymbol{A}^\top\boldsymbol{w})\geq\nu\|\boldsymbol{A}^\top\boldsymbol{w}\|_2^2
=\nu\boldsymbol{w}^\top\boldsymbol{P}\boldsymbol{w}.
\end{equation}
Hence $\boldsymbol{K}_{\boldsymbol{A}}\succeq\nu\boldsymbol{P}$, which proves Eq.~\ref{eq:shared-diversity}.
\end{proof}

This highlights how projection diversity plays a supporting role by maintaining the output behavior of the original reconstructions whenever inputs activate those primary directions.

\paragraph{Role of the assumptions.}
The trace controls the sum of squared singular values of $\boldsymbol{A}$, while the determinant controls their product. Eigenvalues $(\varepsilon,\varepsilon^{-1})$ have determinant one, whereas $(\varepsilon,2-\varepsilon)$ have trace two, but both have a vanishing minimum as $\varepsilon\to0$. The log-determinant assumption also requires full row rank, hence $r\leq k$. The input assumption is needed only on the row space of $\boldsymbol{A}$. It ensures nonzero second-moment energy in the retained directions. Conditioning $\boldsymbol{A}$ cannot supply signal absent from the data. When $\boldsymbol{A}$ changes during training, the condition must be checked for its resulting row space. At $\eta=0$, trace normalization forces $\boldsymbol{P}=\boldsymbol{I}$, giving the sharper bound $\lambda_{\min}(\boldsymbol{K}_{\boldsymbol{A}})\geq\nu$. The same conclusion holds for $r=1$ under trace normalization.

\paragraph{Example 3 (pairwise expert differences with an ambiguous mixture).}\label{app:example-ambiguous-mixture}
We provide a simple example that demonstrates while two routing assignments can yield identical mixture outputs, pairwise expert separation alone is insufficient to guarantee unique routing recovery. Suppose the assumptions of Theorem~\ref{thm:shared-diversity} hold. Choose $\boldsymbol{B}_1\neq\boldsymbol{B}_2$ and set
\begin{equation}
\boldsymbol{B}_3=\frac{1}{2}
(\boldsymbol{B}_1+\boldsymbol{B}_2).
\end{equation}
All three matrices are pairwise different. Eq.~\ref{eq:expert_differences} and the positive lower bound on $\lambda_{\min}(\boldsymbol{K}_{\boldsymbol{A}})$ therefore imply positive mean-square output differences for every pair. Yet linearity gives, for every activation,
\begin{equation}
\boldsymbol{z}_3(\boldsymbol{h})=\frac{1}{2}\boldsymbol{z}_1(\boldsymbol{h})
+\frac{1}{2}\boldsymbol{z}_2(\boldsymbol{h}).
\end{equation}
Consequently, the distinct routing vectors $\boldsymbol{g}=(\frac{1}{2},\frac{1}{2},0)^\top$ and $\widehat{\boldsymbol{g}}=(0,0,1)^\top$ satisfy $c\boldsymbol{Z}(\boldsymbol{h})\boldsymbol{g}
=c\boldsymbol{Z}(\boldsymbol{h})\widehat{\boldsymbol{g}}$. This illustrates why preserving pairwise expert differences does not ensure routing recovery. The shared-projection result preserves existing differences in mean square, while routing recovery requires separation of all nonzero zero-sum combinations of expert outputs.

Theorem~\ref{thm:shared-diversity} assumes $\text{tr}(\boldsymbol{A}\boldsymbol{A}^\top)=\|\boldsymbol{A}\|_{\mathrm{F}}^2=r$. In our experiments with the raw log-determinant penalty, the learned projection $\boldsymbol{A}$ is unconstrained in scale. The following corollary extends the bound to full-row-rank projections of arbitrary scale and separates the contributions of scale and spectral shape.

\begin{corollary}[Arbitrary-scale form of Theorem~\ref{thm:shared-diversity}]
\label{cor:shared-diversity-scale}
Suppose $r>1$ and $\boldsymbol{A}$ has full row rank, and let the input assumption of Theorem~\ref{thm:shared-diversity} hold. Define $\bar{\boldsymbol{A}}=\sqrt{r}\,\boldsymbol{A}/\|\boldsymbol{A}\|_{\mathrm{F}}$ and $\bar{\eta}=-\log\det(\bar{\boldsymbol{A}}\bar{\boldsymbol{A}}^\top)\geq0$. Then
\begin{equation}
\lambda_{\min}(\boldsymbol{K}_{\boldsymbol{A}})
\;\geq\;\nu\,\det(\boldsymbol{A}\boldsymbol{A}^\top)\left(\frac{r-1}{\|\boldsymbol{A}\|_{\mathrm{F}}^2}\right)^{r-1}
=\;\nu\,\frac{\|\boldsymbol{A}\|_{\mathrm{F}}^2}{r}\;e^{-\bar{\eta}}\left(\frac{r-1}{r}\right)^{r-1}.
\label{eq:shared-diversity-scale}
\end{equation}
\end{corollary}
\begin{proof}
Let $\mu_1\leq\dots\leq\mu_r$ be the eigenvalues of $\boldsymbol{A}\boldsymbol{A}^\top$, which are positive since $\boldsymbol{A}$ has full row rank. As in the proof of Theorem~\ref{thm:shared-diversity}, the input assumption gives $\lambda_{\min}(\boldsymbol{K}_{\boldsymbol{A}})\geq\nu\mu_1$. By the AM--GM inequality,
$\prod_{i=2}^{r}\mu_i\leq\big(\tfrac{1}{r-1}\sum_{i=2}^{r}\mu_i\big)^{r-1}\leq\big(\|\boldsymbol{A}\|_{\mathrm{F}}^2/(r-1)\big)^{r-1}$,
so $\mu_1=\det(\boldsymbol{A}\boldsymbol{A}^\top)/\prod_{i=2}^{r}\mu_i\geq\det(\boldsymbol{A}\boldsymbol{A}^\top)\big((r-1)/\|\boldsymbol{A}\|_{\mathrm{F}}^2\big)^{r-1}$. The second form follows from $\det(\boldsymbol{A}\boldsymbol{A}^\top)=(\|\boldsymbol{A}\|_{\mathrm{F}}^2/r)^r\det(\bar{\boldsymbol{A}}\bar{\boldsymbol{A}}^\top)$, and $\bar{\eta}\geq0$ because $\text{tr}(\bar{\boldsymbol{A}}\bar{\boldsymbol{A}}^\top)=r$ implies $\det(\bar{\boldsymbol{A}}\bar{\boldsymbol{A}}^\top)\leq1$ by the AM--GM inequality.
\end{proof}
The bound factors into a scale term $\|\boldsymbol{A}\|_{\mathrm{F}}^2/r$ and a shape term that depends only on the normalized matrix $\bar{\boldsymbol{A}}$, and it reduces to Eq.~\ref{eq:shared-diversity} when $\|\boldsymbol{A}\|_{\mathrm{F}}^2=r$. The raw log-determinant can therefore increase through either term, but only the shape term reflects how evenly $\boldsymbol{A}$ spreads its energy across directions, since a uniform rescaling of $\boldsymbol{A}$ can be absorbed by the expert matrices.

\subsection{Connection to the sampled diversity regularizer}
\label{app:sampled-determinants}
Theorem~\ref{thm:shared-diversity} uses the determinant of the full projection Gram matrix, whereas the regularizer uses sampled columns. We now relate this condition to the penalty evaluated on sampled columns of $\boldsymbol{A}$, keeping the projection fixed during the sampling analysis.
\paragraph{Sampled determinants.}
Assume $r\leq k$, and let $\mathcal{S}$ be sampled uniformly from all sets of exactly $r$ columns. Then $\boldsymbol{A}_{\mathcal{S}}\in\mathbb{R}^{r\times r}$, and the Cauchy--Binet formula gives
\begin{equation}
\det(\boldsymbol{A}\boldsymbol{A}^\top)
=\sum_{|\mathcal{S}|=r}\det(\boldsymbol{A}_{\mathcal{S}})^2
=\sum_{|\mathcal{S}|=r}
\det(\boldsymbol{A}_{\mathcal{S}}^\top\boldsymbol{A}_{\mathcal{S}}).
\end{equation}
Under uniform sampling of $r$ columns, this equals $\binom{k}{r}\mathbb E_{\mathcal{S}}\left[\det(\boldsymbol{A}_{\mathcal{S}}^\top\boldsymbol{A}_{\mathcal{S}})\right]$. Taking logarithms and applying Jensen's inequality gives
\begin{equation}
\log\det(\boldsymbol{A}\boldsymbol{A}^\top)
\geq\log\binom{k}{r}
+\mathbb E_{\mathcal{S}}\left[\log\det
(\boldsymbol{A}_{\mathcal{S}}^\top\boldsymbol{A}_{\mathcal{S}})\right].
\end{equation}
We note that if any sampled matrix has a determinant of zero, the expected log determinant is $-\infty$ under uniform sampling, and the inequality has no finite lower bound.

\paragraph{Finite-$\epsilon$ regularization.}
For the implemented penalty, define its expectation over column sampling as
\begin{equation}
    D_\epsilon=-\mathbb E_{\mathcal{S}}\left[\log\det
(\boldsymbol{A}_{\mathcal{S}}^\top\boldsymbol{A}_{\mathcal{S}}
+\epsilon\boldsymbol{I})\right],\quad \epsilon>0
\end{equation}
Since $\boldsymbol{A}\boldsymbol{A}^\top\succeq\boldsymbol{A}_{\mathcal{S}}\boldsymbol{A}_{\mathcal{S}}^\top$ and $\boldsymbol{A}_{\mathcal{S}}$ is square, we have
\begin{equation}
\log\det(\boldsymbol{A}\boldsymbol{A}^\top+\epsilon\boldsymbol{I})
\geq-D_\epsilon.
\end{equation}
If $\operatorname{tr}(\boldsymbol{A}\boldsymbol{A}^\top)\leq T_A$ and $r>1$, we have
\begin{equation}
\lambda_{\min}(\boldsymbol{A}\boldsymbol{A}^\top)
\geq\left[
e^{-D_\epsilon}
\left(\frac{r-1}{T_A+(r-1)\epsilon}\right)^{r-1}
-\epsilon\right]_+,
\label{eq:app-regularized-spectrum}
\end{equation}
where $[x]_+=\max(x,0)$. Under the activation assumption of Theorem~\ref{thm:shared-diversity}, $\lambda_{\min}(\boldsymbol{K}_{\boldsymbol{A}})\geq \nu\lambda_{\min}(\boldsymbol{A}\boldsymbol{A}^{\top})$. Thus, multiplying $\lambda_{\min}(\boldsymbol{A}\boldsymbol{A}^\top)$ by $\nu$ gives a lower bound on $\lambda_{\min}(\boldsymbol{K}_{\boldsymbol{A}})$.

\paragraph{Scale control.}
For $s>0$, the transformation $\boldsymbol{A}\mapsto s\boldsymbol{A}$ and $\boldsymbol{B}_i\mapsto\boldsymbol{B}_i/s$ keep every expert output unchanged. If a sampled submatrix has positive singular values $\sigma_1,\ldots,\sigma_q$, with $q>0$, its negative log determinant becomes
\begin{equation}
-\sum_{j=1}^{q}\log(s^2\sigma_j^2+\epsilon)
-(r-q)\log\epsilon\ \longrightarrow\ -\infty
\quad\text{as }s\to\infty.
\end{equation}
Thus, the diversity penalty can decrease through rescaling without changing the expert output. A scale constraint or suitable norm control is therefore needed to connect the penalty to preservation of latent directions. Theorem~\ref{thm:shared-diversity} provides this connection under its stated scale and input assumptions, minimizing the log-determinant penalty alone does not enforce them.

\section{Experiment Details}
\label{app:experiment-details}

\subsection{Datasets}
\label{app:datasets}
We summarize the details of the used datasets as follows:
\begin{itemize}
    \item CommonsenseQA (ComQA)~\citep{talmor-etal-2019-commonsenseqa}: it challenges the model's grasp of everyday knowledge.
    \item CosmosQA (CosQA)~\citep{huang-etal-2019-cosmos}: it tests inferential reasoning through narrative comprehension.
    \item SocialIQA (SIQA)~\citep{sap-etal-2019-social}: it examines the model's understanding of social interactions by asking questions about motivations, reactions, or outcomes in social contexts.
    \item PhysicalIQA (PIQA)~\citep{bisk2020piqa}: it examines the model's understanding of physical phenomena to select the most plausible solution to a given problem.
    \item ScienceQA (ScQA)~\citep{lu2022learn}: it examines the model's understanding of scientific principles.
    \item BoolQ~\citep{clark-etal-2019-boolq}: it's a QA task where each example consists of a short passage and a yes/no question about the passage, which is to assess the model's ability to evaluate the veracity of statements.
    \item ARC-Challenge (ARC-C)~\citep{clark2018think}: it's a more difficult subset of ARC, containing questions that require advanced reasoning and knowledge.
    \item OpenBookQA (OBQA)~\citep{mihaylov-etal-2018-suit}: it's a question-answering task requiring reasoning and knowledge from a small "open book" of science facts.
     \item COPA~\citep{roemmele2011choice}: it's a causal reasoning task where a system is given a premise sentence and must determine either the cause or the effect of the premise from two possible choices.

\end{itemize}
\subsection{Baselines}
\label{app:baselines}
\begin{itemize}
    \item LoRA~\citep{hu2022lora}: it utilizes trainable low-rank matrices for parameter-efficient fine-tuning. We change the rank from 8 to 32.
    \item PiSSA~\citep{meng2024pissa}: it initializes LoRA matrices with the principal components of the pre-trained weights.
    \item HydraLoRA~\citep{tian2024hydralora}: it proposes an asymmetric LoRA framework that improves parameter efficiency and performance by addressing training inefficiencies.
    \item MoLoRA~\citep{zadouri2024pushing}: it combines MoE architecture with lightweight experts, enabling extremely parameter-efficient fine-tuning by updating less than 1\% of model parameters.
    \item AdaMoLE~\citep{liu2024adamole}: it uses adaptive mechanisms to optimize the selection of experts.
    \item GOAT~\citep{fan2025make}: it adaptively integrates pre-trained knowledge using an SVD-structured MoE framework, addressing the limitations of non-informative or static priors.
\end{itemize}

\subsection{Dataset-Specific Training Schedules}
\label{app:training-schedules}
Different commonsense reasoning datasets exhibit varying levels of complexity and convergence behavior. To ensure stable and fair evaluation, we adopt dataset-specific training schedules rather than using a fixed number of epochs across all tasks. 
Tab.~\ref{tab:training_epochs} summarizes the number of training epochs used for each dataset. Importantly, we select the training epochs based on validation performance to avoid underfitting or overfitting across tasks. This design ensures that all compared methods are evaluated under comparable convergence conditions, allowing for a fair assessment of model capability rather than training duration. We note that using a uniform training schedule across datasets leads to suboptimal performance for harder tasks and premature convergence for easier ones, highlighting the importance of task-aware training budgets in multi-dataset evaluation.

\begin{table}[t]
    \centering
    \caption{
    Dataset-specific training schedules used in our experiments.}
    \label{tab:training_epochs}
    \begin{tabular}{c|c}
    \toprule
    \textbf{Dataset} & \textbf{\#Epochs} \\
    \midrule
    ComQA   & 2 \\
    CosQA   & 2 \\
    SIQA    & 2 \\
    PIQA    & 3 \\
    ScQA    & 5 \\
    BoolQ   & 2 \\
    OBQA    & 2 \\
    ARC-C   & 3 \\
    \bottomrule
    \end{tabular}
\end{table}

\subsection{Sensitivity to Regularization Coefficients}
\label{app:regularization-sensitivity}

\begin{table}[t]
\centering
\footnotesize
\caption{Sensitivity of OD-MoLE to $\lambda_{\mathrm{orth}}$ and $\lambda_{\mathrm{div}}$ coefficients on COPA using Qwen2.5-7B. Results are reported as accuracy under different coefficient combinations in the expert composition setting (2 experts per submodule).}
\label{tab:lambda_sensitivity}
\begin{tabular}{c|ccc}
\toprule
$\lambda_{\mathrm{orth}} \backslash \lambda_{\mathrm{div}}$ 
 &$10^{-5}$ & $10^{-4}$ & $10^{-3}$ \\
\midrule
$10^{-5}$ &  0.95 & 0.94 & 0.94 \\
$10^{-4}$ & 0.95 & 0.95 & 0.95 \\
$10^{-3}$ &  0.96 & 0.96 & 0.96 \\
\bottomrule
\end{tabular}
\end{table}

We analyze the sensitivity of OD-MoLE to the orthogonality and diversity coefficients, $\lambda_{\mathrm{orth}}$ and $\lambda_{\mathrm{div}}$. We vary one coefficient while fixing the other to $10^{-3}$ and report the results using Qwen2.5-7B on the COPA dataset in Tab.~\ref{tab:lambda_sensitivity}. We use two experts per submodule and compose their outputs during training. As the coefficients increase from $10^{-5}$ to $10^{-3}$, performance improves, suggesting that moderate regularization effectively encourages expert diversity and reduces interference. Overall, the results show that OD-MoLE is robust to a reasonable range of coefficient values, with $10^{-3}$ providing a good balance between promoting diversity and preserving model capacity.

\subsection{Expert Output Similarity Analysis}
\label{app:output-similarity}

\begin{figure}[t]
  \centering

  \begin{subfigure}{0.24\textwidth}
    \centering
    \includegraphics[width=\linewidth]{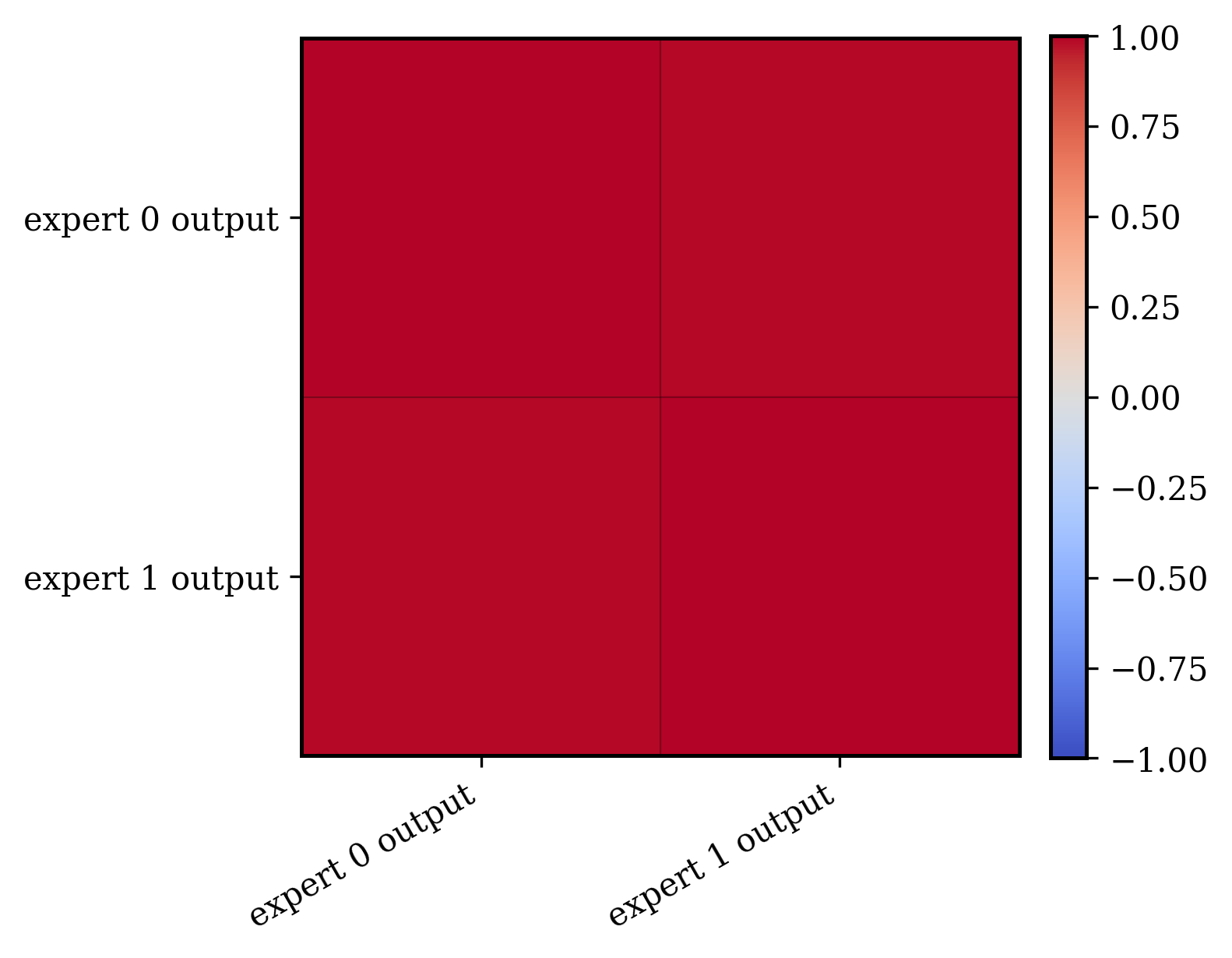}
    \caption*{HydraLoRA $q$-proj}
  \end{subfigure}
  \begin{subfigure}{0.24\textwidth}
    \centering
    \includegraphics[width=\linewidth]{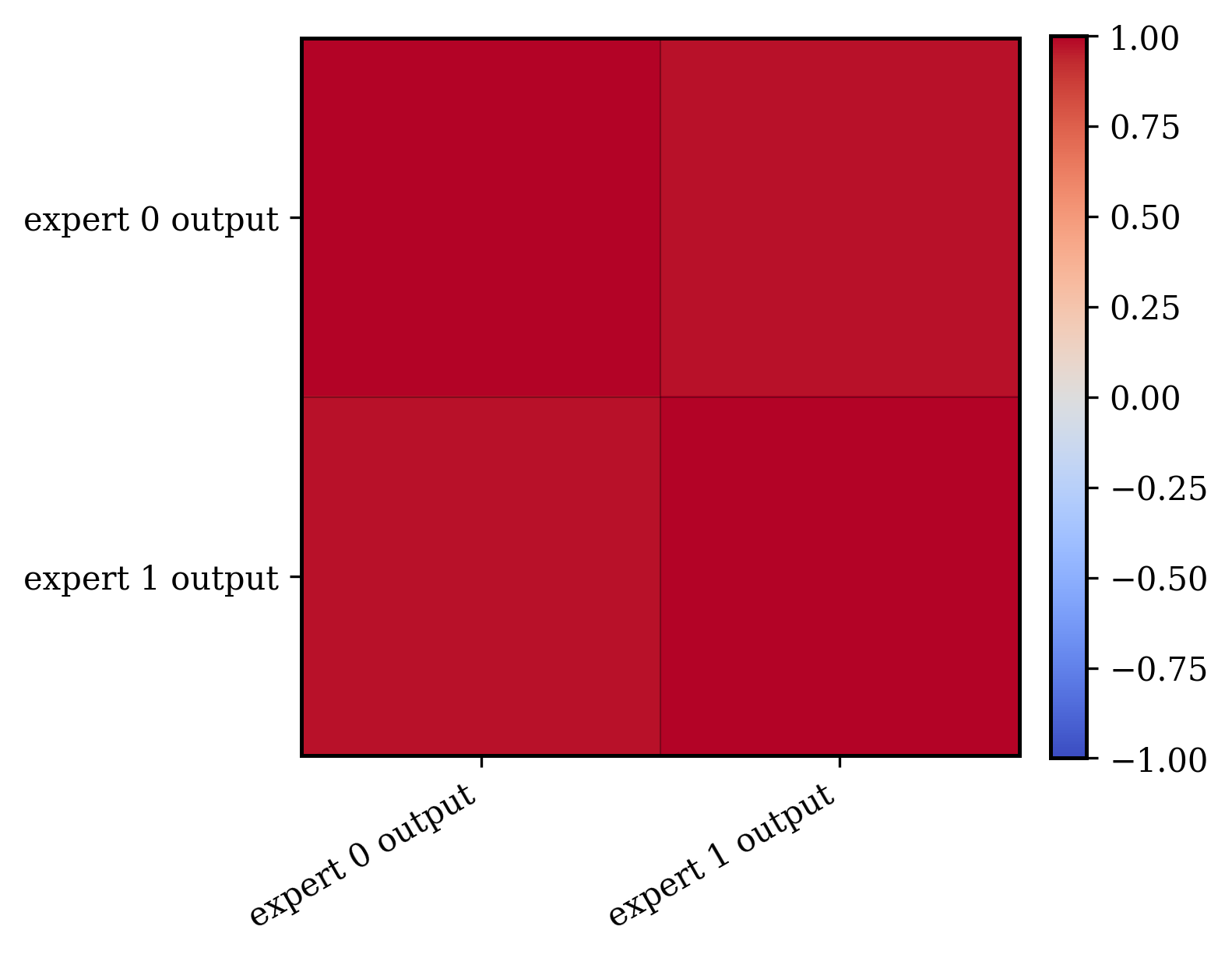}
    \caption*{HydraLoRA $k$-proj}
  \end{subfigure}
  \begin{subfigure}{0.24\textwidth}
    \centering
    \includegraphics[width=\linewidth]{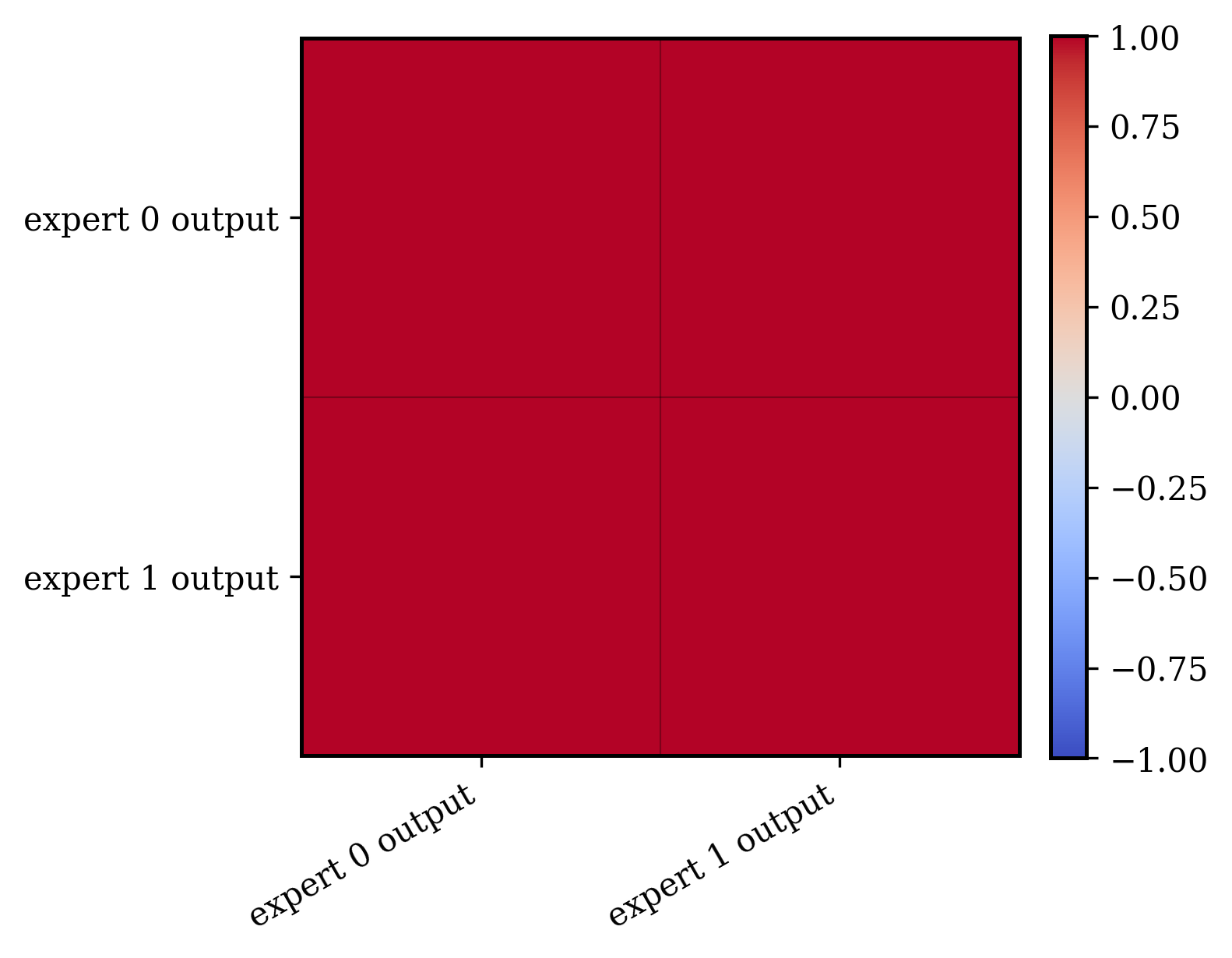}
    \caption*{HydraLoRA $v$-proj}
  \end{subfigure}
  \begin{subfigure}{0.24\textwidth}
    \centering
    \includegraphics[width=\linewidth]{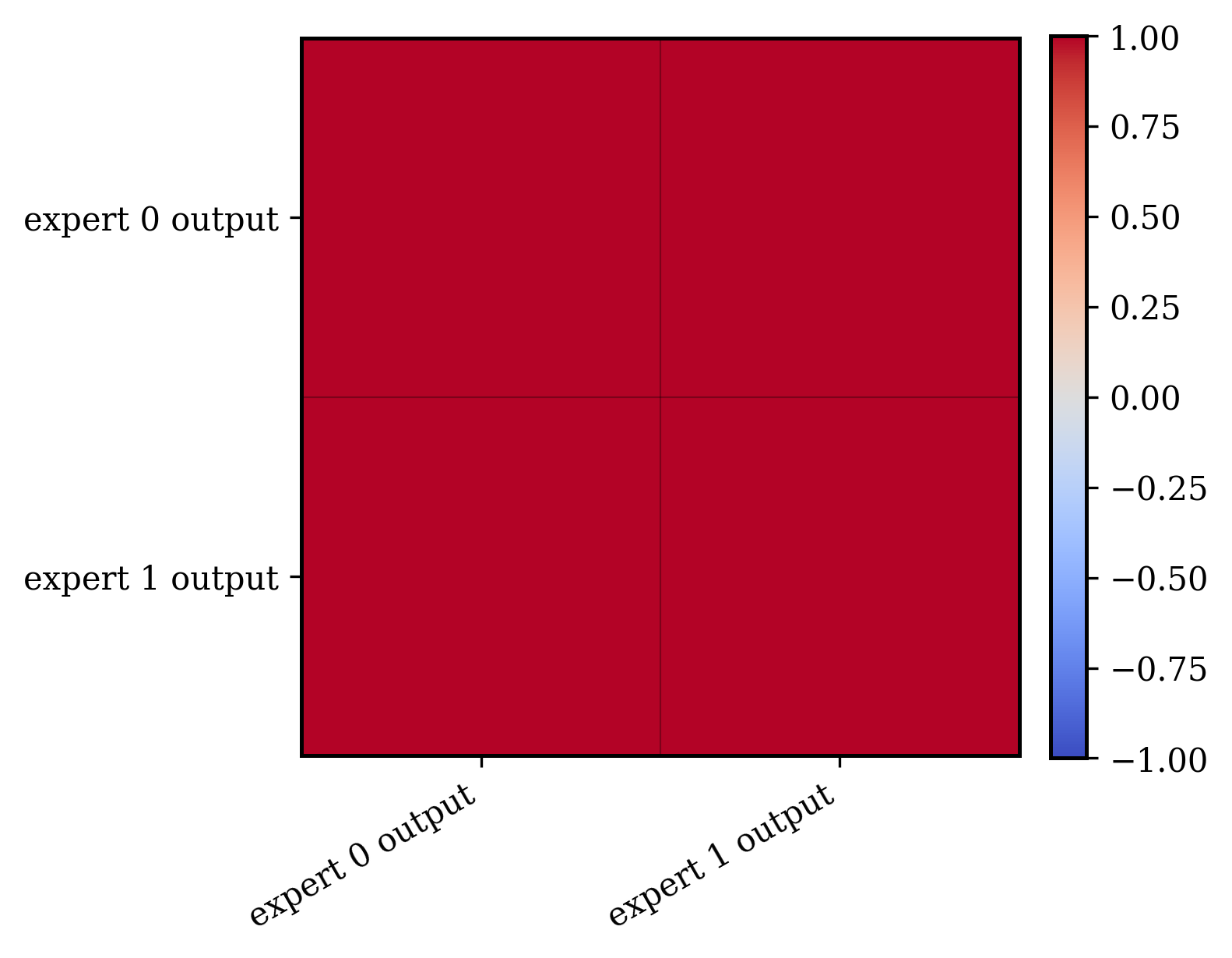}
    \caption*{HydraLoRA $o$-proj}
  \end{subfigure}

  \vspace{0.5em}

  \begin{subfigure}{0.24\textwidth}
    \centering
    \includegraphics[width=\linewidth]{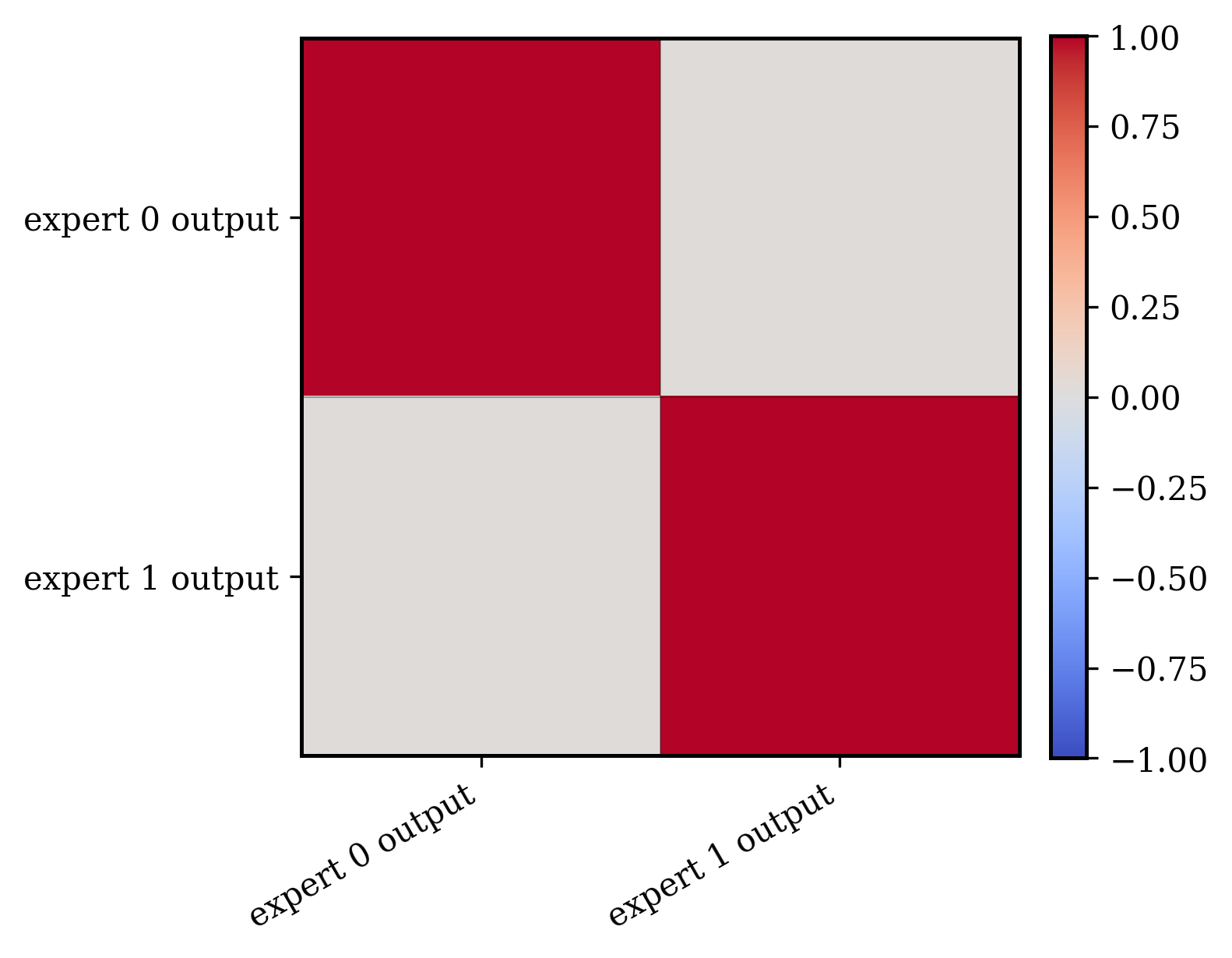}
    \caption*{OD-MoLE $q$-proj}
  \end{subfigure}
  \begin{subfigure}{0.24\textwidth}
    \centering
    \includegraphics[width=\linewidth]{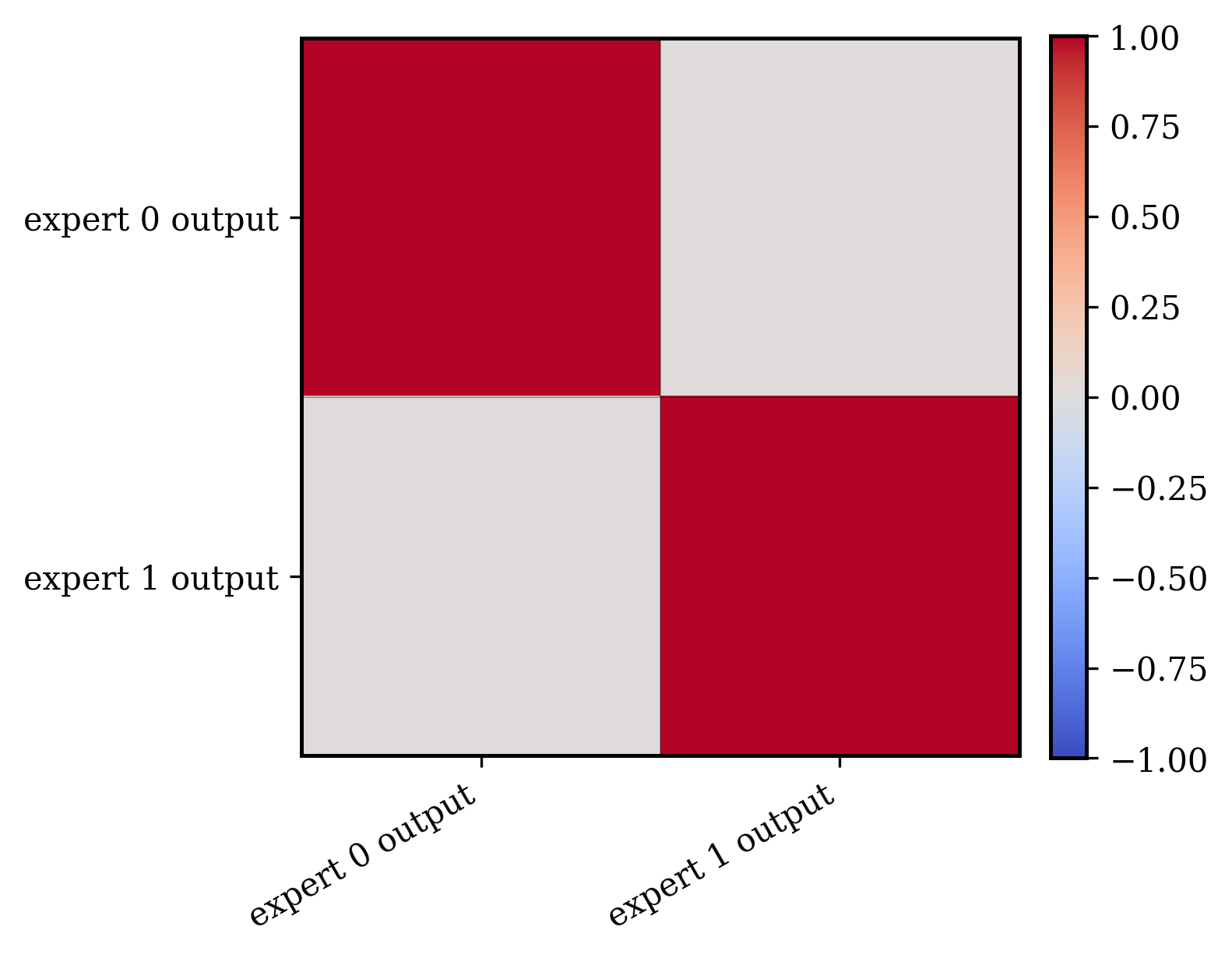}
    \caption*{OD-MoLE $k$-proj}
  \end{subfigure}
  \begin{subfigure}{0.24\textwidth}
    \centering
    \includegraphics[width=\linewidth]{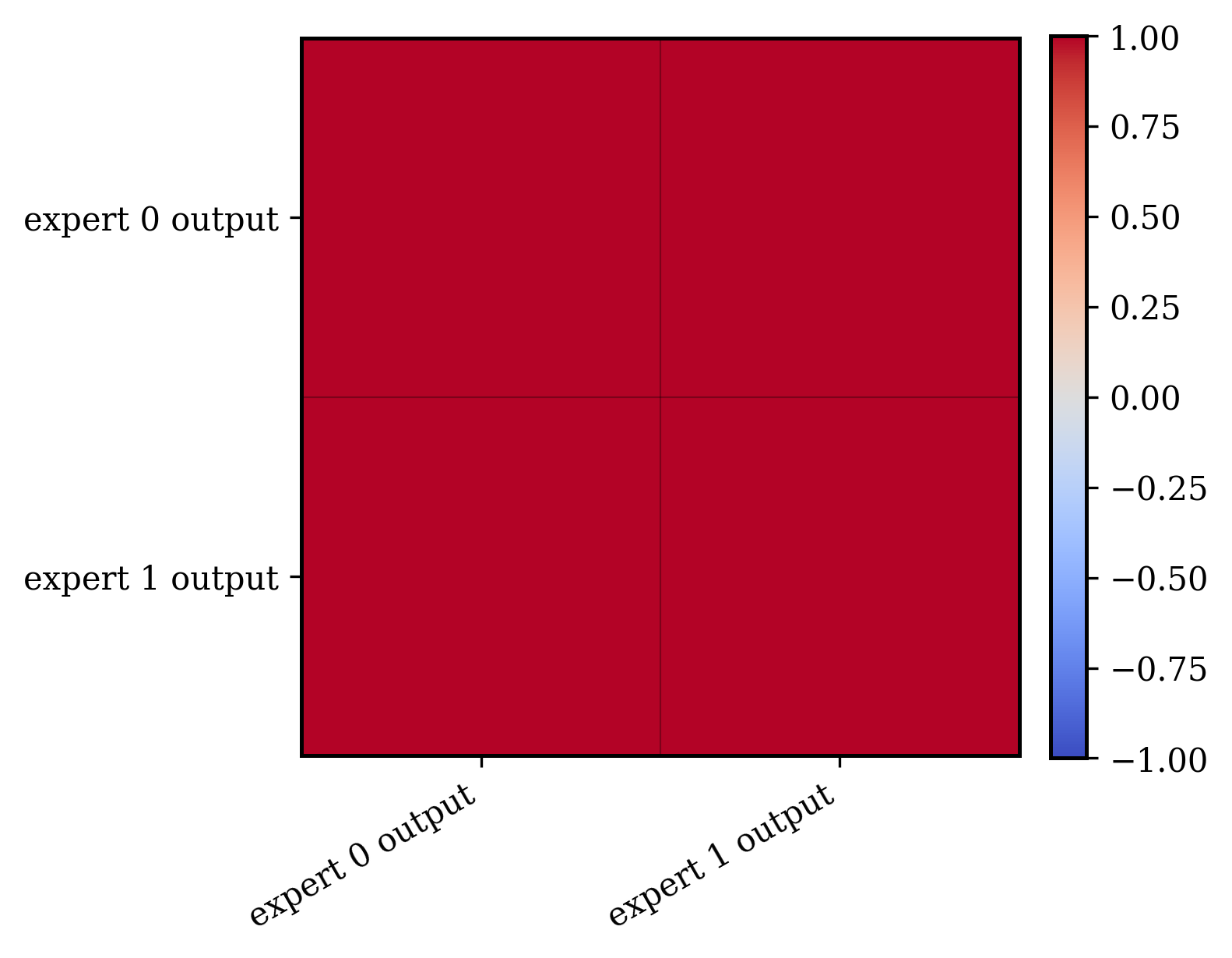}
    \caption*{OD-MoLE $v$-proj}
  \end{subfigure}
  \begin{subfigure}{0.24\textwidth}
    \centering
    \includegraphics[width=\linewidth]{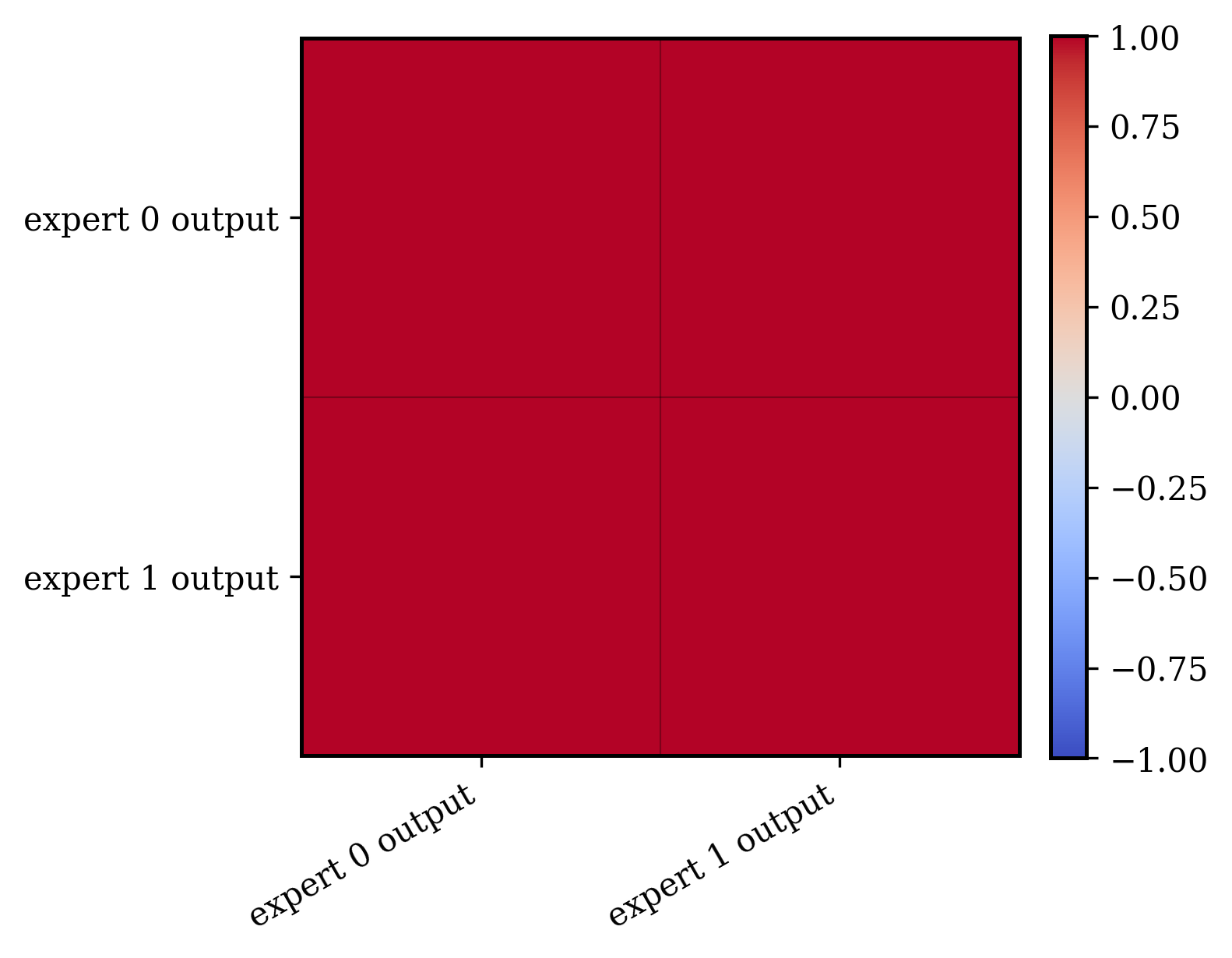}
    \caption*{OD-MoLE $o$-proj}
  \end{subfigure}

  \caption{
  Heatmap of expert output representations $\boldsymbol{B}_i\boldsymbol{A}\boldsymbol{h}$ from four attention projection modules at layer 0.
  The first row shows HydraLoRA, while the second row shows OD-MoLE.
  Compared with HydraLoRA, OD-MoLE exhibits clearer separation between expert output distributions, indicating improved expert specialization and reduced output-level interference.
  }
  \label{fig:heatmap_hydra_odmole_layer0}
\end{figure}

\begin{figure}[t]
  \centering

  \begin{subfigure}{0.24\textwidth}
    \centering
    \includegraphics[width=\linewidth]{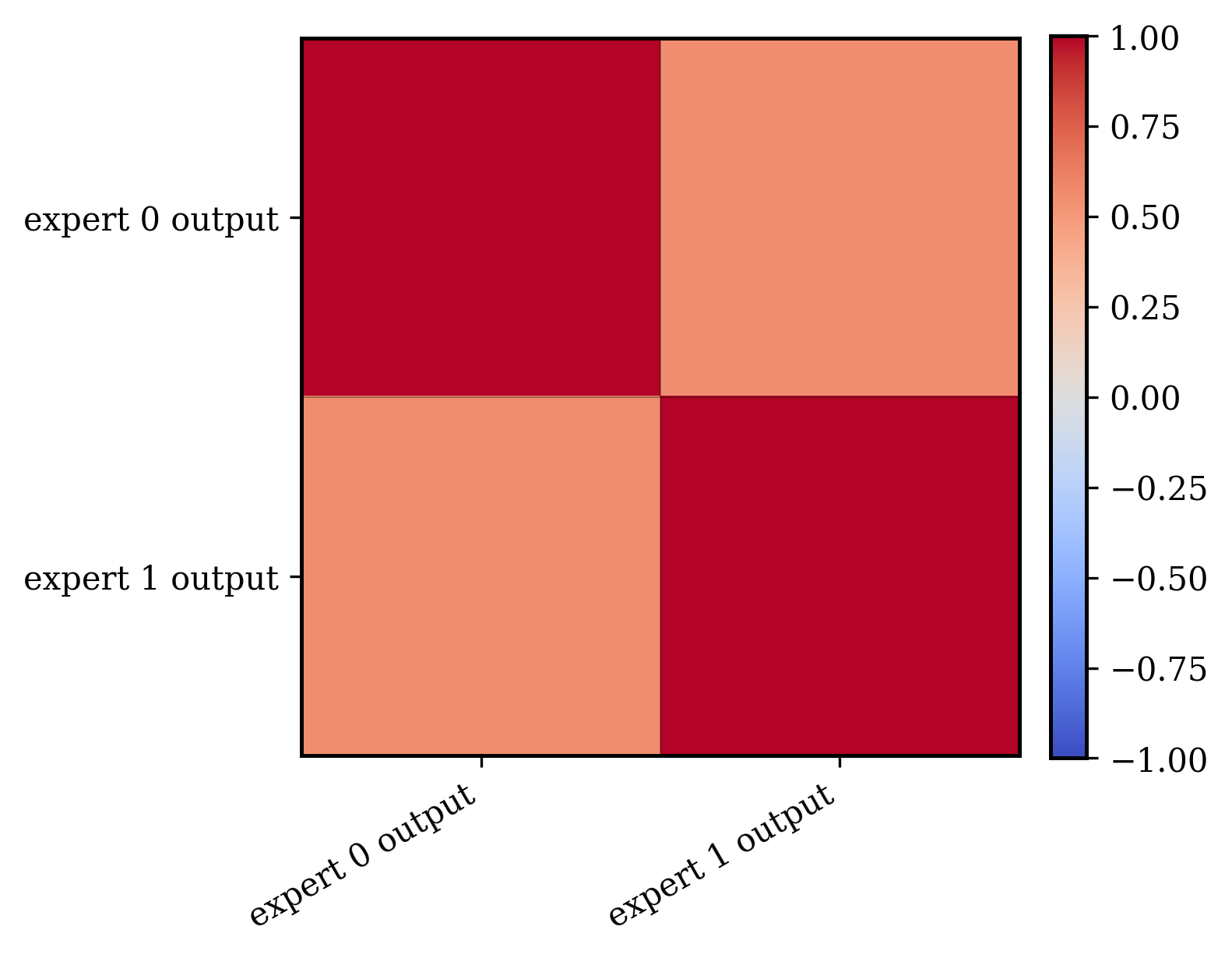}
    \caption*{HydraLoRA $q$-proj}
  \end{subfigure}
  \begin{subfigure}{0.24\textwidth}
    \centering
    \includegraphics[width=\linewidth]{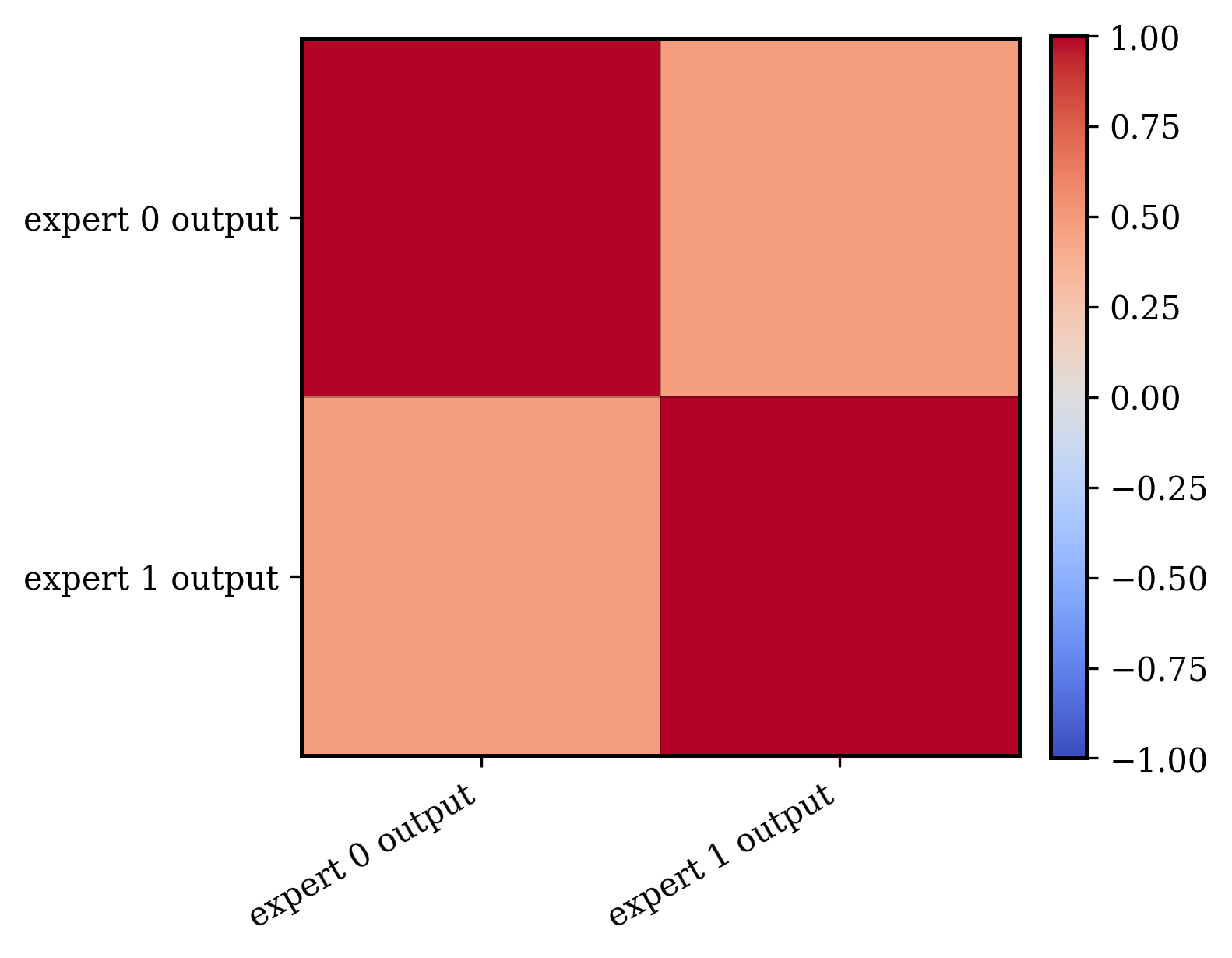}
    \caption*{HydraLoRA $k$-proj}
  \end{subfigure}
  \begin{subfigure}{0.24\textwidth}
    \centering
    \includegraphics[width=\linewidth]{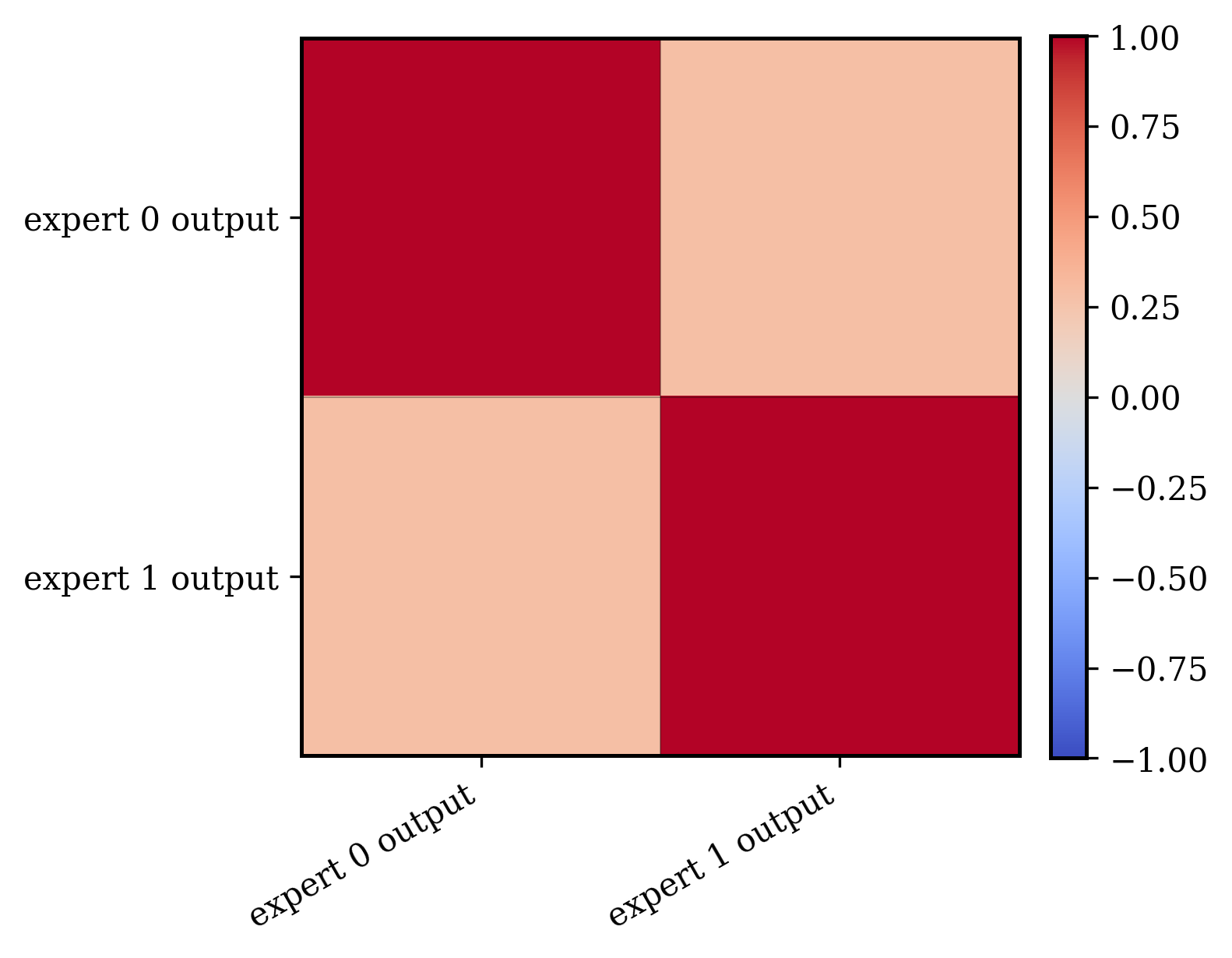}
    \caption*{HydraLoRA $v$-proj}
  \end{subfigure}
  \begin{subfigure}{0.24\textwidth}
    \centering
    \includegraphics[width=\linewidth]{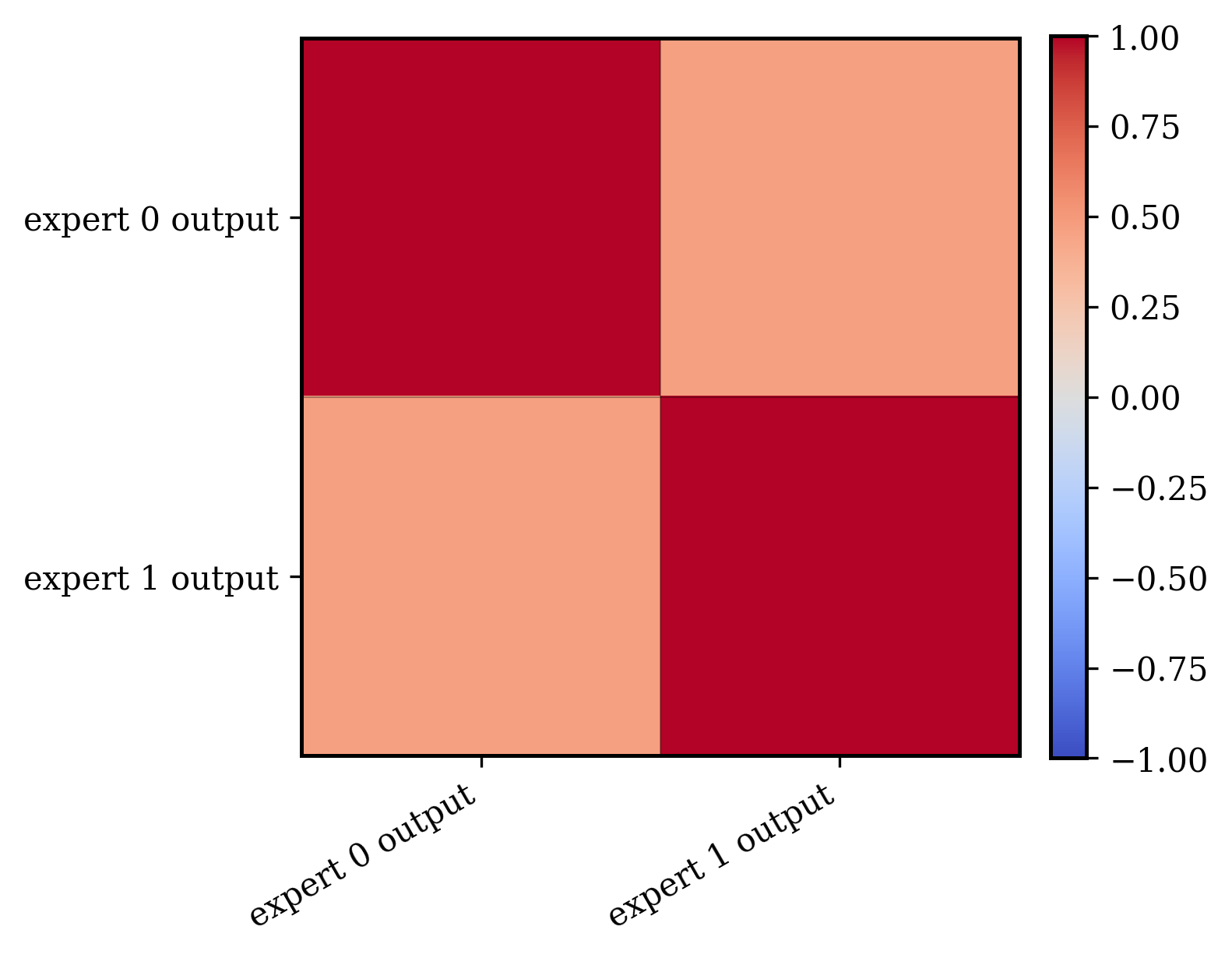}
    \caption*{HydraLoRA $o$-proj}
  \end{subfigure}

  \vspace{0.5em}

  \begin{subfigure}{0.24\textwidth}
    \centering
    \includegraphics[width=\linewidth]{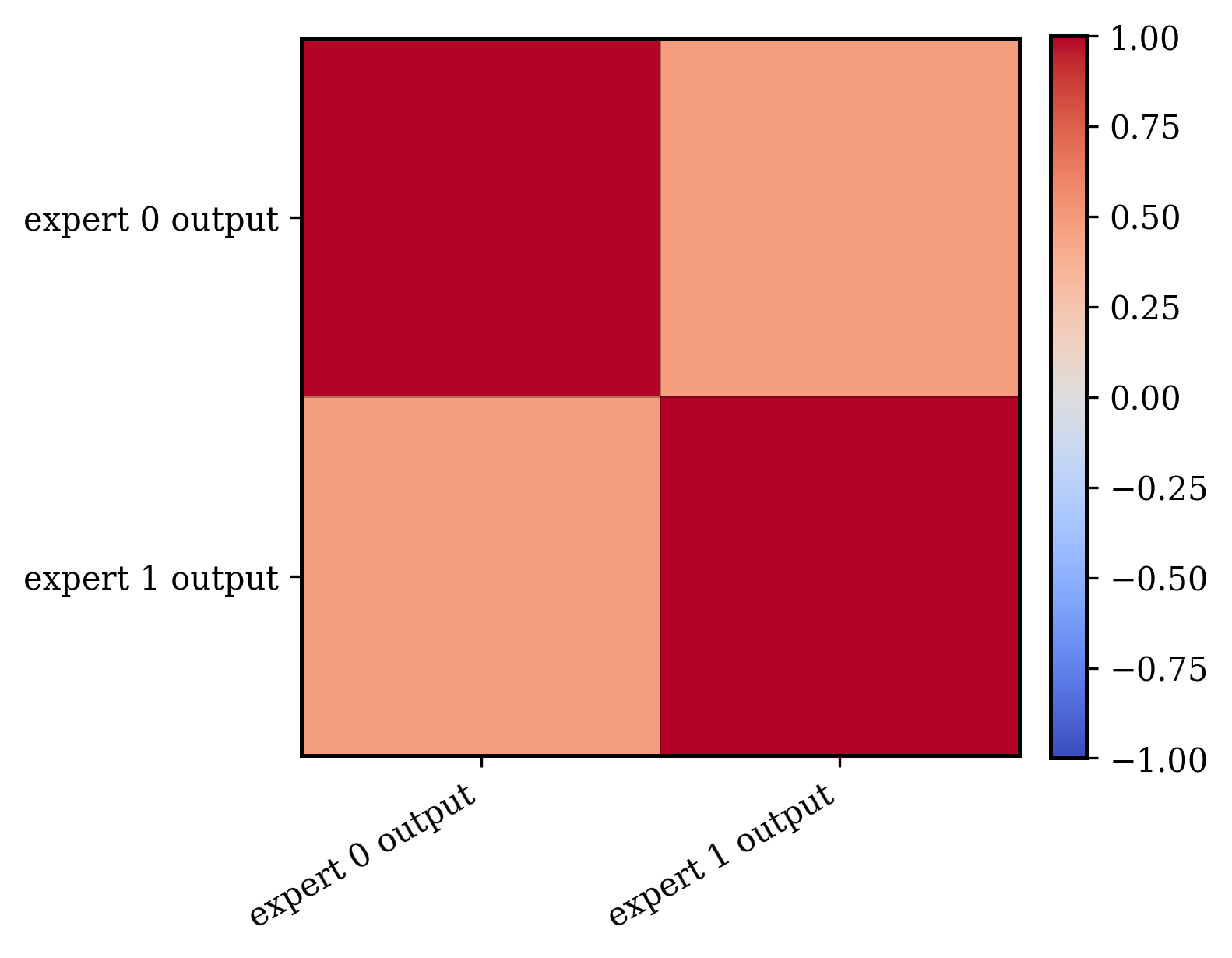}
    \caption*{OD-MoLE $q$-proj}
  \end{subfigure}
  \begin{subfigure}{0.24\textwidth}
    \centering
    \includegraphics[width=\linewidth]{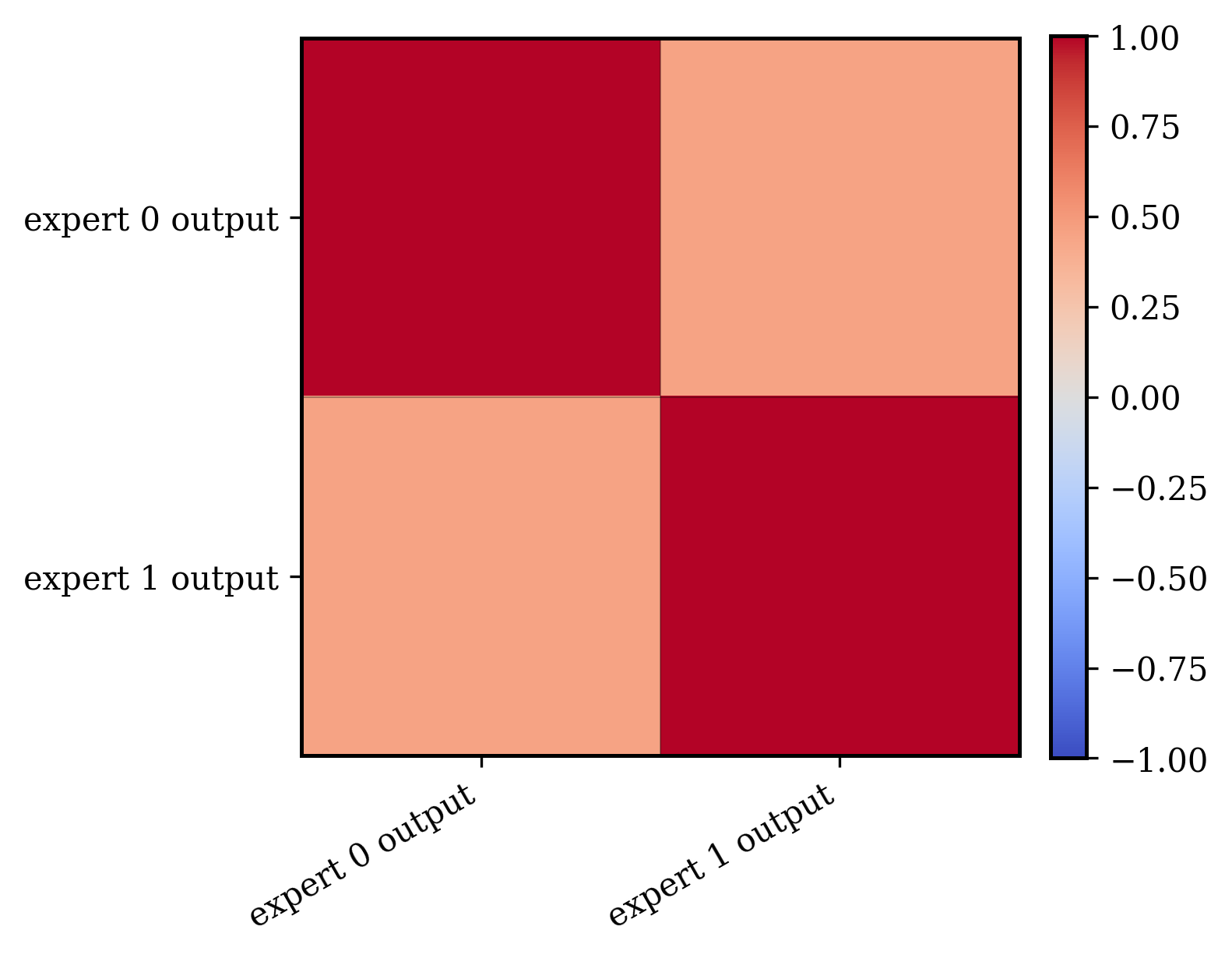}
    \caption*{OD-MoLE $k$-proj}
  \end{subfigure}
  \begin{subfigure}{0.24\textwidth}
    \centering
    \includegraphics[width=\linewidth]{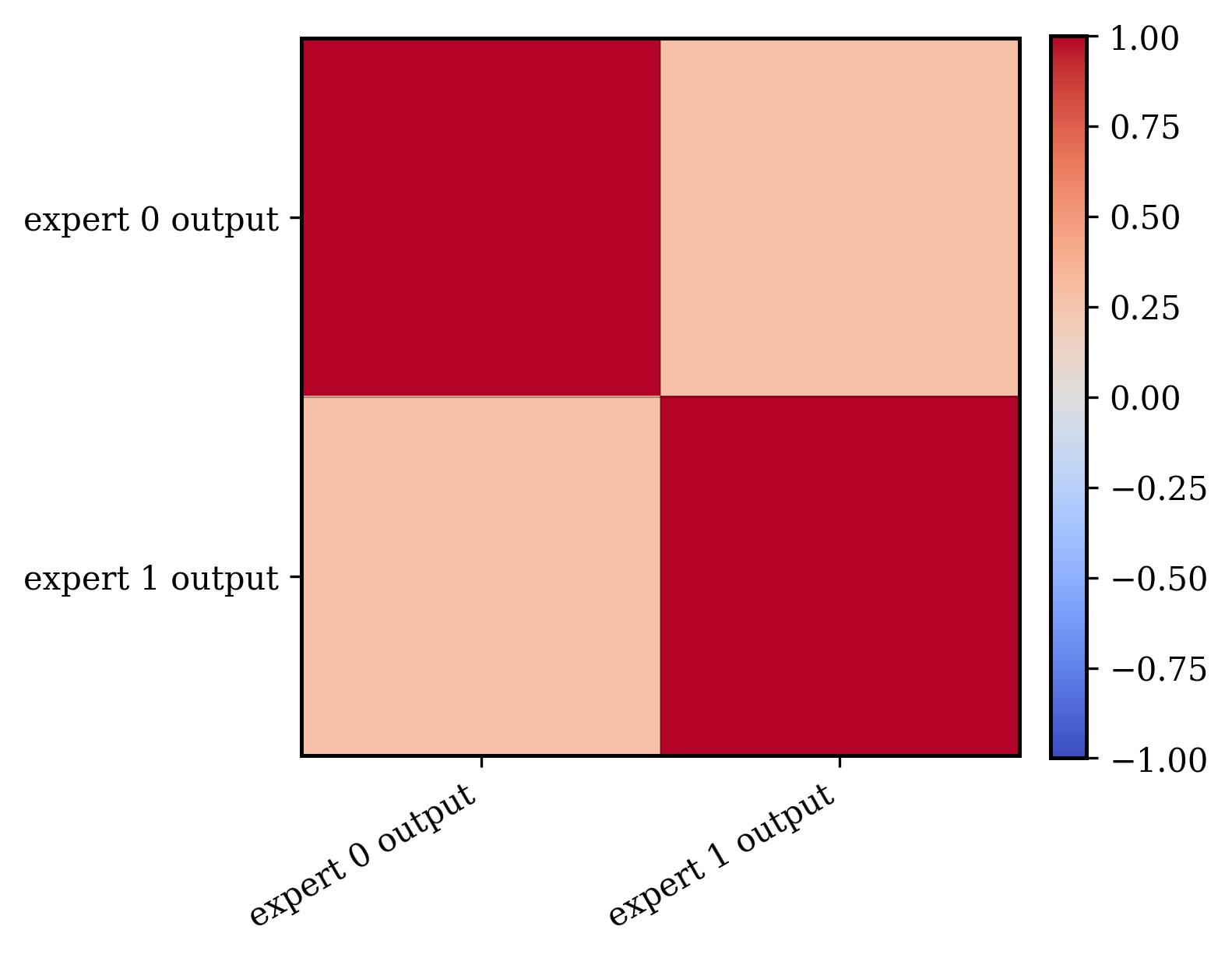}
    \caption*{OD-MoLE $v$-proj}
  \end{subfigure}
  \begin{subfigure}{0.24\textwidth}
    \centering
    \includegraphics[width=\linewidth]{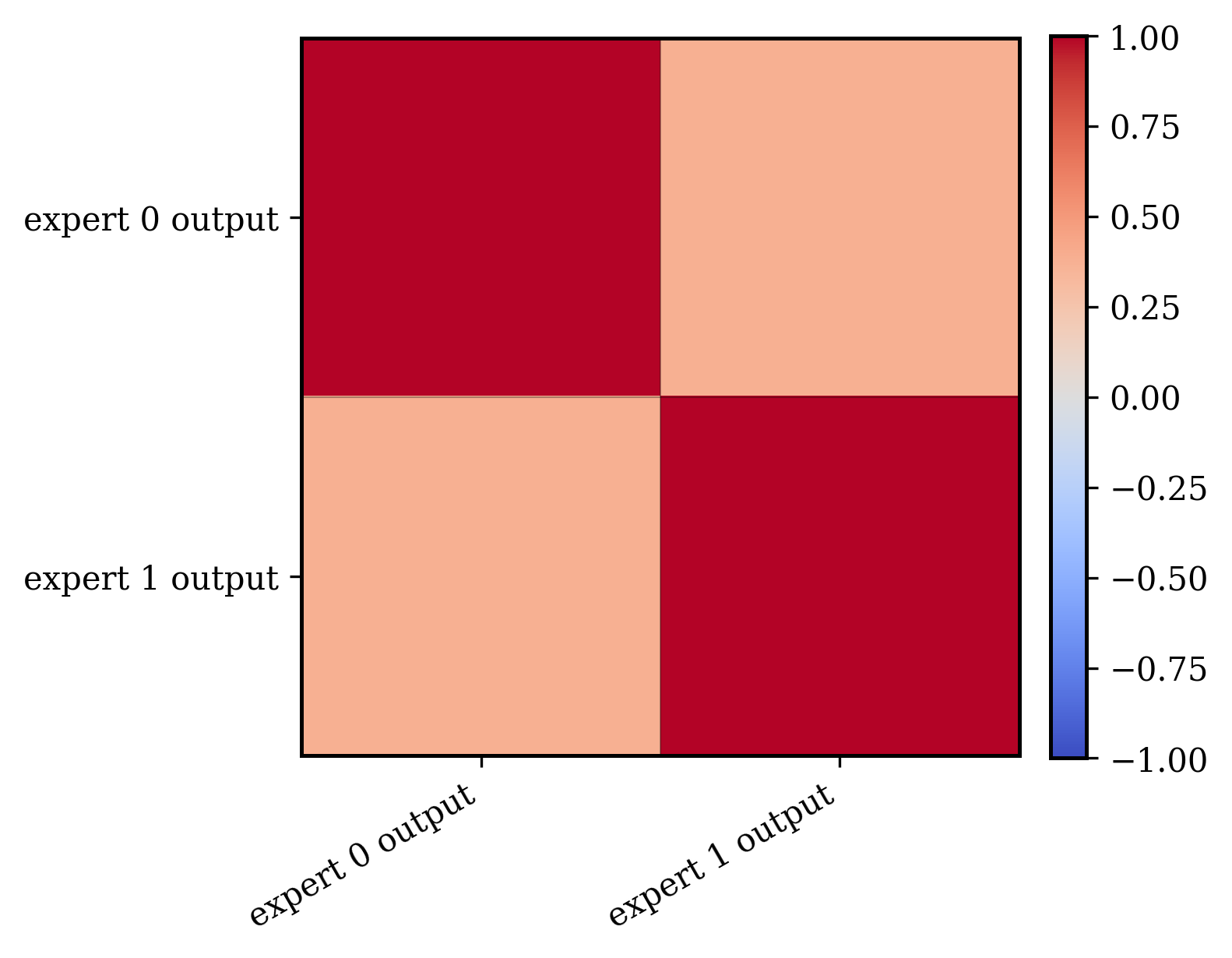}
    \caption*{OD-MoLE $o$-proj}
  \end{subfigure}

  \caption{
  Heatmap of expert output representations $\boldsymbol{B}_i\boldsymbol{A}\boldsymbol{h}$ from four attention projection modules at layer 31.
  The first row shows HydraLoRA, while the second row shows OD-MoLE.
  Compared with HydraLoRA, OD-MoLE exhibits clearer separation between expert output distributions, indicating improved expert specialization and reduced output-level interference.
  }
  \label{fig:heatmap_hydra_odmole_layer31}
\end{figure}

To quantitatively assess expert specialization, we analyze the pairwise cosine similarity between expert output representations $\boldsymbol{B}_i\boldsymbol{A}\boldsymbol{h}$. Figs~\ref{fig:heatmap_hydra_odmole_layer0} and~\ref{fig:heatmap_hydra_odmole_layer31} present the similarity heatmaps across four attention projection modules ($q$, $k$, $v$, and $o$) at both early (layer 0) and deeper (layer 31) layers of Llama-2-7B. We use two experts per submodule and compose their outputs during training. We observe that HydraLoRA consistently exhibits high off-diagonal similarity values across all projection modules, indicating that different experts produce highly correlated outputs and thus learn redundant transformations. In contrast, OD-MoLE substantially reduces off-diagonal cosine similarities, particularly in the $q$- and $k$-projection modules, where the off-diagonal entries approach zero. This indicates that expert outputs become nearly orthogonal, demonstrating that our method effectively encourages diverse and complementary representations. The improvement is consistent across both shallow and deep layers, showing that OD-MoLE promotes specialization throughout the network. Interestingly, both methods exhibit relatively high similarity in the $v$- and $o$-projection modules. Overall, these results provide strong quantitative evidence that OD-MoLE reduces output-level interference and improves expert specialization by enforcing diversity in the representation space, which aligns with the qualitative observations from t-SNE visualizations and the improvements in downstream task performance.

\subsection{Effect of the number of experts}
\label{app:effect_number_of_experts}
We vary $N\in\{8,16,32\}$ while fixing the total expert rank $Nr_i=32$ and routing ratio $K/N=1/4$, keeping the active expert rank at $Kr_i=8$, where $r_i$ is rank of each expert $i$. Tab.~\ref{tab:expert-count} shows that OD-MoLE outperforms HydraLoRA in all three configurations on Qwen2.5-7B and achieves its highest accuracy at $N=8$. Increasing expert count while reducing per-expert rank therefore does not consistently improve performance.

\begin{table}[t]
\caption{Effect of expert count on Qwen2.5-7B across eight commonsense reasoning tasks. The total expert rank $Nr_i=32$ and routing ratio $K/N=1/4$ are fixed, giving an active expert rank of $Kr_i=8$, where $r_i$ is the rank of each expert. Bold indicates the higher accuracy within each setting.}
\label{tab:expert-count}
\centering
\small
\setlength{\tabcolsep}{5pt}
\begin{tabular}{ccccccc}
\toprule
Experts $N$ & Top-$K$ & Rank $r_i$ & Active rank $Kr_i$
& HydraLoRA & OD-MoLE & Gain \\
\midrule
8  & 2 & 4 & 8 & 86.68 & \textbf{87.04} & +0.36 \\
16 & 4 & 2 & 8 & 86.05 & \textbf{86.23} & +0.18 \\
32 & 8 & 1 & 8 & 85.99 & \textbf{86.53} & +0.54 \\
\bottomrule
\end{tabular}
\end{table}

\subsection{Multi-Task Learning with Task-Assigned Experts}
\label{app:task-assigned-routing}

\paragraph{Experimental setup.}
We evaluate OD-MoLE and HydraLoRA in a multi-task setting using Qwen2.5-7B and all eight commonsense reasoning benchmarks. We use 1,000 training examples per task, eight experts with rank four per expert, and five training epochs. Each task $i$ is assigned to one expert-specific reconstruction matrix $\boldsymbol{B}_i$, and all tokens from that task are routed to the assigned expert, bypassing the learned router. The task loss for task $i$ therefore updates $\boldsymbol{B}_i$ and the shared projection $\boldsymbol{A}$. This restriction concerns task-loss gradients and regularization may additionally couple the experts.
\begin{table}[t]
    \caption{Multi-task results on Qwen2.5-7B with one expert assigned to each task. Difference denotes OD-MoLE minus HydraLoRA in percentage points. \textbf{Bold} indicates the higher accuracy in each column.}
    \label{tab:task-assigned-routing}
    \centering
    \small
    \setlength{\tabcolsep}{4pt}
    \resizebox{\linewidth}{!}{%
    \begin{tabular}{l*{9}{c}}
        \toprule
        Method & ComQA & CosQA & SIQA & PIQA
        & ScQA & BoolQ & ARC-C & OBQA & Avg. \\
        \midrule
        HydraLoRA
        & \textbf{84.68} & 79.77 & \textbf{74.77}
        & \textbf{85.50} & 84.84 & 85.29
        & 85.24 & 91.60 & 83.96 \\
        OD-MoLE
        & 84.11 & \textbf{80.74} & 74.72
        & 84.90 & \textbf{85.40} & \textbf{85.57}
        & \textbf{85.75} & \textbf{93.20}
        & \textbf{84.30} \\
        \midrule
        Difference
        & $-0.57$ & $+0.97$ & $-0.05$ & $-0.60$
        & $+0.56$ & $+0.28$ & $+0.51$ & $+1.60$
        & $+0.34$ \\
        \bottomrule
    \end{tabular}%
    }
\end{table}
\paragraph{Results and discussion.}
Tab.~\ref{tab:task-assigned-routing} shows that OD-MoLE achieves an average accuracy of 84.30, exceeding HydraLoRA by 0.34 percentage points and improving five of eight tasks. These results suggest that OD-MoLE's benefits extend to explicit task-to-expert assignment, although improvements are not uniform across tasks.

Task assignment does not eliminate cross-task interaction, since all tasks update the shared projection $\boldsymbol{A}$. The experiment therefore evaluates task-specific reconstruction mappings within a jointly trained shared representation. It requires task identity at inference and does not measure learned-router quality.

\subsection{Comparison with Alternative Specialization Objectives}
\label{app:guo-comparison}

We compare OD-MoLE with the specialization objectives proposed by~\citet{guo2025advancing}. We implement their objectives within the same LoRA MoE framework and evaluate both methods using Qwen2.5-7B, the same datasets and LoRA rank, and two experts per submodule with both experts active. This comparison evaluates the objectives within our experimental setting.

\begin{table}[h!]
    \caption{Comparison with the objectives of~\citet{guo2025advancing} implemented in the same LoRA MoE framework. Both methods use Qwen2.5-7B with all-expert composition. \textbf{Bold} indicates the higher score in each column.}
    \label{tab:guo-comparison}
    \centering
    \small
    \setlength{\tabcolsep}{4pt}
    \resizebox{\linewidth}{!}{%
    \begin{tabular}{l*{9}{c}}
        \toprule
        Method & ComQA & CosQA & SIQA & PIQA
        & ScQA & BoolQ & ARC-C & OBQA & Avg. \\
        \midrule
        Guo et al.\ (adapted)
        & 83.46 & 76.25 & 76.41 & \textbf{87.87}
        & 78.13 & 80.90 & 82.76 & \textbf{95.20}
        & 82.62 \\
        OD-MoLE
        & \textbf{85.50} & \textbf{84.85}
        & \textbf{79.63} & 87.65
        & \textbf{92.58} & \textbf{83.80}
        & \textbf{83.96} & 94.80
        & \textbf{86.60} \\
        \bottomrule
    \end{tabular}%
    }
\end{table}

Tab.~\ref{tab:guo-comparison} shows that OD-MoLE outperforms ~\citet{guo2025advancing} on average. The adapted objectives achieve higher scores on PIQA and OBQA. ~\citet{guo2025advancing} regularizes selected dense-expert representations and routing-score dispersion in a general MoE, while OD-MoLE is designed in the asymmetric LoRA framework. We regularize data-dependent residuals $\boldsymbol{B}_i\boldsymbol{A}\boldsymbol{h}$ to reduce redundancy among expert-specific mappings while separately preserving the capacity of shared low-rank projection $\boldsymbol{A}$. These results support the effectiveness of OD-MoLE's regularization design in the evaluated LoRA MoE setting.

\subsection{Statistical Significance of the Comparison with HydraLoRA}
\label{app:five-seed}
Tab.~\ref{tab:five-seed} compares OD-MoLE with HydraLoRA under selective routing with top-2 over 8 experts on five paired seeds using Qwen2.5-7B. OD-MoLE improves the eight-task average in 4 of 5 seeds, with a mean paired gain of $0.425$ percentage points and a 95\% confidence interval of $[0.039,\,0.811]$ that excludes zero. The mean difference is positive on all eight tasks. Thus, OD-MoLE improves the performance on average.

\begin{table}[t]
\centering
\caption{Paired per-seed comparison with HydraLoRA under selective routing. Each seed cell reports HydraLoRA / OD-MoLE accuracy. $\Delta$ is OD-MoLE minus HydraLoRA in percentage points. Intervals are paired $t$ intervals across the five seed differences. Avg.\ first averages the eight tasks within each seed.}
\label{tab:five-seed}
\small
\resizebox{\linewidth}{!}{%
\begin{tabular}{lccccccc}
\toprule
Task & Seed 40 & Seed 41 & Seed 42 & Seed 43 & Seed 44 & Mean $\Delta$ & 95\% CI \\
\midrule
ComQA & 84.77 / 85.18 & 84.60 / 84.68 & 84.36 / 84.68 & 85.67 / 86.00 & 85.09 / 84.85 & $+0.180$ & $[-0.149,\,+0.509]$ \\
CosQA & 87.67 / 87.81 & 84.57 / 85.76 & 84.93 / 85.71 & 84.03 / 84.14 & 83.89 / 84.53 & $+0.572$ & $[+0.006,\,+1.138]$ \\
SIQA & 79.43 / 79.12 & 79.48 / 79.89 & 79.22 / 80.25 & 79.12 / 80.19 & 78.35 / 79.48 & $+0.666$ & $[-0.102,\,+1.434]$ \\
PIQA & 89.72 / 89.99 & 88.41 / 89.34 & 89.72 / 90.37 & 89.66 / 88.85 & 89.83 / 90.37 & $+0.316$ & $[-0.519,\,+1.151]$ \\
ScQA & 92.30 / 93.61 & 91.70 / 92.82 & 91.26 / 92.26 & 92.96 / 92.96 & 91.98 / 91.93 & $+0.676$ & $[-0.131,\,+1.483]$ \\
BoolQ & 87.06 / 88.20 & 87.40 / 87.92 & 87.58 / 88.50 & 87.86 / 86.21 & 88.01 / 87.89 & $+0.162$ & $[-1.229,\,+1.553]$ \\
ARC-C & 87.97 / 88.57 & 88.74 / 88.99 & 88.05 / 88.48 & 87.54 / 88.05 & 87.97 / 88.31 & $+0.426$ & $[+0.255,\,+0.597]$ \\
OBQA & 95.20 / 96.40 & 95.00 / 95.40 & 95.40 / 95.80 & 95.20 / 95.20 & 96.20 / 96.20 & $+0.400$ & $[-0.208,\,+1.008]$ \\
\midrule
Avg. & 88.015 / 88.610 & 87.4875 / 88.100 & 87.565 / 88.25625 & 87.755 / 87.700 & 87.665 / 87.945 & $+0.425$ & $[+0.039,\,+0.811]$ \\
\bottomrule
\end{tabular}%
}
\end{table}

\subsection{Regularizer Ablation in Per-Seed Results}
\label{app:ablation-per-seed}
Tabs.~\ref{tab:orth-only-seeds}--\ref{tab:pairs-seeds} give the per-seed results behind Tab.~\ref{tab:regularizer-ablation-selective}, all under selective routing on seeds $\{40,41,42\}$. Output orthogonality alone (ortho, no div) improves the average accuracy over (no ortho, no div) in all three seeds, by $0.506$ points on average with a 95\% interval of $[0.038,\,0.974]$. Shared-projection diversity alone (no ortho, div) also improves the average in all three seeds, although its interval $[-0.071,\,0.970]$ includes zero. Restricting the orthogonality loss to the router-selected expert pair gives the nearly identical average accuracy as computing it over all pairs, which indicates that penalizing the unselected experts does not interfere with what the selected experts learn.

\begin{table}[t]
\centering
\caption{Comparison of output orthogonality only (ortho, no div) and (no ortho, no div). Each seed cell reports (no ortho, no div) / (ortho, no div) accuracy. $\Delta$ is (ortho, no div) minus (no ortho, no div), with paired $t$ 95\% intervals across seeds.}
\label{tab:orth-only-seeds}
\small
\setlength{\tabcolsep}{4pt}
\begin{tabular}{lccccc}
\toprule
Task & Seed 40 & Seed 41 & Seed 42 & Mean $\Delta$ & 95\% CI \\
\midrule
ComQA & 84.77 / 85.18 & 84.60 / 84.03 & 84.36 / 85.09 & $+0.190$ & $[-1.493,\,+1.873]$ \\
CosQA & 87.67 / 86.34 & 84.57 / 86.50 & 84.93 / 87.10 & $+0.923$ & $[-3.933,\,+5.780]$ \\
SIQA & 79.43 / 79.63 & 79.48 / 80.09 & 79.22 / 79.53 & $+0.373$ & $[-0.154,\,+0.900]$ \\
PIQA & 89.72 / 89.88 & 88.41 / 89.55 & 89.72 / 89.34 & $+0.307$ & $[-1.607,\,+2.221]$ \\
ScQA & 92.30 / 92.40 & 91.70 / 92.35 & 91.26 / 92.35 & $+0.613$ & $[-0.619,\,+1.846]$ \\
BoolQ & 87.06 / 88.23 & 87.40 / 88.04 & 87.58 / 88.10 & $+0.777$ & $[-0.083,\,+1.636]$ \\
ARC-C & 87.97 / 87.97 & 88.74 / 88.31 & 88.05 / 87.88 & $-0.200$ & $[-0.738,\,+0.338]$ \\
OBQA & 95.20 / 96.80 & 95.00 / 96.00 & 95.40 / 96.00 & $+1.067$ & $[-0.184,\,+2.317]$ \\
\midrule
Avg. & 88.02 / 88.30 & 87.49 / 88.11 & 87.57 / 88.17 & $+0.506$ & $[+0.038,\,+0.974]$ \\
\bottomrule
\end{tabular}
\end{table}

\begin{table}[t]
\centering
\caption{Comparison of shared-projection diversity only (no ortho, div) and (no ortho, no div). Each seed cell reports (no ortho, no div) / (no ortho, div) accuracy. $\Delta$ is (no ortho, div) minus (no ortho, no div), with paired $t$ 95\% intervals across seeds.}
\label{tab:div-only-seeds}
\small
\setlength{\tabcolsep}{4pt}
\begin{tabular}{lccccc}
\toprule
Task & Seed 40 & Seed 41 & Seed 42 & Mean $\Delta$ & 95\% CI \\
\midrule
ComQA & 84.77 / 85.26 & 84.60 / 84.28 & 84.36 / 84.36 & $+0.057$ & $[-0.957,\,+1.070]$ \\
CosQA & 87.67 / 87.14 & 84.57 / 86.63 & 84.93 / 87.71 & $+1.437$ & $[-2.888,\,+5.761]$ \\
SIQA & 79.43 / 79.43 & 79.48 / 80.19 & 79.22 / 79.99 & $+0.493$ & $[-0.571,\,+1.557]$ \\
PIQA & 89.72 / 89.99 & 88.41 / 90.15 & 89.72 / 88.74 & $+0.343$ & $[-3.039,\,+3.725]$ \\
ScQA & 92.30 / 92.82 & 91.70 / 91.79 & 91.26 / 92.02 & $+0.457$ & $[-0.387,\,+1.300]$ \\
BoolQ & 87.06 / 87.28 & 87.40 / 88.04 & 87.58 / 87.71 & $+0.330$ & $[-0.346,\,+1.006]$ \\
ARC-C & 87.97 / 88.40 & 88.74 / 88.74 & 88.05 / 88.65 & $+0.343$ & $[-0.425,\,+1.112]$ \\
OBQA & 95.20 / 95.60 & 95.00 / 95.20 & 95.40 / 95.20 & $+0.133$ & $[-0.626,\,+0.892]$ \\
\midrule
Avg. & 88.02 / 88.24 & 87.49 / 88.13 & 87.57 / 88.05 & $+0.449$ & $[-0.071,\,+0.970]$ \\
\bottomrule
\end{tabular}
\end{table}

\begin{table}[t]
\centering
\caption{Comparison of orthogonality loss over all expert pairs and the router-selected pair only, with $\lambda_{\mathrm{div}}=0$. Each seed cell reports all pairs / selected pair accuracy. $\Delta$ is selected pair minus all pairs, with paired $t$ 95\% intervals across seeds.}
\label{tab:pairs-seeds}
\small
\setlength{\tabcolsep}{4pt}
\begin{tabular}{lccccc}
\toprule
Task & Seed 40 & Seed 41 & Seed 42 & Mean $\Delta$ & 95\% CI \\
\midrule
ComQA & 85.18 / 85.18 & 84.03 / 84.03 & 85.09 / 84.28 & $-0.270$ & $[-1.432,\,+0.892]$ \\
CosQA & 86.34 / 87.94 & 86.50 / 86.30 & 87.10 / 88.34 & $+0.880$ & $[-1.486,\,+3.246]$ \\
SIQA & 79.63 / 79.32 & 80.09 / 79.22 & 79.53 / 80.19 & $-0.173$ & $[-2.096,\,+1.750]$ \\
PIQA & 89.88 / 89.93 & 89.55 / 89.12 & 89.34 / 88.85 & $-0.290$ & $[-1.025,\,+0.445]$ \\
ScQA & 92.40 / 92.91 & 92.35 / 92.54 & 92.35 / 92.12 & $+0.157$ & $[-0.765,\,+1.079]$ \\
BoolQ & 88.23 / 87.28 & 88.04 / 88.10 & 88.10 / 88.53 & $-0.153$ & $[-1.928,\,+1.621]$ \\
ARC-C & 87.97 / 87.46 & 88.31 / 88.91 & 87.88 / 88.57 & $+0.260$ & $[-1.400,\,+1.920]$ \\
OBQA & 96.80 / 95.40 & 96.00 / 95.80 & 96.00 / 96.20 & $-0.467$ & $[-2.535,\,+1.602]$ \\
\midrule
Avg. & 88.30 / 88.18 & 88.11 / 88.00 & 88.17 / 88.39 & $-0.007$ & $[-0.477,\,+0.463]$ \\
\bottomrule
\end{tabular}
\end{table}

\subsection{Analysis of Output Orthogonality}
\label{app:orth-analysis}
All analyses in this subsection use selective routing and compare (no ortho, no div) with applying only orthogonality (ortho, no div), where $\lambda_{\mathrm{orth}}=0.001$ and $\lambda_{\mathrm{div}}=0$.

\paragraph{Routing sensitivity.}
Tab.~\ref{tab:orth-mechanism} reports the median of the routing sensitivity $S$. Tab.~\ref{tab:orth-p05} adds its fifth percentile, which describes the least sensitive tokens. The fifth percentile increases on five tasks and decreases on three, so the orthogonality penalty improves typical routing distinguishability.

\begin{table}[t]
\centering
\caption{Fifth percentile of the routing sensitivity $S$ under the setting of Tab.~\ref{tab:orth-mechanism}. Orth. and Div. denote output orthogonality and shared-projection diversity.}
\label{tab:orth-p05}
\footnotesize
\begin{tabular}{cc*{8}{c}}
\toprule
Orth. & Div. & ComQA & CosQA & SIQA & PIQA & ScQA & BoolQ & ARC-C & OBQA \\
\midrule
$\times$ & $\times$ & 0.176 & 0.853 & 0.408 & 0.673 & 0.832 & 0.346 & 0.309 & 0.197 \\
$\checkmark$ & $\times$ & 0.186 & 0.843 & 0.415 & 0.677 & 0.718 & 0.328 & 0.319 & 0.260 \\
\bottomrule
\end{tabular}
\end{table}

\paragraph{Nonvanishing expert outputs.}
Theorem~\ref{thm:routing-recovery} assumes a positive lower bound on the expert output norms, which the training objective does not enforce. To check whether near-zero outputs occur in practice, we compute, for each evaluated token and projection, the minimum expert output norm before multiplication by its routing weight, over all eight experts and over the two selected experts in Tab.~\ref{tab:min-norm}. The fraction of minimum norms below $10^{-8}$ is $0\%$ on all eight tasks in both models and for both sets of experts. With orthogonality, the median minimum norm increases on 7 of 8 tasks in both cases. These measurements support the nonvanishing-output assumption on the evaluated activations, and the routing-recovery guarantee remains conditional on it.

\begin{table}[t]
\centering
\caption{Minimum expert output norm per evaluated token and projection, over all eight experts and over the two router-selected experts. Median and p05 are the median and fifth percentile across tokens and projections. No minimum norm falls below $10^{-8}$ in any setting.}
\label{tab:min-norm}
\footnotesize
\resizebox{\linewidth}{!}{%
\begin{tabular}{llc*{8}{c}}
\toprule
Experts & Statistic & Orth. & ComQA & CosQA & SIQA & PIQA & ScQA & BoolQ & ARC-C & OBQA \\
\midrule
\multirow{4}{*}{All 8} & \multirow{2}{*}{Median} & $\times$ & 0.533 & 3.036 & 1.415 & 2.024 & 2.669 & 1.213 & 1.181 & 1.563 \\
 &  & $\checkmark$ & 0.566 & 3.112 & 1.457 & 2.167 & 2.730 & 1.255 & 1.219 & 1.457 \\
\cmidrule(lr){2-11}
 & \multirow{2}{*}{p05} & $\times$ & 0.145 & 0.688 & 0.358 & 0.373 & 0.534 & 0.262 & 0.221 & 0.145 \\
 &  & $\checkmark$ & 0.158 & 0.696 & 0.337 & 0.382 & 0.600 & 0.246 & 0.211 & 0.201 \\
\midrule
\multirow{4}{*}{Selected 2} & \multirow{2}{*}{Median} & $\times$ & 0.622 & 3.450 & 1.710 & 2.540 & 3.174 & 1.386 & 1.385 & 1.804 \\
 &  & $\checkmark$ & 0.681 & 3.550 & 1.781 & 2.596 & 3.180 & 1.475 & 1.401 & 1.791 \\
\cmidrule(lr){2-11}
 & \multirow{2}{*}{p05} & $\times$ & 0.178 & 0.816 & 0.409 & 0.624 & 0.815 & 0.310 & 0.300 & 0.188 \\
 &  & $\checkmark$ & 0.187 & 0.830 & 0.397 & 0.647 & 0.676 & 0.326 & 0.303 & 0.250 \\
\bottomrule
\end{tabular}%
}
\end{table}

\paragraph{Routing distributions.}
Tab.~\ref{tab:routing-dist} reports the most frequently selected expert and the most frequently co-selected expert pair in each attention projection at layers 7, 14 and 21 for BoolQ. In both models, the most selected expert is chosen for $27$--$31\%$ of tokens and the most frequent pair for $4.5$--$7.7\%$, close to the values of $25\%$ and $3.6\%$ that uniform top-2 routing over 8 experts would give. Routing therefore does not collapse onto a few experts with or without the orthogonality penalty.

\begin{table}[t]
\centering
\caption{Routing distributions on BoolQ under selective routing. For each layer and attention projection, we report the most frequently selected expert and the most frequently co-selected expert pair, with their selection frequencies, without ($\times$) and with ($\checkmark$) output orthogonality.}
\label{tab:routing-dist}
\small
\begin{tabular}{clcccc}
\toprule
Layer & Projection & Top expert ($\times$) & Top expert ($\checkmark$) & Top pair ($\times$) & Top pair ($\checkmark$) \\
\midrule
7 & \texttt{q\_proj} & E4 (29.1\%) & E1 (28.2\%) & E0--E1 (6.7\%) & E3--E4 (6.6\%) \\
7 & \texttt{k\_proj} & E1 (28.2\%) & E6 (27.8\%) & E1--E7 (5.0\%) & E4--E6 (6.2\%) \\
7 & \texttt{v\_proj} & E1 (28.2\%) & E3 (29.6\%) & E1--E4 (6.5\%) & E0--E3 (5.1\%) \\
7 & \texttt{o\_proj} & E2 (29.7\%) & E7 (31.2\%) & E3--E7 (6.1\%) & E5--E7 (6.2\%) \\
\midrule
14 & \texttt{q\_proj} & E1 (30.2\%) & E6 (27.6\%) & E3--E6 (6.0\%) & E4--E5 (6.5\%) \\
14 & \texttt{k\_proj} & E4 (28.4\%) & E2 (28.7\%) & E2--E4 (5.9\%) & E4--E5 (5.4\%) \\
14 & \texttt{v\_proj} & E0 (27.0\%) & E4 (26.8\%) & E1--E3 (5.0\%) & E1--E3 (6.4\%) \\
14 & \texttt{o\_proj} & E6 (29.4\%) & E7 (29.3\%) & E0--E2 (7.7\%) & E0--E6 (6.3\%) \\
\midrule
21 & \texttt{q\_proj} & E5 (31.1\%) & E5 (26.9\%) & E3--E7 (5.5\%) & E5--E6 (4.8\%) \\
21 & \texttt{k\_proj} & E1 (28.5\%) & E7 (29.3\%) & E1--E5 (5.3\%) & E4--E7 (5.5\%) \\
21 & \texttt{v\_proj} & E7 (28.8\%) & E0 (29.9\%) & E2--E7 (4.8\%) & E1--E7 (5.4\%) \\
21 & \texttt{o\_proj} & E6 (28.0\%) & E5 (27.1\%) & E3--E6 (5.8\%) & E3--E5 (4.5\%) \\
\bottomrule
\end{tabular}
\end{table}

\paragraph{Expert removal and replacement.}
To test whether the two selected experts make complementary contributions, we intervene on BoolQ at one layer at a time, jointly across the four attention projections. We either remove the contribution of the higher- or lower-weight selected expert, or replace it with the response of its co-selected partner, rescaled to the original response norm and applied at the original routing weight in Tab.~\ref{tab:intervention}. For OD-MoLE, removal lowers accuracy in 4 of 6 conditions and replacement in 5 of 6 conditions. For HydraLoRA, each intervention lowers accuracy in only 1 of 6 conditions, and accuracy slightly increases on average. The consistent direction suggests that the experts selected by OD-MoLE are less redundant than those of HydraLoRA.

\begin{table}[t]
\centering
\caption{Expert removal and norm-matched partner replacement on BoolQ under selective routing. Slot is the routing slot of the targeted expert (higher or lower routing weight). Accuracy drop is the original accuracy minus the accuracy after intervention, in percentage points, where positive accuracy drops and positive NLL increases indicate worse predictions after the intervention.}
\label{tab:intervention}
\small
\resizebox{\linewidth}{!}{%
\begin{tabular}{cllcccc}
\toprule
Layer & Slot & Intervention & \multicolumn{1}{c}{HydraLoRA acc.\ drop} & \multicolumn{1}{c}{OD-MoLE acc.\ drop} & \multicolumn{1}{c}{HydraLoRA $\Delta$NLL} & \multicolumn{1}{c}{OD-MoLE $\Delta$NLL} \\
\midrule
7 & Higher & Removal & $-0.3364$ & $+0.0917$ & $+0.0020$ & $+0.0065$ \\
7 & Higher & Replacement & $-0.2141$ & $+0.2752$ & $+0.0068$ & $+0.0075$ \\
7 & Lower & Removal & $-0.3058$ & 0.0000 & $-0.0010$ & $-0.0014$ \\
7 & Lower & Replacement & $-0.3976$ & $-0.0612$ & $-0.0033$ & $-0.0033$ \\
\midrule
14 & Higher & Removal & $-0.5505$ & $+0.2446$ & $-0.0048$ & $+0.0044$ \\
14 & Higher & Replacement & $-0.2141$ & $+0.2141$ & $-0.0040$ & $+0.0020$ \\
14 & Lower & Removal & $-0.6116$ & $+0.1223$ & $-0.0082$ & $+0.0035$ \\
14 & Lower & Replacement & $-0.4281$ & $+0.1835$ & $-0.0046$ & $+0.0011$ \\
\midrule
21 & Higher & Removal & $-0.5199$ & $-0.1835$ & $-0.0092$ & $+0.0064$ \\
21 & Higher & Replacement & $-0.4281$ & $+0.0612$ & $-0.0073$ & $+0.0040$ \\
21 & Lower & Removal & $+0.0917$ & $+0.0306$ & $-0.0012$ & $-0.0007$ \\
21 & Lower & Replacement & $+0.3058$ & $+0.1835$ & $+0.0007$ & $+0.0023$ \\
\bottomrule
\end{tabular}%
}
\end{table}

\subsection{Analysis of Shared-Projection Diversity}
\label{app:div-analysis}
The spectrum analyses in this subsection use all-expert composition with output orthogonality disabled, and summarize the 112 shared projection matrices of each run.

\paragraph{Spectrum during training.}
Tab.~\ref{tab:spectrum-epochs} tracks the shared projection $\boldsymbol{A}$ at initialization and at the end of each epoch, without and with the diversity regularizer ($\lambda_{\mathrm{div}}=0.001$). On both tasks, the Frobenius norm grows during training and ends slightly higher with the regularizer, so part of its effect is on scale. Effective rank, stable rank, and the normalized spectrum are invariant to uniform rescaling of $\boldsymbol{A}$. Effective and stable rank decline in both runs but more slowly with the regularizer, and the mean of the normalized median spectrum is also higher at every epoch. The regularizer therefore changes the shape of the spectrum and not only its scale, which is the shape term in Corollary~\ref{cor:shared-diversity-scale}.

\begin{table}[t]
\centering
\caption{Norm and singular-value spectrum of the shared projection $\boldsymbol{A}$ during training, without ($\times$) and with ($\checkmark$) shared-projection diversity ($\lambda_{\mathrm{div}}=0.001$). Each cell reports mean / max over 112 shared projection matrices. Epoch 0 is the initialization. Norm.\ spectrum is the per-index median singular value normalized by $\|\boldsymbol{A}\|_{\mathrm{F}}$.}
\label{tab:spectrum-epochs}
\small
\resizebox{\linewidth}{!}{%
\begin{tabular}{lcccccc}
\toprule
Task & Div. & Epoch & $\|\boldsymbol{A}\|_{\mathrm{F}}$ & Eff. rank & Stable rank & Norm.\ spectrum \\
\midrule
\multirow{6}{*}{SIQA} & \multirow{3}{*}{$\times$} & 0 & 2.3089 / 2.3194 & 15.9649 / 15.9764 & 14.3419 / 14.6667 & 0.24989 / 0.26392 \\
 &  & 1 & 2.6812 / 3.0775 & 14.8210 / 15.5969 & 6.8792 / 10.0413 & 0.24553 / 0.36576 \\
 &  & 2 & 3.1958 / 4.0492 & 13.3581 / 14.7045 & 5.2882 / 7.7874 & 0.23962 / 0.40654 \\
\cmidrule(lr){2-7}
 & \multirow{3}{*}{$\checkmark$} & 0 & 2.3089 / 2.3194 & 15.9649 / 15.9764 & 14.3419 / 14.6667 & 0.24989 / 0.26392 \\
 &  & 1 & 2.6955 / 3.0807 & 15.0661 / 15.7013 & 7.3503 / 10.4430 & 0.24635 / 0.38040 \\
 &  & 2 & 3.2623 / 4.2240 & 14.1296 / 15.1488 & 5.9818 / 8.7138 & 0.24305 / 0.43637 \\
\midrule
\multirow{8}{*}{PIQA} & \multirow{4}{*}{$\times$} & 0 & 2.3089 / 2.3194 & 15.9649 / 15.9764 & 14.3419 / 14.6667 & 0.24989 / 0.26392 \\
 &  & 1 & 2.5999 / 2.8697 & 14.6758 / 15.6223 & 6.2983 / 10.2325 & 0.24532 / 0.39706 \\
 &  & 2 & 2.9368 / 3.5961 & 13.3540 / 14.7972 & 4.8435 / 7.9145 & 0.24103 / 0.44940 \\
 &  & 3 & 3.2760 / 4.2558 & 12.1875 / 13.9444 & 4.2116 / 6.5360 & 0.23542 / 0.47673 \\
\cmidrule(lr){2-7}
 & \multirow{4}{*}{$\checkmark$} & 0 & 2.3089 / 2.3194 & 15.9649 / 15.9764 & 14.3419 / 14.6667 & 0.24989 / 0.26392 \\
 &  & 1 & 2.6088 / 2.8766 & 14.7700 / 15.6293 & 6.4475 / 9.7441 & 0.24589 / 0.40120 \\
 &  & 2 & 2.9229 / 3.5557 & 13.7568 / 15.0367 & 5.1794 / 7.9312 & 0.24256 / 0.46390 \\
 &  & 3 & 3.3494 / 4.3883 & 12.9154 / 14.4494 & 4.6261 / 7.4242 & 0.23907 / 0.49564 \\
\bottomrule
\end{tabular}%
}
\end{table}

\paragraph{Condition number.}
Tab.~\ref{tab:kappa-epochs} reports the spectral condition number $\kappa_2(\boldsymbol{A})$ on PIQA. The condition number grows during training in both runs. With the regularizer, both its median and its 90th percentile are lower after every training epoch, and the gap widens over training to about $10\%$ at the final epoch.

\begin{table}[t]
\centering
\caption{Spectral condition number $\kappa_2(\boldsymbol{A})$ during training on PIQA, without ($\times$) and with ($\checkmark$) shared-projection diversity ($\lambda_{\mathrm{div}}=0.001$). Median and p90 are taken over 112 shared projection matrices.}
\label{tab:kappa-epochs}
\small
\begin{tabular}{cccc}
\toprule
Div. & Epoch & Median $\kappa_2$ & p90 $\kappa_2$ \\
\midrule
\multirow{4}{*}{$\times$} & 0 & 1.1196 & 1.1325 \\
 & 1 & 1.8845 & 2.2502 \\
 & 2 & 2.3881 & 2.7517 \\
 & 3 & 2.8333 & 3.1920 \\
\midrule
\multirow{4}{*}{$\checkmark$} & 0 & 1.1196 & 1.1325 \\
 & 1 & 1.8560 & 2.2313 \\
 & 2 & 2.2328 & 2.5952 \\
 & 3 & 2.5447 & 2.8726 \\
\bottomrule
\end{tabular}
\end{table}

\paragraph{Sensitivity to $\lambda_{\mathrm{div}}$.}
Tab.~\ref{tab:lambda-sweep} varies $\lambda_{\mathrm{div}}$ and reports the spectrum at the end of training. The mean stable rank increases monotonically with $\lambda_{\mathrm{div}}$ on both tasks, so a stronger penalty retains more spectral energy outside the leading singular direction. Accuracy varies by a few tenths of a point across $\lambda_{\mathrm{div}}$. A better-conditioned shared projection is thus not sufficient by itself for higher accuracy, and the other experiments in this appendix use $\lambda_{\mathrm{div}}=0.001$.

\begin{table}[t]
\centering
\caption{Effect of $\lambda_{\mathrm{div}}$ on the shared projection at the end of training in a single run. Stable rank and raw singular values are reported as mean / max, and the raw singular value is the per-index median across 112 shared projection matrices.}
\label{tab:lambda-sweep}
\small
\begin{tabular}{lcccc}
\toprule
Task & $\lambda_{\mathrm{div}}$ & Stable rank & Raw singular value & Accuracy \\
\midrule
\multirow{5}{*}{SIQA} & 0 & 5.2882 / 7.7874 & 0.70820 / 1.08094 & 80.04 \\
 & 0.001 & 5.9818 / 8.7138 & 0.76706 / 1.30416 & 80.14 \\
 & 0.01 & 6.0256 / 8.9137 & 0.76219 / 1.27646 & 79.94 \\
 & 0.1 & 6.1871 / 9.0568 & 0.75281 / 1.24540 & 80.30 \\
 & 1 & 7.3886 / 11.2498 & 0.77154 / 1.40283 & 80.09 \\
\midrule
\multirow{5}{*}{PIQA} & 0 & 4.2116 / 6.5360 & 0.71079 / 1.23412 & 89.72 \\
 & 0.001 & 4.6261 / 7.4242 & 0.77307 / 1.52501 & 90.21 \\
 & 0.01 & 4.6340 / 7.5714 & 0.77097 / 1.51441 & 89.77 \\
 & 0.1 & 4.7589 / 7.4488 & 0.76204 / 1.48070 & 89.66 \\
 & 1 & 5.8482 / 10.7028 & 0.77909 / 1.63407 & 89.61 \\
\bottomrule
\end{tabular}
\end{table}

\paragraph{Normalized log-determinant penalty.}
Tab.~\ref{tab:normalized-logdet} compares the raw log-determinant penalty with a variant computed on the normalized matrix $\bar{\boldsymbol{A}}=\sqrt{r}\,\boldsymbol{A}/\max(\|\boldsymbol{A}\|_{\mathrm{F}},10^{-12})$, which satisfies the scale condition of Theorem~\ref{thm:shared-diversity} by construction. The full matrix is normalized before its columns are sampled, and both variants use $\lambda_{\mathrm{div}}=0.001$ and $\lambda_{\mathrm{orth}}=0.001$. The two variants reach comparable average accuracy in both routing settings, so enforcing the scale condition does not change the empirical conclusions.

\begin{table}[t]
\centering
\caption{Comparison of raw and normalized log-determinant penalty in a single run.}
\label{tab:normalized-logdet}
\footnotesize
\resizebox{\linewidth}{!}{%
\begin{tabular}{ll*{9}{c}}
\toprule
Routing & Penalty & ComQA & CosQA & SIQA & PIQA & ScQA & BoolQ & ARC-C & OBQA & Avg. \\
\midrule
\multirow{2}{*}{All experts} & Normalized & 85.75 & 88.41 & 80.35 & 90.32 & 91.65 & 86.94 & 88.65 & 95.80 & 88.48 \\
 & Raw & 85.83 & 88.51 & 80.19 & 90.21 & 92.02 & 87.03 & 88.48 & 96.00 & 88.53 \\
\midrule
\multirow{2}{*}{Selective} & Normalized & 84.77 & 87.14 & 79.89 & 89.61 & 92.63 & 87.83 & 88.05 & 95.60 & 88.19 \\
 & Raw & 85.18 & 87.81 & 79.12 & 89.99 & 93.61 & 88.20 & 88.57 & 96.40 & 88.61 \\
\bottomrule
\end{tabular}%
}
\end{table}

\subsection{Mathematical Reasoning and Code Generation}
\label{app:more-tasks}
To evaluate beyond the commonsense benchmarks, we train on MetaMathQA and report exact-match accuracy on GSM8K, and we report pass@1 on HumanEval in Tab.~\ref{tab:more-tasks}. For GSM8K, OD-MoLE improves over HydraLoRA in both routing settings, by $0.68$ points in all-expert composition and $1.90$ points in selective routing. For HumanEval, OD-MoLE is higher by $0.61$ points in both settings.

\begin{table}[t]
\centering
\caption{Mathematical reasoning and code generation using Qwen2.5-7B in a single run. $\Delta$ is OD-MoLE minus HydraLoRA in percentage points.}
\label{tab:more-tasks}
\small
\begin{tabular}{lllccc}
\toprule
Benchmark & Metric & Routing & HydraLoRA (\%) & OD-MoLE (\%) & $\Delta$ \\
\midrule
GSM8K & Exact match & All experts & 82.79 & 83.47 & $+0.68$ \\
GSM8K & Exact match & Selective & 81.65 & 83.55 & $+1.90$ \\
\midrule
HumanEval & Pass@1 & All experts & 59.15 & 59.76 & $+0.61$ \\
HumanEval & Pass@1 & Selective & 57.32 & 57.93 & $+0.61$ \\
\bottomrule
\end{tabular}
\end{table}

\subsection{Inference Latency}
\label{app:latency}
Both regularizers are applied only during training, so OD-MoLE and HydraLoRA perform the same computation at inference. Tab.~\ref{tab:latency-bench} reports synchronized wall-clock latency per request on the eight benchmarks with batch size 1, using the same 128 held-out prompts for all configurations after five warm-up requests each. The median latency of OD-MoLE is within $2.3\%$ of that of HydraLoRA in 15 of the 16 tasks and routing combinations. Tab.~\ref{tab:latency-gen} reports generation latency on GSM8K and HumanEval, where the two methods are again close. In our implementation, selective routing has higher latency than all-expert composition for both methods.

\begin{table}[t]
\centering
\caption{Inference latency on the eight benchmarks with Qwen2.5-7B and batch size 1. Each cell reports median / 95th-percentile latency per request in milliseconds over held-out prompts.}
\label{tab:latency-bench}
\small
\begin{tabular}{lcccc}
\toprule
 & \multicolumn{2}{c}{All-expert composition} & \multicolumn{2}{c}{Selective routing} \\
\cmidrule(lr){2-3}\cmidrule(lr){4-5}
Task & HydraLoRA & OD-MoLE & HydraLoRA & OD-MoLE \\
\midrule
ComQA & 185.92 / 196.12 & 183.59 / 191.34 & 613.04 / 635.67 & 613.25 / 633.57 \\
CosQA & 206.15 / 264.60 & 241.75 / 267.27 & 660.01 / 690.79 & 659.32 / 698.27 \\
SIQA & 182.04 / 203.74 & 181.01 / 201.02 & 373.95 / 612.40 & 373.95 / 394.35 \\
PIQA & 129.01 / 180.98 & 129.02 / 180.50 & 374.10 / 428.03 & 375.27 / 429.95 \\
ScQA & 128.06 / 147.63 & 128.07 / 147.57 & 372.70 / 394.28 & 372.57 / 394.17 \\
BoolQ & 179.81 / 224.49 & 179.84 / 226.21 & 427.93 / 473.11 & 421.32 / 468.78 \\
ARC-C & 130.51 / 149.01 & 133.47 / 149.71 & 381.24 / 399.17 & 379.17 / 397.60 \\
OBQA & 128.88 / 148.14 & 129.10 / 147.95 & 375.42 / 396.52 & 374.26 / 395.03 \\
\bottomrule
\end{tabular}
\end{table}

\begin{table}[t]
\centering
\caption{Generation latency on GSM8K and HumanEval using Qwen2.5-7B. Each cell reports median / 95th-percentile wall time per request in seconds.}
\label{tab:latency-gen}
\small
\begin{tabular}{llcc}
\toprule
Benchmark & Routing & HydraLoRA & OD-MoLE \\
\midrule
GSM8K & All experts & 5.128 / 5.387 & 5.116 / 5.313 \\
GSM8K & Selective & 9.412 / 9.971 & 9.379 / 9.926 \\
HumanEval & All experts & 5.181 / 5.338 & 5.098 / 5.472 \\
HumanEval & Selective & 9.403 / 9.886 & 9.403 / 9.755 \\
\bottomrule
\end{tabular}
\end{table}

\section{Parameter Analysis}\label{subsec:param_num}
We provide a parameter analysis for each baseline and our proposed method based on the Llama-2-7B model in Tab.~\ref{tab:parameter_comparison}. We assume $d$ represents the model dimension, $r$ denotes the rank of LoRA, $N$ indicates the number of experts, $L$ indicates the number of layers, top-$k$ is top-2, and $a$ is the selected number of experts using threshold in AdaMoLE. LoRA is applied to the attention projection matrices $\boldsymbol{W}_q,\boldsymbol{W}_k,\boldsymbol{W}_v,\boldsymbol{W}_o$, and only the corresponding LoRA parameters are trainable. The proportion is measured relative to the total number of whole model parameters.

\begin{table*}[t]
    \centering
    \caption{Comparison of trainable and activated parameters
    across vanilla LoRA with varying rank and different adaptation methods.}
    \label{tab:parameter_comparison}
    \renewcommand{\arraystretch}{1.25}
    \setlength{\tabcolsep}{8pt}

    \resizebox{\textwidth}{!}{%
        \begin{tabular}{lcccc}
            \toprule
            \textbf{Method}
            & \multicolumn{2}{c}{\textbf{Trainable parameters}}
            & \multicolumn{2}{c}{\textbf{Activated parameters}} \\
            \cmidrule(lr){2-3}
            \cmidrule(lr){4-5}
            & Count & Proportion & Count & Proportion \\
            \midrule

            \rowcolor{lightblue}
            LoRA ($r=8$)
            & $8drL$
            & $0.12\%$
            & ---
            & --- \\

            LoRA ($r=16$)
            & $8drL$
            & $0.25\%$
            & ---
            & --- \\

            \rowcolor{lightblue}
            LoRA ($r=32$)
            & $8drL$
            & $0.50\%$
            & ---
            & --- \\

            MoLoRA
            & $8drNL + 4dNL$
            & $0.56\%$
            & $8drkL + 4dNL$
            & $0.19\%$ \\

            \rowcolor{lightblue}
            GOAT
            & $8drNL + 4dNL$
            & $0.56\%$
            & $8drkL + 4dNL$
            & $0.19\%$ \\

            AdaMoLE
            & $8drNL + 4dNL$
            & $0.56\%$
            & $8drNaL + 4dNL$
            & $0.19\%$ \\

            \rowcolor{lightblue}
            HydraLoRA
            & $4(N+1)drL + 4dNL$
            & $0.34\%$
            & $4(N+1)drL + 4dNL$
            & $0.34\%$ \\

            OD-MoLE
            & $4(N+1)drL + 4dNL$
            & $0.34\%$
            & $4(N+1)drL + 4dNL$
            & $0.34\%$ \\

            \bottomrule
        \end{tabular}%
    }
\end{table*}
\end{document}